\documentclass[twoside,11pt]{article}

\usepackage{jmlr2e}

\usepackage{amsmath,amsbsy,amsfonts,amssymb,dsfont,units}
\usepackage{algorithm,algorithmic}
\usepackage{tikz}
\usetikzlibrary{calc,shapes}
\usepackage{subfigure}
\usepackage[colorlinks=false,allbordercolors={1 1 1}]{hyperref}

\newtheorem{lem}{Lemma}[section]
\newtheorem{prop}{Proposition}[section]
\newtheorem{thm}{Theorem}[section]
\newtheorem{condition}{Condition}[section]

\newenvironment{thm2}[1]{\def\thetheorem{\ref{#1} restated}\begin{thm}}{\end{thm}\addtocounter{theorem}{-1}}

\newcommand\E{\mathbb{E}}
\newcommand\R{\mathbb{R}}
\newcommand\N{\mathcal{N}}

\newcommand\calL{\mathcal{L}}

\renewcommand\t{{\scriptscriptstyle\top}}

\DeclareMathOperator{\diag}{diag}
\newcommand\wt{\widetilde}

\newcommand\eps{\epsilon}
\newcommand\veps{\varepsilon}
\newcommand\tl{\tilde}
\newcommand\teps{\tl{\eps}}
\newcommand\poly{\operatorname{poly}}
\newcommand\sign{\operatorname{sign}}

\newcommand\med{\operatorname{median}}
\newcommand\hv{\ensuremath{\hat{v}}}
\newcommand\hlambda{\ensuremath{\hat{\lambda}}}
\newcommand\tlambda{\ensuremath{\tilde{\lambda}}}
\newcommand\tlambdamax{\ensuremath{\tilde{\lambda}_{\max}}}
\newcommand\tlambdamin{\ensuremath{\tilde{\lambda}_{\min}}}
\newcommand\tlambdaavg{\ensuremath{\tilde{\lambda}_{\operatorname{avg}}}}
\newcommand\oalpha{\ensuremath{\overline{\alpha}}}
\newcommand\obeta{\ensuremath{\overline{\beta}}}
\newcommand\odelta{\ensuremath{\overline{\delta}}}
\renewcommand\th[1]{\ensuremath{\theta_{#1}}}
\newcommand\ut[1]{\ensuremath{\check{\theta}_{#1}}}
\newcommand\lambdamax{\ensuremath{\lambda_{\max}}}
\newcommand\lambdamin{\ensuremath{\lambda_{\min}}}
\newcommand\event{\mathsf{Event}}
\newcommand\deflate{\mathcal{E}}
\newcommand\vectorize{\operatorname{vec}}

\newcommand\Dir{\operatorname{Dir}}
\newcommand\coherence{\operatorname{coherence}}

\jmlrheading{15}{2014}{2773-2832}{2/13; Revised 3/14}{8/14}{Animashree Anandkumar, Rong Ge, Daniel Hsu, Sham M. Kakade, and Matus Telgarsky}
\ShortHeadings
{Tensor Decompositions for Learning Latent Variable Models}
{Anandkumar, Ge, Hsu, Kakade, and Telgarsky}
\firstpageno{2773}

\begin{document}
\title{Tensor Decompositions for Learning Latent Variable Models}

\author{%
  \name Animashree Anandkumar
  \email a.anandkumar@uci.edu \\
  \addr Electrical Engineering and Computer Science \\
  University of California, Irvine \\
  2200 Engineering Hall \\
  Irvine, CA 92697
  \AND
  \name Rong Ge
  \email rongge@microsoft.com \\
  \addr Microsoft Research \\
  One Memorial Drive \\
  Cambridge, MA 02142
  \AND
  \name Daniel Hsu
  \email djhsu@cs.columbia.edu \\
  \addr Department of Computer Science \\
  Columbia University \\
  1214 Amsterdam Avenue, \#0401 \\
  New York, NY 10027
  \AND
  \name Sham M. Kakade
  \email skakade@microsoft.com \\
  \addr Microsoft Research \\
  One Memorial Drive \\
  Cambridge, MA 02142
  \AND
  \name Matus Telgarsky
  \email mtelgars@cs.ucsd.edu \\
  \addr Department of Statistics \\
  Rutgers University \\
  110 Frelinghuysen Road \\
  Piscataway, NJ 08854
}
\editor{Benjamin Recht}

\maketitle

\begin{abstract}%
This work considers a computationally and statistically efficient
parameter estimation method for a wide class of latent variable
models---including Gaussian mixture models, hidden Markov models, and
latent Dirichlet allocation---which exploits a certain tensor structure in
their low-order observable moments (typically, of second- and third-order).
Specifically, parameter estimation is reduced to the problem of extracting
a certain (orthogonal) decomposition of a symmetric tensor derived from the
moments; this decomposition can be viewed as a natural generalization of
the singular value decomposition for matrices.
Although tensor decompositions are generally intractable to compute, the
decomposition of these specially structured tensors can be efficiently
obtained by a variety of approaches, including power iterations and
maximization approaches (similar to the case of matrices).
A detailed analysis of a robust tensor power method is provided,
establishing an analogue of Wedin's perturbation theorem for the singular
vectors of matrices.
This implies a robust and computationally tractable estimation approach for
several popular latent variable models.
\end{abstract}


\section{Introduction}

The method of moments is a classical parameter estimation
technique~\citep{Pearson94} from statistics which has proved invaluable in a
number of application domains.
The basic paradigm is simple and intuitive: (i) compute certain statistics
of the data---often empirical moments such as means and correlations---and (ii) find model parameters that give rise to (nearly) the same
corresponding population quantities.
In a number of cases, the method of moments leads to consistent estimators
which can be efficiently computed; this is especially relevant in the
context of latent variable models, where standard maximum likelihood
approaches are typically computationally prohibitive, and heuristic methods
can be unreliable and difficult to validate with high-dimensional data.
Furthermore, the method of moments can be viewed as complementary to the
maximum likelihood approach; simply taking a single step of Newton-Raphson
on the likelihood function starting from the moment based
estimator~\citep{LeCam86} often leads to the best of both worlds: a
computationally efficient estimator that is (asymptotically) statistically
optimal.

The primary difficulty in learning latent variable models is that the
latent (hidden) state of the data is not directly observed; rather
only observed variables correlated with the hidden state are observed.
As such, it is not evident the method of moments should fare any
better than maximum likelihood in terms of computational performance:
matching the model parameters to the observed moments may involve
solving computationally intractable systems of multivariate polynomial
equations.  Fortunately, for many classes of latent variable models,
there is rich structure in low-order moments (typically second- and
third-order) which allow for this inverse moment problem to be solved
efficiently~\citep{cattell1944parallel,
  cardoso1991super,Chang96,MR06,HKZ12,AHK12,SpectralLDA,HK13-mog}.  What
is more is that these decomposition problems are often amenable to
simple and efficient iterative methods, such as gradient descent and
the power iteration method.

\subsection{Contributions}

In this work, we
observe that a number of important and well-studied latent
variable models---including Gaussian mixture models, hidden Markov
models, and Latent Dirichlet allocation---share a certain structure in
their low-order moments, and this permits certain tensor decomposition
approaches to parameter estimation. In particular, this
decomposition can be viewed as a natural generalization of the
singular value decomposition for matrices.

While much of this (or similar) structure was implicit in several
previous works~\citep{Chang96,MR06,HKZ12,AHK12,SpectralLDA,HK13-mog}, here
we make the decomposition explicit under a unified framework.
Specifically, we express the observable moments as sums of rank-one
terms, and reduce the parameter estimation task to the problem of
extracting a symmetric orthogonal decomposition of a symmetric tensor
derived from these observable moments.  The problem can then be solved
by a variety of approaches, including fixed-point and variational
methods.

One approach for obtaining the orthogonal decomposition is the tensor power
method of~\citet[Remark 3]{SHOPM}.
We provide a convergence analysis of this method for orthogonally
decomposable symmetric tensors, as well as a detailed perturbation analysis
for a robust (and a computationally tractable) variant
(Theorem~\ref{thm:robustpower}).
This perturbation analysis can be viewed as an analogue of Wedin's
perturbation theorem for singular vectors of
matrices~\citep{wedin1972perturbation}, providing a bound on the error of
the recovered decomposition in terms of the operator norm of the tensor
perturbation.
This analysis is subtle in at least two ways.
First, unlike for matrices (where every matrix has a singular value
decomposition), an orthogonal decomposition need not exist for the
perturbed tensor.
Our robust variant uses random restarts and deflation to extract an
approximate decomposition in a computationally tractable manner.
Second, the analysis of the deflation steps is non-trivial; a na\"ive
argument would entail error accumulation in each deflation step, which we
show can in fact be avoided.
When this method is applied for parameter estimation in latent variable
models previously discussed, improved sample complexity bounds (over
previous work) can be obtained using this perturbation analysis.

Finally, we also address computational issues that arise when applying the
tensor decomposition approaches to estimating latent variable models.
Specifically, we show that the basic operations of simple iterative
approaches (such as the tensor power method) can be efficiently executed in
time linear in the dimension of the observations and the size of the
training data.
For instance, in a topic modeling application, the proposed methods require
time linear in the number of words in the vocabulary and in the number of
non-zero entries of the term-document matrix.
The combination of this computational efficiency and the robustness of the
tensor decomposition techniques makes the overall framework a promising
approach to parameter estimation for latent variable models.

\subsection{Related Work}

The connection between tensor decompositions and latent variable
models has a long history across many scientific and mathematical
disciplines.
We review some of the key works that are most closely related to ours.

\subsubsection{Tensor Decompositions}

The role of tensor decompositions in the context of latent variable models
dates back to early uses in psychometrics~\citep{cattell1944parallel}.
These ideas later gained popularity in chemometrics, and more recently in
numerous science and engineering disciplines, including neuroscience,
phylogenetics, signal processing, data mining, and computer vision.
A thorough survey of these techniques and applications is given by~\citet{kolda2009tensor}.
Below, we discuss a few specific connections to two applications in
machine learning and statistics, independent component analysis and latent
variable models (between which there is also significant overlap).

Tensor decompositions have been used in signal processing and computational
neuroscience for blind source separation and independent component analysis
(ICA)~\citep{Comon:book}.
Here, statistically independent non-Gaussian sources are linearly mixed in
the observed signal, and the goal is to recover the mixing matrix (and
ultimately, the original source signals).
A typical solution is to locate projections of the observed signals that
correspond to local extrema of the so-called ``contrast functions'' which
distinguish Gaussian variables from non-Gaussian variables.
This method can be effectively implemented using fast descent
algorithms~\citep{hyvarinen1999fast}.
When using the excess kurtosis (\emph{i.e.}, fourth-order cumulant) as the
contrast function, this method reduces to a generalization of the power
method for symmetric
tensors~\citep{SHOPM,ZG01,kofidis_regalia_power_convexity}.
This case is particularly important, since all local extrema of the
kurtosis objective correspond to the true sources (under the assumed
statistical model)~\citep{delfosse1995adaptive};
the descent methods can therefore be rigorously analyzed, and their
computational and statistical complexity can be
bounded~\citep{FriezeLinear,Nguyen-Regev,AroraICA}.

Higher-order tensor decompositions have also been used to develop
estimators for commonly used mixture models, hidden Markov models, and
other related latent variable models, often using the the algebraic
procedure of R.~Jennrich~\citep[as reported in the article
of][]{Harshman}, which is based on a simultaneous diagonalization of
different ways of flattening a tensor to matrices.
Jennrich's procedure
was employed for parameter estimation of discrete
Markov models by~\citet{Chang96} via pair-wise and triple-wise
probability tables; and it was later used for other latent variable
models such as hidden Markov models (HMMs), latent trees, Gaussian
mixture models, and topic models such as latent Dirichlet allocation
(LDA) by many others~\citep{MR06,HKZ12,AHK12,SpectralLDA,HK13-mog}.
In these contexts, it is often also possible to establish
strong identifiability results, without giving an explicit estimators,
by invoking the non-constructive identifiability argument
of~\citet{kruskal77}---see the article by~\citet{AMR09} for several
examples.

Related simultaneous diagonalization approaches have also been used
for blind source separation and ICA (as discussed above), and a number
of efficient algorithms have been developed for this
problem~\citep{bunse1993numerical,
CS93,PertDJ,CC96,corless1997reordered,ziehe2004fast}.
A rather different technique that uses tensor flattening and matrix
eigenvalue decomposition has been developed
by~\citet{cardoso1991super} and later by~\citet{de2007fourth}.
A significant advantage of this technique is that it can be used to
estimate overcomplete mixtures, where the number of sources is larger
than the observed dimension.

The relevance of tensor analysis to latent variable modeling has been long
recognized in the field of algebraic
statistics~\citep{pachter2005algebraic}, and many works characterize the
algebraic varieties corresponding to the moments of various classes of
latent variable models~\citep{drton2007algebraic,sturmfels2011binary}.
These works typically do not address computational or finite sample issues,
but rather are concerned with basic questions of identifiability.

The specific tensor structure considered in the present work is the
symmetric orthogonal decomposition.
This decomposition expresses a tensor as a linear combination of simple
tensor forms; each form is the tensor product of a vector (\emph{i.e.}, a
rank-$1$ tensor), and the collection of vectors form an orthonormal basis.
An important property of tensors with such decompositions is that they have
eigenvectors corresponding to these basis vectors.
Although the concepts of eigenvalues and eigenvectors of tensors is
generally significantly more complicated than their matrix counterpart---both
algebraically~\citep{qi2005eigenvalues,cartwright2011number,lim_tensoreig}
and
computationally~\citep{hillar2009most,kofidis_regalia_power_convexity}---the special symmetric orthogonal structure we consider permits simple
algorithms to efficiently and stably recover the desired decomposition.
In particular, a generalization of the matrix power method to symmetric
tensors, introduced by~\citet[Remark 3]{SHOPM} and analyzed
by~\citet{kofidis_regalia_power_convexity}, provides such a decomposition.
This is in fact implied by the characterization of~\cite{ZG01}, which shows
that iteratively obtaining the best rank-$1$ approximation of such
orthogonally decomposable tensors also yields the exact decomposition.
We note that in general, obtaining such approximations for general
(symmetric) tensors is NP-hard~\citep{hillar2009most}.

\subsubsection{Latent Variable Models}

This work focuses on the particular application of tensor decomposition
methods to estimating latent variable models, a significant
departure from many previous approaches in the machine learning and
statistics literature.
By far the most popular heuristic for parameter estimation for such models
is the Expectation-Maximization (EM) algorithm~\citep{DLR77,RW84}.
Although EM has a number of merits, it may suffer from slow convergence and
poor quality local optima~\citep{RW84}, requiring practitioners to
employ many additional heuristics to obtain good solutions.
For some models such as latent trees~\citep{Roch:CompBio} and topic
models~\citep{Arora:1439928}, maximum likelihood estimation is NP-hard,
which suggests that other estimation approaches may be more attractive.
More recently, algorithms from theoretical computer science and machine
learning have addressed computational and sample complexity issues related
to estimating certain latent variable models such as Gaussian mixture
models and
HMMs~\citep{Das99,AK01,DS07,VW02,KSV05,AM05,CR08,BV08,KMV10,BS10,MV10,HK13-mog,
Chang96,MR06,HKZ12,AHK12,Arora:1439928,SpectralLDA}.
See the works by \citet{AHK12} and \citet{HK13-mog} for a discussion of these methods, together with the
computational and statistical hardness barriers that they face.
The present work reviews a broad range of latent variables where a mild
non-degeneracy condition implies the symmetric orthogonal decomposition
structure in the tensors of low-order observable moments.

Notably, another class of methods, based on subspace
identification~\citep{SubIDMoor} and observable operator models/multiplicity
automata~\citep{S61,jaeger,LSS01}, have been proposed for a number of
latent variable models.
These methods were successfully developed for HMMs by~\citet{HKZ12}, and subsequently
generalized and extended for a number of related sequential and tree Markov
models models~\citep{SBG10,Bailly11,BSG10-psr,PSX11,FRU12,BQC12,BM12}, as well
as certain classes of parse tree models~\citep{LQBC12,CSCFU12,DRCFU12}.
These methods use low-order moments to learn an ``operator'' representation
of the distribution, which can be used for density estimation and belief
state updates.
While finite sample bounds can be given to establish the learnability of
these models~\citep{HKZ12}, the algorithms do not actually give parameter
estimates (\emph{e.g.}, of the emission or transition matrices in the case
of HMMs).

\subsection{Organization}

The rest of the paper is organized as
follows. Section~\ref{sec:prelims} reviews some basic definitions of
tensors. Section~\ref{sec:examples} provides examples of a number of
latent variable models which, after appropriate manipulations of their
low order moments, share a certain natural tensor
structure. Section~\ref{sec:decomps} reduces the problem of
parameter estimation to that of extracting a certain
(symmetric orthogonal) decomposition of a tensor. We then provide a
detailed analysis of a robust tensor power method and establish an
analogue of Wedin's perturbation theorem for the singular vectors of
matrices.  The discussion in Section~\ref{sec:discussion} addresses a
number of practical concerns that arise when dealing with moment
matrices and tensors.

\section{Preliminaries}\label{sec:prelims}

We introduce some tensor notations borrowed from~\citet{lim_tensoreig}.
A real \emph{$p$-th order tensor} $A \in \bigotimes_{i=1}^p \R^{n_i}$ is a
member of the tensor product of Euclidean spaces $\R^{n_i}$, $i \in [p]$.
We generally restrict to the case where $n_1 = n_2 = \dotsb = n_p = n$, and
simply write $A \in \bigotimes^p \R^n$.
For a vector $v \in \R^n$, we use $v^{\otimes p} := v \otimes v \otimes
\dotsb \otimes v \in \bigotimes^p \R^n$ to denote its $p$-th tensor power.
As is the case for vectors (where $p=1$) and matrices (where $p=2$), we may
identify a $p$-th order tensor with the $p$-way array of real numbers $[
A_{i_1,i_2,\dotsc,i_p} \colon i_1,i_2,\dotsc,i_p \in [n] ]$, where
$A_{i_1,i_2,\dotsc,i_p}$ is the $(i_1,i_2,\dotsc,i_p)$-th coordinate of $A$
(with respect to a canonical basis).

We can consider $A$ to be a multilinear map in the following sense: for a
set of matrices $\{ V_i \in \R^{n \times m_i} : i \in [p] \}$, the
$(i_1,i_2,\dotsc,i_p)$-th entry in the $p$-way array representation of
$A(V_1,V_2,\dotsc,V_p) \in \R^{m_1 \times m_2 \times \dotsb \times m_p}$ is
\begin{equation*}
[ A(V_1, V_2,\dotsc,V_p) ]_{i_1,i_2,\dotsc,i_p}
\ := \
\sum_{j_1,j_2,\dotsc,j_p \in [n]}
A_{j_1,j_2,\dotsc,j_p} \ 
[V_1]_{j_1, i_1} \ [V_2]_{j_2, i_2} \ \dotsb \ [V_p]_{j_p, i_p}
.
\end{equation*}
Note that if $A$ is a matrix ($p=2$), then
\[
A(V_1,V_2) =  V_1^\t A V_2
.
\]
Similarly, for a matrix $A$ and vector $v \in \R^n$, we can express $Av$ as
\[
A(I,v) = Av \in \R^n ,
\]
where $I$ is the $n \times n$ identity matrix.
As a final example of this notation, observe
\[
A(e_{i_1},e_{i_2},\dotsc,e_{i_p}) = A_{i_1,i_2,\dotsc,i_p}  ,
\]
where $\{ e_1, e_2, \dotsc, e_n \}$ is the canonical basis for $\R^n$.

Most tensors $A \in \bigotimes^p \R^n$ considered in this work will be
\emph{symmetric} (sometimes called \emph{supersymmetric}), which means that
their $p$-way array representations are invariant to permutations of the
array indices:
\emph{i.e.}, for all indices $i_1,i_2,\dotsc,i_p \in [n]$,
$A_{i_1,i_2,\dotsc,i_p} = A_{i_{\pi(1)},i_{\pi(2)},\dotsc,i_{\pi(p)}}$ for
any permutation $\pi$ on $[p]$.
It can be checked that this reduces to the usual definition of a symmetric
matrix for $p=2$.

The \emph{rank} of a $p$-th order tensor $A \in \bigotimes^p \R^n$ is the
smallest non-negative integer $k$ such that $A = \sum_{j=1}^k u_{1,j}
\otimes u_{2,j} \otimes \dotsb \otimes u_{p,j}$ for some $u_{i,j} \in \R^n,
i \in [p], j \in [k]$,
and the \emph{symmetric rank} of a symmetric $p$-th order tensor $A$ is the
smallest non-negative integer $k$ such that $A = \sum_{j=1}^k u_j^{\otimes
p}$ for some $u_j \in \R^n, j \in [k]$.\footnote{For even $p$, the definition is
slightly different~\citep{CGLM08}.}
The notion of rank readily reduces to the usual definition of matrix rank
when $p=2$, as revealed by the singular value decomposition.
Similarly, for symmetric matrices, the symmetric rank is equivalent to the
matrix rank as given by the spectral theorem.
A decomposition into such rank-one terms is known as a \emph{canonical
polyadic decomposition}~\citep{hitchcock1,hitchcock2}.

The notion of tensor (symmetric) rank is considerably more delicate than
matrix (symmetric) rank.
For instance, it is not clear \emph{a priori} that the symmetric rank of a
tensor should even be finite~\citep{CGLM08}.
In addition, removal of the best rank-$1$ approximation of a (general)
tensor may increase the tensor rank of the residual~\citep{tensor_rank_increase}.

Throughout, we use $\|v\| = (\sum_i v_i^2)^{1/2}$ to denote the Euclidean
norm of a vector $v$, and $\|M\|$ to denote the spectral (operator) norm of
a matrix.
We also use $\|T\|$ to denote the operator norm of a tensor, which we
define later.

\section{Tensor Structure in Latent Variable Models}\label{sec:examples}

In this section, we give several examples of latent variable models whose
low-order moments can be written as symmetric tensors of low symmetric
rank; some of these examples can be deduced using the techniques developed
in the text by~\citet{TMS}.
The basic form is demonstrated in Theorem~\ref{thm:single-topic} for the
first example, and the general pattern will emerge from subsequent
examples.

\subsection{Exchangeable Single Topic Models}\label{sec:singletopic}

We first consider a simple bag-of-words model for documents in which the
words in the document are assumed to be \emph{exchangeable}.
Recall that a collection of random variables $x_1, x_2, \dotsc, x_\ell$ are
exchangeable if their joint probability distribution is invariant to
permutation of the indices.
The well-known De Finetti's theorem~\citep{austin2008exchangeable} implies
that such exchangeable models can be viewed as mixture models in which
there is a latent variable $h$ such that $x_1, x_2, \dotsc, x_\ell$ are
conditionally i.i.d.~given $h$ (see Figure~\ref{fig:exchangeable} for the
corresponding graphical model) and the conditional distributions are
identical at all the nodes.

In our simplified topic model for documents, the latent variable $h$ is
interpreted as the (sole) topic of a given document, and it is assumed to
take only a finite number of distinct values.
Let $k$ be the number of distinct topics in the corpus, $d$ be the number
of distinct words in the vocabulary, and $\ell \geq 3$ be the number of
words in each document.
The generative process for a document is as follows: the document's
topic is drawn according to the discrete distribution specified by the
probability vector $w := (w_1,w_2,\dotsc,w_k) \in \Delta^{k-1}$.
This is modeled as a discrete random variable $h$ such that
\[ \Pr[h = j] = w_j , \quad j \in [k] . \]
Given the topic $h$, the document's $\ell$ words are drawn independently
according to the discrete distribution specified by the probability
vector $\mu_h \in \Delta^{d-1}$.
It will be convenient to represent the $\ell$ words in the document by
$d$-dimensional random \emph{vectors} $x_1, x_2, \dotsc, x_\ell \in \R^d$.
Specifically, we set
\[ x_t = e_i \quad \text{if and only if} \quad
\text{the $t$-th word in the document is $i$} , \quad t \in [\ell] , \]
where $e_1,e_2,\ldots e_d$ is the standard coordinate basis for $\R^d$.

One advantage of this encoding of words is that the (cross) moments of
these random vectors correspond to joint probabilities over words.
For instance, observe that
\begin{align*}
\E[ x_1 \otimes x_2 ]
& = \sum_{1 \leq i,j \leq d} \Pr[ x_1 = e_i, x_2 = e_j ] \ e_i \otimes e_j
\\
& = \sum_{1 \leq i,j \leq d} \Pr[ \text{$1$st word} = i, \text{$2$nd word} =
j ] \ e_i \otimes e_j ,
\end{align*}
so the $(i,j)$-the entry of the matrix $\E[ x_1 \otimes x_2 ]$ is $\Pr[
\text{$1$st word} = i, \text{$2$nd word} = j ]$.
More generally, the $(i_1,i_2,\dotsc,i_\ell)$-th entry in the tensor $\E[
x_1 \otimes x_2 \otimes \dotsb \otimes x_\ell ]$ is $\Pr[ \text{$1$st word}
= i_1, \text{$2$nd word} = i_2, \dotsc, \text{$\ell$-th word} = i_\ell ]$.
This means that estimating cross moments, say, of $x_1 \otimes x_2 \otimes
x_3$, is the same as estimating joint probabilities of the first three
words over all documents.
(Recall that we assume that each document has at least three words.)

The second advantage of the vector encoding of words is that the conditional
expectation of $x_t$ given $h = j$ is simply $\mu_j$, the vector of word
probabilities for topic $j$:
\[
\E[ x_t | h = j ]
\ = \ \sum_{i=1}^d \Pr[ \text{$t$-th word} = i | h = j] \ e_i
\ = \ \sum_{i=1}^d [ \mu_j ]_i \ e_i
\ = \ \mu_j
, \quad j \in [k]
\]
(where $[\mu_j]_i$ is the $i$-th entry in the vector $\mu_j$).
Because the words are conditionally independent given the topic, we can use
this same property with conditional cross moments, say, of $x_1$ and $x_2$:
\[
\E[ x_1 \otimes x_2 | h = j ]
\ = \ \E[ x_1 | h = j] \otimes \E[ x_2 | h = j]
\ = \ \mu_j \otimes \mu_j
, \quad j \in [k] .
\]
This and similar calculations lead one to the following theorem.

\begin{thm}[\citealp{AHK12}] \label{thm:single-topic}
If
\begin{eqnarray*}
M_2 & := & \E[ x_1 \otimes x_2 ] \\
M_3 & := & \E[ x_1 \otimes x_2 \otimes x_3 ] ,
\end{eqnarray*}
then
\begin{eqnarray*}
M_2 & = & \sum_{i=1}^k w_i \ \mu_i \otimes \mu_i \\
M_3 & = &
\sum_{i=1}^k w_i \ \mu_i \otimes \mu_i \otimes \mu_i .
\end{eqnarray*}
\end{thm}

As we will see in Section~\ref{sec:estimation}, the structure of $M_2$ and
$M_3$ revealed in Theorem~\ref{thm:single-topic} implies that the topic
vectors $\mu_1, \mu_2, \dotsc, \mu_k$ can be estimated by computing a
certain symmetric tensor decomposition.
Moreover, due to exchangeability, all triples (resp., pairs) of words in a
document---and not just the first three (resp., two) words---can be used in
forming $M_3$ (resp., $M_2$); see Section~\ref{sec:impl}.

\subsection{Beyond Raw Moments}
\label{sec:beyond}

In the single topic model above, the raw (cross) moments of the observed
words directly yield the desired symmetric tensor structure.
In some other models, the raw moments do not explicitly have this form.
Here, we show that the desired tensor structure can be found through
various manipulations of different moments.

\subsubsection{Spherical Gaussian Mixtures: Common Covariance}

We now consider a mixture of $k$ Gaussian distributions with spherical
covariances.
We start with the simpler case where all of the covariances are identical;
this probabilistic model is closely related to the (non-probabilistic)
$k$-means clustering problem~\citep{kmeans}.

Let $w_i \in (0,1)$ be the probability of choosing component $i \in [k]$,
$\{
\mu_1, \mu_2, \dotsc, \mu_k \} \subset \R^d$ be the component mean
vectors, and $\sigma^2 I$ be the common covariance matrix.  An
observation in this model is given by
\begin{align*}
x & := \mu_h + z ,
\end{align*}
where $h$ is the discrete random variable with $\Pr[h = i] = w_i$ for $i
\in [k]$ (similar to the exchangeable single topic model), and $z \sim
\N(0,\sigma^2 I)$ is an independent multivariate Gaussian random vector in
$\R^d$ with zero mean and spherical covariance $\sigma^2 I$.

The Gaussian mixture model differs from the exchangeable single topic model
in the way observations are generated.
In the single topic model, we observe multiple draws (words in a particular
document) $x_1, x_2, \dotsc, x_\ell$ given the same fixed $h$ (the topic of
the document).
In contrast, for the Gaussian mixture model, every realization of $x$
corresponds to a different realization of $h$.

\begin{thm}[\citealp{HK13-mog}] \label{thm:spherical_same}
Assume $d \geq k$.
The variance $\sigma^2$ is the smallest eigenvalue of the 
covariance matrix $\E[ x \otimes x ] - \E[x]\otimes \E[x]$.
Furthermore, if
\begin{eqnarray*}
M_2 & := & \E[ x \otimes x ] - \sigma^2 I \\
M_3 & := &
\E[ x \otimes x \otimes x] - \sigma^2 \sum_{i=1}^d \bigl(
\E[x] \otimes e_i \otimes e_i + e_i \otimes \E[x] \otimes e_i
+ e_i \otimes e_i \otimes \E[x] \bigr) ,
\end{eqnarray*}
then
\begin{eqnarray*}
M_2& = & \sum_{i=1}^k w_i \ \mu_i \otimes \mu_i \\
M_3 & = &
\sum_{i=1}^k w_i \ \mu_i \otimes \mu_i \otimes \mu_i .
\end{eqnarray*}
\end{thm}

\subsubsection{Spherical Gaussian Mixtures: Differing Covariances}

The general case is where each component may have a
\emph{different} spherical covariance.
An observation in this model is again $x = \mu_h + z$, but now $z \in \R^d$
is a random vector whose conditional distribution given $h = i$ (for some
$i \in [k]$) is a multivariate Gaussian $\N(0,\sigma_i^2 I)$ with zero mean
and spherical covariance $\sigma_i^2 I$.

\begin{thm}[\citealp{HK13-mog}] \label{thm:spherical}
Assume $d \geq k$.
The average variance $\bar\sigma^2 := \sum_{i=1}^k w_i \sigma_i^2$ is
the smallest eigenvalue of the covariance matrix $\E[ x \otimes x ] -
\E[x]\otimes \E[x]$.
Let $v$ be any unit norm eigenvector corresponding to the eigenvalue
$\bar\sigma^2$.
If
\begin{eqnarray*}
M_1 & := & \E[ x (v^\t (x - \E[x]))^2 ] \\
M_2 & := & \E[ x \otimes x ] - \bar\sigma^2 I \\
M_3 & := &
\E[ x \otimes x \otimes x] - \sum_{i=1}^d \bigl(
M_1 \otimes e_i \otimes e_i + e_i \otimes M_1 \otimes e_i
+ e_i \otimes e_i \otimes M_1 \bigr) ,
\end{eqnarray*}
then
\begin{eqnarray*}
M_2 & = & \sum_{i=1}^k w_i \ \mu_i \otimes \mu_i \\
M_3 & = &
\sum_{i=1}^k w_i \ \mu_i \otimes \mu_i \otimes \mu_i .
\end{eqnarray*}
\end{thm}
As shown by~\citet{HK13-mog}, $M_1 = \sum_{i=1}^k w_i \sigma_i^2 \mu_i$.
Note that for the common covariance case, where $\sigma_i^2 = \sigma^2$, we
have that $M_1 = \sigma^2 \E[x]$
(\emph{cf}.~Theorem~\ref{thm:spherical_same}).

\subsubsection{Independent Component Analysis (ICA)}

The standard model for ICA~\citep{Comon94,CC96,HO00,Comon:book}, in which
independent signals are linearly mixed and corrupted with Gaussian noise
before being observed, is specified as follows.
Let $h \in \R^k$ be a latent random \emph{vector} with independent
coordinates, $A \in \R^{d \times k}$ the mixing matrix, and $z$ be a
multivariate Gaussian random vector. The random vectors $h$ and $z$ are
assumed to be independent.
The observed random vector is
\begin{align*}
x & := A h + z .
\end{align*}
Let $\mu_i$ denote the $i$-th column of the mixing matrix $A$.

\begin{thm}[\citealp{Comon:book}]
Define
\begin{eqnarray*}
M_4 & := & \E[ x \otimes x \otimes x \otimes x ]
- T
\end{eqnarray*}
where $T$ is the fourth-order tensor with
\begin{multline*}
[T]_{i_1,i_2,i_3,i_4} := \E[ x_{i_1} x_{i_2} ] \E[ x_{i_3} x_{i_4} ] + \E[
x_{i_1} x_{i_3} ] \E[ x_{i_2} x_{i_4} ] \\
+ \E[ x_{i_1} x_{i_4} ] \E[ x_{i_2}
x_{i_3} ] ,
\quad 1 \leq i_1, i_2, i_3, i_4 \leq k 
\end{multline*}
(\emph{i.e.}, $T$ is the fourth derivative tensor of the function $v
\mapsto 8^{-1} \E[ (v^\t x)^2 ]^2$, so $M_4$ is the fourth cumulant tensor).
Let $\kappa_i := \E[h_i^4] - 3$ for each $i \in [k]$.
Then
\begin{eqnarray*}
M_4 & = &
\sum_{i=1}^k \kappa_i \ \mu_i \otimes \mu_i \otimes \mu_i \otimes \mu_i .
\end{eqnarray*}
\end{thm}
Note that $\kappa_i$ corresponds to the excess kurtosis, a measure of
non-Gaussianity as $\kappa_i = 0$ if $h_i$ is a standard normal random
variable.
Furthermore, note that $A$ is not identifiable if $h$ is a multivariate
Gaussian.

We may derive forms similar to that of $M_2$ and $M_3$ from
Theorem~\ref{thm:single-topic} using $M_4$ by observing that
\begin{align*}
M_4(I,I,u,v)
& = \sum_{i=1}^k \kappa_i (\mu_i^\t u) (\mu_i^\t v) \ \mu_i \otimes \mu_i
, \\
M_4(I,I,I,v)
& = \sum_{i=1}^k \kappa_i (\mu_i^\t v) \ \mu_i \otimes \mu_i \otimes \mu_i
\end{align*}
for any vectors $u,v \in \R^d$.

\subsubsection{Latent Dirichlet Allocation (LDA)}

An increasingly popular class of latent variable models are \emph{mixed
membership models}, where each datum may belong to several different latent
classes simultaneously.
LDA is one such model for the case of document modeling; here, each
document corresponds to a mixture over topics (as opposed to just a single
topic).
The distribution over such topic mixtures is a Dirichlet distribution
$\Dir(\alpha)$ with parameter vector $\alpha \in \R_{++}^k$ with strictly
positive entries; its density over the probability simplex $\Delta^{k-1} :=
\{ v \in \R^k : v_i \in [0,1] \forall i \in [k], \ \sum_{i=1}^k v_i = 1 \}$
is given by
\[
p_\alpha(h) =
\frac{\Gamma(\alpha_0)}{\prod_{i=1}^k \Gamma(\alpha_i)}
\prod_{i=1}^k h_i^{\alpha_i-1}
, \quad h \in \Delta^{k-1}
\]
where
\[ \alpha_0 := \alpha_1 + \alpha_2 + \dotsb + \alpha_k . \]

As before, the $k$ topics are specified by probability vectors $\mu_1,
\mu_2, \dotsc, \mu_k \in \Delta^{d-1}$.
To generate a document, we first draw the topic mixture $h =
(h_1,h_2,\dotsc,h_k) \sim \Dir(\alpha)$, and then conditioned on $h$, we
draw $\ell$ words $x_1,x_2,\dotsc,x_\ell$ independently from the discrete
distribution specified by the probability vector $\sum_{i=1}^k h_i
\mu_i$ (\emph{i.e.}, for each $x_t$, we independently sample a topic
$j$ according to $h$ and then sample $x_t$ according to $\mu_j$).
Again, we encode a word $x_t$ by setting $x_t = e_i$ iff the $t$-th word in
the document is $i$.

The parameter $\alpha_0$ (the sum of the ``pseudo-counts'') characterizes
the concentration of the distribution.
As $\alpha_0\rightarrow 0$, the distribution degenerates to a single  
topic model (\emph{i.e.}, the limiting density has, with probability
$1$, exactly one entry of $h$ being $1$ and the rest are $0$).
At the other extreme, if $\alpha = (c,c,\dotsc,c)$ for some scalar $c > 0$,
then as $\alpha_0 = ck \to \infty$, the distribution of $h$ becomes peaked
around the uniform vector $(1/k,1/k,\dotsc,1/k)$ (furthermore, the
distribution behaves like a product distribution).
We are typically interested in the case where $\alpha_0$ is small
(\emph{e.g.}, a constant independent of $k$), whereupon $h$ typically has
only a few large entries.
This corresponds to the setting where the documents are mainly comprised of
just a few topics.

\begin{thm}[\citealp{SpectralLDA}] \label{thm:lda}
Define
\begin{eqnarray*}
M_1 & := & \E[x_1] \\
M_2 & := &
\E[x_1 \otimes x_2 ] -
\frac{\alpha_0}{\alpha_0+1}
M_1\otimes M_1 \\
M_3 & : = &
\E[ x_1 \otimes x_2 \otimes x_3 ] \\
& & -\frac{\alpha_0}{\alpha_0+2}
\Bigl(\E[x_1 \otimes x_2 \otimes M_1]
+ \E[x_1 \otimes M_1 \otimes x_2]  + \E[M_1 \otimes x_1 \otimes x_2]
\Bigr) \\
& & + \frac{2\alpha_0^2}{(\alpha_0+2)(\alpha_0+1)}
M_1 \otimes M_1 \otimes M_1
.
\end{eqnarray*}
Then
\begin{eqnarray*}
M_2 & = & \sum_{i=1}^k \frac{\alpha_i}{(\alpha_0+1)\alpha_0} \ \mu_i \otimes \mu_i \\
M_3 & = &
\sum_{i=1}^k \frac{2\alpha_i}{(\alpha_0+2)(\alpha_0+1)\alpha_0} \ \mu_i
\otimes \mu_i \otimes \mu_i .
\end{eqnarray*}
\end{thm}

Note that $\alpha_0$ needs to be known to form $M_2$ and $M_3$ from
the raw moments.
This, however, is a much weaker than assuming that the entire distribution
of $h$ is known (\emph{i.e.}, knowledge of the whole parameter vector
$\alpha$).

\subsection{Multi-View Models}
\label{sec:multi}

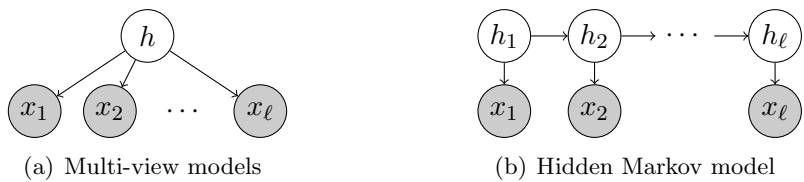
\begin{figure}
\begin{center}
\subfigure[Multi-view models]{\label{fig:exchangeable}
\begin{tikzpicture}
  [
    scale=1.0,
    observed/.style={circle,minimum size=0.7cm,inner sep=0mm,draw=black,fill=black!20},
    hidden/.style={circle,minimum size=0.7cm,inner sep=0mm,draw=black},
  ]
  \node [hidden,name=h] at ($(0,0)$) {$h$};
  \node [observed,name=x1] at ($(-1.5,-1)$) {$x_1$};
  \node [observed,name=x2] at ($(-0.5,-1)$) {$x_2$};
  \node at ($(0.5,-1)$) {$\dotsb$};
  \node [observed,name=xl] at ($(1.5,-1)$) {$x_\ell$};
  \draw [->] (h) to (x1);
  \draw [->] (h) to (x2);
  \draw [->] (h) to (xl);
\end{tikzpicture}}
\hfil\subfigure[Hidden Markov model]{\label{fig:hmm}
\begin{tikzpicture}
  [
    scale=1.0,
    observed/.style={circle,minimum size=0.7cm,inner sep=0mm,draw=black,fill=black!20},
    hidden/.style={circle,minimum size=0.7cm,inner sep=0mm,draw=black},
  ]
  \node [hidden,name=h1] at ($(-1.2,0)$) {$h_1$};
  \node [hidden,name=h2] at ($(0,0)$) {$h_2$};
  \node [name=hd] at ($(1.2,0)$) {$\dotsb$};
  \node [hidden,name=hl] at ($(2.4,0)$) {$h_\ell$};
  \node [observed,name=x1] at ($(-1.2,-1)$) {$x_1$};
  \node [observed,name=x2] at ($(0,-1)$) {$x_2$};
  \node [observed,name=xl] at ($(2.4,-1)$) {$x_\ell$};
  \draw [->] (h1) to (h2);
  \draw [->] (h2) to (hd);
  \draw [->] (hd) to (hl);
  \draw [->] (h1) to (x1);
  \draw [->] (h2) to (x2);
  \draw [->] (hl) to (xl);
\end{tikzpicture}}
\end{center}
\caption{Examples of latent variable models.
}
\label{fig:graphical-model}
\vspace{-1mm}
\end{figure}

Multi-view models (also sometimes called na\"ive Bayes models) are a
special class of Bayesian networks in which observed variables $x_1, x_2,
\ldots, x_\ell$ are conditionally independent given a latent variable $h$.
This is similar to the exchangeable single topic model, but here we do not
require the conditional distributions of the $x_t, t \in [\ell]$ to be
identical.
Techniques developed for this class can be used to handle a number of
widely used models including hidden Markov models \citep{MR06,AHK12},
phylogenetic tree models~\citep{Chang96,MR06}, certain tree mixtures~\citep{AnandkumarHsuKakade:graphmixturesNIPS12}, and certain probabilistic
grammar models~\citep{unmixing}.

As before, we let $h \in [k]$ be a discrete random variable with $\Pr[h =
j] = w_j$ for all $j \in [k]$.
Now consider random vectors $x_1 \in \R^{d_1}$, $x_2 \in \R^{d_2}$, and
$x_3 \in \R^{d_3}$ which are conditionally independent given $h$, and
\begin{align*}
\E[x_t | h = j] & = \mu_{t,j}
, \quad j \in [k], \ t \in \{1,2,3\}
\end{align*}
where the $\mu_{t,j} \in \R^{d_t}$ are the conditional means of the $x_t$
given $h = j$.
Thus, we allow the observations $x_1, x_2, \dotsc, x_\ell$ to be random vectors,
parameterized only by their conditional means. Importantly, these conditional
distributions may be discrete, continuous, or even a mix of both.

We first note the form for the raw (cross) moments.
\begin{prop} We have that:
\begin{eqnarray*} 
\E[ x_t \otimes x_{t'} ]
& = & \sum_{i=1}^k w_i \ \mu_{t,i} \otimes \mu_{t',i} ,
\quad \{t,t'\} \subset \{1,2,3\} , t \neq t' \\
\E[ x_1 \otimes x_2 \otimes x_3]
& = & \sum_{i=1}^k w_i \ \mu_{1,i} \otimes \mu_{2,i} \otimes \mu_{3,i} .
\end{eqnarray*}
\end{prop}

The cross moments do not possess a symmetric tensor form when the
conditional distributions are different.
Nevertheless, the moments can be ``symmetrized'' via a simple linear
transformation of $x_1$ and $x_2$ (roughly speaking, this relates $x_1$ and
$x_2$ to $x_3$); this leads to an expression from which the conditional
means of $x_3$ (\emph{i.e.}, $\mu_{3,1}, \mu_{3,2}, \dotsc, \mu_{3,k}$) can
be recovered.
For simplicity, we assume $d_1 = d_2 = d_3 = k$; the general case (with
$d_t \geq k$) is easily handled using low-rank singular value
decompositions.

\begin{thm}[\citealp{SpectralLDA}] \label{thm:multiview}
Assume that $\{ \mu_{v,1}, \mu_{v,2}, \dotsc, \mu_{v,k} \}$ are
linearly independent for each $v \in \{1,2,3\}$.
Define
\begin{eqnarray*}
\tl x_1 & := &  \E[x_3 \otimes x_2] \E[x_1 \otimes x_2]^{-1}  x_1 \\
\tl x_2 & := &  \E[x_3 \otimes x_1] \E[x_2 \otimes x_1]^{-1}  x_2 \\
M_2 & := & \E[\tl x_1 \otimes \tl x_2] \\
M_3 & := & \E[\tl x_1 \otimes \tl x_2 \otimes x_3] .
\end{eqnarray*}
Then
\begin{eqnarray*}
M_2
& = & \sum_{i=1}^k w_i \ \mu_{3,i} \otimes \mu_{3,i} \\
M_3
& = & \sum_{i=1}^k w_i \ \mu_{3,i} \otimes \mu_{3,i} \otimes \mu_{3,i} .
\end{eqnarray*}
\end{thm}

We now discuss three examples \citep[taken mostly from][]{AHK12} where the
above observations can be applied.
The first two concern mixtures of product distributions, and
the last one is the time-homogeneous hidden Markov model.

\subsubsection{Mixtures of Axis-Aligned Gaussians and Other Product
Distributions}

The first example is a mixture of $k$ product distributions in $\R^n$ under
a mild incoherence assumption~\citep{AHK12}.
Here, we allow each of the $k$ component distributions to have a different
product distribution (\emph{e.g.}, Gaussian distribution with an axis-aligned
covariance matrix), but require the matrix of component means $A := [ \mu_1
| \mu_2 | \dotsb | \mu_k ] \in \R^{n \times k}$ to satisfy a certain (very
mild) incoherence condition.
The role of the incoherence condition is explained below.

For a mixture of product distributions, any partitioning of the dimensions
$[n]$ into three groups creates three (possibly asymmetric) ``views'' which
are conditionally independent once the mixture component is selected.
However, recall that Theorem~\ref{thm:multiview} requires that for each
view, the $k$ conditional means be linearly independent.
In general, this may not be achievable; consider, for instance, the case
$\mu_i = e_i$ for each $i \in [k]$.
Such cases, where the component means are very aligned with the coordinate
basis, are precluded by the incoherence condition.

Define $\coherence(A) := \max_{i \in [n]} \{ e_i^\t \Pi_A e_i \}$ to be the
largest diagonal entry of the orthogonal projector to the range of $A$, and
assume $A$ has rank $k$.
The coherence lies between $k/n$ and $1$; it is largest when the range of
$A$ is spanned by the coordinate axes, and it is $k/n$ when the range is
spanned by a subset of the Hadamard basis of cardinality $k$.
The incoherence condition requires, for some $\veps, \delta \in (0,1)$,
$\coherence(A) \leq (\veps^2/6)/\ln(3k/\delta)$.
Essentially, this condition ensures that the non-degeneracy of the
component means is not isolated in just a few of the $n$ dimensions.
Operationally, it implies the following.
\begin{prop}[\citealp{AHK12}] \label{prop:product}
Assume $A$ has rank $k$, and
\[ \coherence(A) \leq \frac{\veps^2/6}
{\ln(3k/\delta)} \]
for some $\veps,\delta \in (0,1)$.
With probability at least $1-\delta$, a random partitioning of the
dimensions $[n]$ into three groups (for each $i \in [n]$, independently
pick $t \in \{1,2,3\}$ uniformly at random and put $i$ in group $t$)
has the following property.
For each $t \in \{1,2,3\}$ and $j \in [k]$, let $\mu_{t,j}$ be the entries
of $\mu_j$ put into group $t$, and let $A_t := [ \mu_{t,1} | \mu_{t,2} |
\dotsb | \mu_{t,k} ]$.
Then for each $t \in \{1,2,3\}$, $A_t$ has full column rank, and the $k$-th
largest singular value of $A_t$ is at least $\sqrt{(1-\veps)/3}$ times that
of $A$.
\end{prop}
Therefore, three asymmetric views can be created by randomly partitioning
the observed random vector $x$ into $x_1$, $x_2$, and $x_3$, such that the
resulting component means for each view satisfy the conditions of
Theorem~\ref{thm:multiview}.

\subsubsection{Spherical Gaussian Mixtures, Revisited}

Consider again the case of spherical Gaussian mixtures
(\emph{cf}.~Section~\ref{sec:beyond}).  As we shall see in
Section~\ref{sec:estimation}, the previous techniques (based on
Theorem~\ref{thm:spherical_same} and Theorem~\ref{thm:spherical}) lead
to estimation procedures when the dimension of $x$ is $k$ or greater (and
when the $k$ component means are linearly independent).  We now show
that when the dimension is slightly larger, say greater than $3k$, a
different (and simpler) technique based on the multi-view structure
can be used to extract the relevant structure.

We again use a randomized reduction.
Specifically, we create three views by (i) applying a random rotation to
$x$, and then (ii) partitioning $x \in \R^n$ into three views $\tl{x}_1,
\tl{x}_2, \tl{x}_3 \in \R^d$ for $d := n/3$.
By the rotational invariance of the multivariate Gaussian distribution, the
distribution of $x$ after random rotation is still a mixture of spherical
Gaussians (\emph{i.e.}, a mixture of product distributions), and thus
$\tl{x}_1, \tl{x}_2, \tl{x}_3$ are conditionally independent given $h$.
What remains to be checked is that, for each view $t \in \{1,2,3\}$, the
matrix of conditional means of $\tl{x}_t$ for each view has full column
rank.
This is true with probability $1$ as long as the matrix of conditional
means $A := [ \mu_1 | \mu_2 | \dotsb | \mu_k ] \in \R^{n \times k}$  has
rank $k$ and $n \geq 3k$.
To see this, observe that a random rotation in $\R^n$ followed by a
restriction to $d$ coordinates is simply a random projection from $\R^n$ to
$\R^d$, and that a random projection of a linear subspace of dimension $k$
to $\R^d$ is almost surely injective as long as $d \geq k$.
Applying this observation to the range of $A$ implies the following.
\begin{prop}[\citealp{HK13-mog}]
Assume $A$ has rank $k$ and that $n \geq 3k$.
Let $R \in \R^{n \times n}$ be chosen uniformly at random among all
orthogonal $n \times n$ matrices, and set $\tl{x} := Rx \in \R^n$ and
$\tl{A} := RA = [ R\mu_1 | R\mu_2 | \dotsb | R\mu_k ] \in \R^{n \times k}$.
Partition $[n]$ into three groups of sizes $d_1, d_2, d_3$ with $d_t \geq
k$ for each $t \in \{1,2,3\}$.
Furthermore, for each $t$, define $\tl{x}_t \in \R^{d_t}$ (respectively,
$\tl{A}_t \in \R^{d_t \times k}$) to be the subvector of $\tl{x}$ (resp.,
submatrix of $\tl{A}$) obtained by selecting the $d_t$ entries (resp.,
rows) in the $t$-th group.
Then $\tl{x}_1, \tl{x}_2, \tl{x}_3$ are conditionally independent given
$h$; $\E[\tl{x}_t | h = j] = \tl{A}_t e_j$ for each $j \in [k]$ and $t \in
\{1,2,3\}$; and with probability $1$, the matrices $\tl{A}_1, \tl{A}_2,
\tl{A}_3$ have full column rank.
\end{prop}

It is possible to obtain a quantitative bound on the $k$-th largest
singular value of each $A_t$ in terms of the $k$-th largest singular value
of $A$ (analogous to Proposition~\ref{prop:product}).
One avenue is to show that a random rotation in fact causes $\tl{A}$ to
have low coherence, after which we can apply Proposition~\ref{prop:product}.
With this approach, it is sufficient to require $n = O(k \log k)$ (for
constant $\veps$ and $\delta$), which results in the $k$-th largest
singular value of each $A_t$ being a constant fraction of the $k$-th
largest singular value of $A$.
We conjecture that, in fact, $n \geq c \cdot k$ for some $c > 3$ suffices.

\subsubsection{Hidden Markov Models}

Our last example is the time-homogeneous HMM for sequences of vector-valued
observations $x_1, x_2, \dotsc \in \R^d$.
Consider a Markov chain of discrete hidden states $y_1 \to y_2 \to y_3 \to
\dotsb$ over $k$ possible states $[k]$; given a state $y_t$ at time
$t$, the observation $x_t$ at time $t$ (a random vector taking values in
$\R^d$) is independent of all other observations and hidden states.
See Figure~\ref{fig:hmm}.

Let $\pi \in \Delta^{k-1}$ be the initial state distribution (\emph{i.e.},
the distribution of $y_1$), and $T \in \R^{k \times k}$ be the stochastic
transition matrix for the hidden state Markov chain: for all times $t$,
\[ \Pr[y_{t+1} = i | y_t = j] = T_{i,j} , \quad i,j \in [k] . \]
Finally, let $O \in \R^{d \times k}$ be the matrix whose $j$-th column is
the conditional expectation of $x_t$ given $y_t = j$: for all times $t$,
\[ \E[ x_t | y_t = j] = O e_j , \quad j \in [k] . \]

\begin{prop}[\citealp{AHK12}]
Define $h := y_2$, where $y_2$ is the second hidden state in the Markov
chain.
Then
\begin{itemize}
\item $x_1, x_2, x_3$ are conditionally independent given $h$;
\item the distribution of $h$ is given by the vector $w := T\pi \in
\Delta^{k-1}$;
\item for all $j \in [k]$,
\begin{align*}
\E[x_1 | h = j] & = O \diag(\pi) T^\t \diag(w)^{-1} e_j \\
\E[x_2 | h = j] & = O e_j \\
\E[x_3 | h = j] & = OT e_j .
\end{align*}
\end{itemize}
\end{prop}

Note the matrix of conditional means of $x_t$ has full column rank, for each $t
\in \{1,2,3\}$, provided that: (i) $O$ has full column rank, (ii) $T$ is
invertible, and (iii) $\pi$ and $T\pi$ have positive entries.

\section{Orthogonal Tensor Decompositions}\label{sec:decomps}

We now show how recovering the $\mu_i$'s in our aforementioned
problems reduces to the problem of finding a certain orthogonal
tensor decomposition of a symmetric tensor.
We start by reviewing the spectral decomposition of symmetric matrices, and
then discuss a generalization to the higher-order tensor case.
Finally, we show how orthogonal tensor decompositions can be used for
estimating the latent variable models from the previous section.

\subsection{Review: The Matrix Case}

We first build intuition by reviewing the matrix setting, where the
desired decomposition is the eigendecomposition of a symmetric rank-$k$
matrix $M = V \Lambda V^\t$, where $V = [v_1 | v_2 | \dotsb | v_k] \in
\R^{n \times k}$ is the matrix with orthonormal eigenvectors as columns,
and $\Lambda = \diag(\lambda_1,\lambda_2,\dotsc,\lambda_k) \in \R^{k \times
k}$ is diagonal matrix of non-zero eigenvalues.
In other words,
\begin{eqnarray}
M & = & \sum_{i=1}^k \lambda_i \ v_i v_i^\t  = \sum_{i=1}^k \lambda_i
\  v_i^{\otimes 2}.
\label{eq:spectral-decomp}
\end{eqnarray}
Such a decomposition is guaranteed to exist for every symmetric matrix.

Recovery of the $v_i$'s and $\lambda_i$'s can be viewed at least two ways.
First, each $v_i$ is fixed under the mapping $u \mapsto Mu$, up to a
scaling factor $\lambda_i$:
\[
Mv_i = \sum_{j=1}^k \lambda_j (v_j^\t v_i) v_j = \lambda_i v_i
\]
as $v_j^\t v_i = 0$ for all $j \neq i$ by orthogonality.
The $v_i$'s are not necessarily the only such fixed points.
For instance, with the multiplicity $\lambda_1 = \lambda_2 = \lambda$, then any linear combination
of $v_1$ and $v_2$ is similarly fixed under $M$.
However, in this case, the decomposition in~\eqref{eq:spectral-decomp} is
not unique, as $\lambda_1 v_1 v_1^\t + \lambda_2 v_2 v_2^\t$ is equal to $\lambda
(u_1 u_1^\t + u_2 u_2^\t)$ for any pair of orthonormal vectors, $u_1$ and
$u_2$ spanning the same subspace as $v_1$ and $v_2$.
Nevertheless, the decomposition is unique when $\lambda_1, \lambda_2,
\dotsc, \lambda_k$ are distinct, whereupon the $v_j$'s are the only
directions fixed under $u \mapsto Mu$ up to non-trivial scaling.

The second view of recovery is via the variational characterization of the
eigenvalues.
Assume $\lambda_1 > \lambda_2 > \dotsb > \lambda_k$; the case of repeated
eigenvalues again leads to similar non-uniqueness as discussed above.
Then the \emph{Rayleigh quotient}
\[ u \mapsto \frac{u^\t M u}{u^\t u} \]
is maximized over non-zero vectors by $v_1$.
Furthermore, for any $s \in [k]$, the maximizer of the Rayleigh quotient,
subject to being orthogonal to $v_1, v_2, \dotsc, v_{s-1}$, is $v_s$.
Another way of obtaining this second statement is to consider the
\emph{deflated} Rayleigh quotient
\[ u \mapsto \frac{u^\t \Bigl( M - \sum_{j=1}^{s-1} \lambda_j v_j v_j^\t
\Bigr) u}{u^\t u} \]
and observe that $v_s$ is the maximizer.

Efficient algorithms for finding these matrix decompositions are well
studied \citep[Section 8.2.3]{GVL96}, and iterative power methods are one
effective class of algorithms.

We remark that in our multilinear tensor notation, we may write the maps $u
\mapsto Mu$ and $u \mapsto u^\t M u / \|u\|_2^2$ as
\begin{align}
u \mapsto Mu
& \ \equiv \ u \mapsto M(I,u) ,
\label{eq:matrix-map}
\\
u \mapsto \frac{u^\t M u}{u^\t u}
& \ \equiv \ u \mapsto \frac{M(u,u)}{u^\t u}
.
\label{eq:matrix-rq}
\end{align}

\subsection{The Tensor Case}
\label{sec:tensor-decomp}

Decomposing general tensors is a delicate issue; tensors may not even have
unique decompositions.
Fortunately, the orthogonal tensors that arise in the aforementioned models
have a structure which permits a unique decomposition under a mild non-degeneracy condition.
We focus our attention to the case $p = 3$, \emph{i.e.}, a third order
tensor; the ideas extend to general $p$ with minor modifications.

An \emph{orthogonal decomposition} of a symmetric tensor $T \in
\bigotimes^3 \R^n$ is a collection of orthonormal (unit) vectors $\{ v_1,
v_2, \dotsc, v_k \}$
together with corresponding positive scalars $\lambda_i > 0$ such that
\begin{align}
T & = \sum_{i=1}^k \lambda_i v_i^{\otimes 3}
.
\label{eq:orthogonal-decomp}
\end{align}
Note that since we are focusing on odd-order tensors ($p = 3$), we have
added the requirement that the $\lambda_i$ be positive.
This convention can be followed without loss of generality since
$-\lambda_i v_i^{\otimes p} = \lambda_i (-v_i)^{\otimes p}$ whenever $p$ is
odd.
Also, it should be noted that orthogonal decompositions do not necessarily
exist for every symmetric tensor.

In analogy to the matrix setting, we consider two ways to view this
decomposition: a fixed-point characterization and a variational
characterization.
Related characterizations based on optimal rank-$1$ approximations are given
by~\citet{ZG01}.

\subsubsection{Fixed-Point Characterization}

For a tensor $T$, consider the vector-valued map
\begin{equation} \label{eq:tensor-map}
u \mapsto T(I,u,u)
\end{equation}
which is the third-order generalization of~\eqref{eq:matrix-map}.
This can be explicitly written as
\[
T(I,u,u) = \sum_{i=1}^d \sum_{1 \leq j,l \leq d} T_{i,j,l} (e_j^\t u) (e_l^\t u) e_i
.
\]
Observe that~\eqref{eq:tensor-map} is \emph{not} a linear map, which is a
key difference compared to the matrix case.

An eigenvector $u$ for a matrix $M$ satisfies $M(I,u) = \lambda u$, for
some scalar $\lambda$.
We say a unit vector $u \in \R^n$ is an \emph{eigenvector} of $T$, with
corresponding \emph{eigenvalue} $\lambda \in \R$, if
\[
T(I,u,u) = \lambda u .
\]
(To simplify the discussion, we assume throughout that eigenvectors have
unit norm; otherwise, for scaling reasons, we replace the above equation with $T(I,u,u) =
\lambda \|u\| u$.)
This concept was originally introduced
by~\citet{lim_tensoreig} and \citet{qi2005eigenvalues}.
For orthogonally decomposable tensors $T =
\sum_{i=1}^k \lambda_i v_i^{\otimes 3}$,
\[
T(I,u,u) = \sum_{i=1}^k \lambda_i (u^\t v_i)^2 v_i \ .
\]
By the orthogonality of the $v_i$, it is clear that $T(I,v_i,v_i) =
\lambda_i v_i$ for all $i \in [k]$.
Therefore each $(v_i,\lambda_i)$ is an eigenvector/eigenvalue pair.

There are a number of subtle differences compared to the matrix case that
arise as a result of the non-linearity of~\eqref{eq:tensor-map}.
First, even with the multiplicity $\lambda_1 = \lambda_2 = \lambda$, a
linear combination $u := c_1 v_1 +c_2 v_2$ may \emph{not} be an
eigenvector.
In particular,
\[ T(I,u,u)
= \lambda_1 c_1^2 v_1 +\lambda_2 c_2^2 v_2
= \lambda ( c_1^2 v_1 + c_2^2 v_2)
\]
may not be a multiple of $c_1 v_1 +c_2 v_2$.
This indicates that the issue of repeated eigenvalues does not have the
same status as in the matrix case.
Second, even if all the eigenvalues are distinct, it turns out that the $v_i$'s are not the only eigenvectors.
For example, set $u := (1/\lambda_1) v_1+(1/\lambda_2) v_2$.
Then,
\[ T(I,u,u)
= \lambda_1 (1/\lambda_1)^2 v_1 +\lambda_2 (1/\lambda_2)^2 v_2 = u
,
\]
so $u / \|u\|$ is an eigenvector.
More generally, for any subset $S \subseteq [k]$, the vector
\[ \sum_{i
\in S} \frac1{\lambda_i} \cdot v_i \]
is (proportional to) an eigenvector.

As we now see, these additional eigenvectors can be viewed as spurious.
We say a unit vector $u$ is a \emph{robust eigenvector} of $T$ if there
exists an $\epsilon > 0$ such that for all $\theta \in \{ u' \in \R^n :
\|u' - u\| \leq \epsilon \}$, repeated iteration of the map
\begin{equation} \label{eq:map2}
\bar \theta \mapsto \frac{
T(I,\bar\theta,\bar\theta)}{\|T(I,\bar\theta,\bar\theta)\|} \ ,
\end{equation}
starting from $\theta$ converges to $u$.
Note that the map~\eqref{eq:map2} rescales the output to have unit
Euclidean norm.
Robust eigenvectors are also called attracting fixed points
of~\eqref{eq:map2}~\citep[see, \emph{e.g.},][]{SIMAX-080148-Tensor-Eigenvalues}.

The following theorem implies that if $T$ has an orthogonal decomposition
as given in~\eqref{eq:orthogonal-decomp}, then the set of robust
eigenvectors of $T$ are precisely the set $\{v_1,v_2,\ldots v_k\}$,
implying that the orthogonal decomposition is unique.
(For even order tensors, the uniqueness is true up to sign-flips of the
$v_i$.)

\begin{thm}\label{thm:robust}
Let $T$ have an orthogonal decomposition as given in
\eqref{eq:orthogonal-decomp}.
\begin{enumerate}
\item The set of $\theta \in \R^n$ which do not converge to some $v_i$
under repeated iteration of \eqref{eq:map2} has measure zero.

\item The set of robust eigenvectors of $T$ is equal to $\{ v_1, v_2,
\dotsc, v_k \}$.

\end{enumerate}
\end{thm}
The proof of Theorem~\ref{thm:robust} is given in
Appendix~\ref{app:proof-robust}, and follows readily from simple
orthogonality considerations.
Note that \emph{every} $v_i$ in the orthogonal tensor decomposition is
robust, whereas for a symmetric matrix $M$, for almost all initial points,
the map $\bar\theta \mapsto \frac{M\bar\theta}{\|M\bar\theta\|}$ converges
only to an eigenvector corresponding to the largest magnitude eigenvalue.
Also, since the tensor order is odd, the signs of the robust eigenvectors
are fixed, as each $-v_i$ is mapped to $v_i$ under~\eqref{eq:map2}.

\subsubsection{Variational Characterization}

We now discuss a variational characterization of the orthogonal
decomposition.
The \emph{generalized Rayleigh quotient}~\citep{ZG01} for a third-order
tensor is
\[
u \mapsto \frac{T(u,u,u)}{(u^\t u)^{3/2}}
,
\]
which can be compared to~\eqref{eq:matrix-rq}.
For an orthogonally decomposable tensor, the following theorem shows that a
non-zero vector $u \in \R^n$ is an \emph{isolated local
maximizer}~\citep{NW99} of the generalized Rayleigh quotient if and only if
$u = v_i$ for some $i \in [k]$.

\begin{thm} \label{thm:variational}
Let $T$ have an orthogonal decomposition as given in
\eqref{eq:orthogonal-decomp}, and consider the optimization problem
\[ \max_{u \in \R^n} T(u,u,u) \ \text{s.t.} \ \|u\| \leq 1 . \]
\begin{enumerate}
\item The stationary points are eigenvectors of $T$.

\item A stationary point $u$ is an isolated local maximizer if and only if
$u = v_i$ for some $i \in [k]$.

\end{enumerate}
\end{thm}
The proof of Theorem~\ref{thm:variational} is given in
Appendix~\ref{app:proof-variational}.
It is similar to local optimality analysis for ICA methods using
fourth-order cumulants
\citep[\emph{e.g.},][]{delfosse1995adaptive,FriezeLinear}.

Again, we see similar distinctions to the matrix case.
In the matrix case, the only local maximizers of the Rayleigh quotient are
the eigenvectors with the largest eigenvalue (and these maximizers take on
the globally optimal value).
For the case of orthogonal tensor forms, the robust eigenvectors are
precisely the isolated local maximizers.

An important implication of the two characterizations is that, for
orthogonally decomposable tensors $T$, (i) the local maximizers of the
objective function $u \mapsto T(u,u,u) / (u^\t u)^{3/2}$ correspond
precisely to the vectors $v_i$ in the decomposition, and (ii) these local
maximizers can be reliably identified using a simple fixed-point iteration
(\emph{i.e.}, the tensor analogue of the matrix power method).
Moreover, a second-derivative test based on $T(I,I,u)$ can be employed to
test for local optimality and rule out other stationary points. 

\subsection{Estimation via Orthogonal Tensor Decompositions}
\label{sec:estimation}

We now demonstrate how the moment tensors obtained for various latent
variable models in Section~\ref{sec:examples} can be reduced to an
orthogonal form. 
For concreteness, we take the specific form from the exchangeable single
topic model (Theorem~\ref{thm:single-topic}):
\begin{eqnarray*}
M_2 & = & \sum_{i=1}^k w_i \ \mu_i \otimes \mu_i , \\
M_3 & = &
\sum_{i=1}^k w_i \ \mu_i \otimes \mu_i \otimes \mu_i .
\end{eqnarray*}
(The more general case allows the weights $w_i$ in $M_2$ to differ in
$M_3$, but for simplicity we keep them the same in the following
discussion.)  We now show how to reduce these forms to an orthogonally
decomposable tensor from which the $w_i$ and $\mu_i$ can be
recovered. See Appendix~\ref{app:simul} for a discussion as to how
previous approaches~\citep{MR06,AHK12,SpectralLDA,HK13-mog} achieved this
decomposition through a certain simultaneous diagonalization method.

Throughout, we assume the following non-degeneracy condition.
\begin{condition}[Non-degeneracy] \label{cond:rank}
The vectors $\mu_1, \mu_2, \dotsc, \mu_k \in \R^d$ are linearly
independent, and the scalars $w_1, w_2, \dotsc, w_k > 0$ are strictly
positive.
\end{condition}
Observe that Condition~\ref{cond:rank} implies that $M_2 \succeq 0$ is
positive semidefinite and has rank $k$.
This is often a mild condition in applications.
When this condition is not met, learning is conjectured to be generally hard
for both computational~\citep{MR06} and information-theoretic
reasons~\citep{MV10}.
As discussed by \citet{HKZ12} and \citet{HK13-mog}, when the non-degeneracy condition does not
hold, it is often possible to combine multiple observations using tensor
products to increase the rank of the relevant matrices.
Indeed, this observation has been rigorously formulated in very recent works
of \citet{smoothed} and \citet{merrier} using the framework of smoothed
analysis~\citep{ST09}.

\subsubsection{The Reduction}

First, let $W \in \R^{d \times k}$ be a linear transformation such that
\[
M_2(W,W) \ = \ W^\t M_2 W \ = \ I
\]
where $I$ is the $k \times k$ identity matrix (\emph{i.e.}, $W$
whitens $M_2$).
Since $M_2 \succeq 0$, we may for concreteness take $W := U D^{-1/2}$,
where $U \in \R^{d \times k}$ is the matrix of orthonormal eigenvectors of
$M_2$, and $D \in \R^{k \times k}$ is the diagonal matrix of positive
eigenvalues of $M_2$.
Let
\[ \tl\mu_i := \sqrt{w_i} \ W^\t \mu_i . \]
Observe that
\[ M_2(W,W)
\ = \ \sum_{i=1}^k W^\t (\sqrt{w_i} \mu_i) (\sqrt{w_i} \mu_i)^\t W
\ = \ \sum_{i=1}^k \tl\mu_i \tl\mu_i^\t
\ = \ I , \]
so the $\tl\mu_i \in \R^k$ are orthonormal vectors.

Now define $\wt{M}_3 := M_3(W,W,W) \in \R^{k \times k \times k}$, so that
\[
\wt{M}_3
\ = \ \sum_{i=1}^k w_i \ (W^\t \mu_i)^{\otimes 3}
\ = \ \sum_{i=1}^k \frac1{\sqrt{w_i}} \ \tl\mu_i^{\otimes 3}
.
\]
As the following theorem shows, the orthogonal decomposition of $\wt{M}_3$
can be obtained by identifying its robust eigenvectors, upon which the
original parameters $w_i$ and $\mu_i$ can be recovered.
For simplicity, we only state the result in terms of robust
eigenvector/eigenvalue pairs; one may also easily state everything in
variational form using Theorem~\ref{thm:variational}.

\begin{thm} \label{thm:main}
Assume Condition~\ref{cond:rank} and take $\wt{M}_3$ as defined above.
\begin{enumerate}
\item The set of robust eigenvectors of $\wt{M}_3$ is equal to $\{
\tl\mu_1, \tl\mu_2, \dotsc, \tl\mu_k \}$.

\item The eigenvalue corresponding to the robust eigenvector $\tl\mu_i$
of $\wt{M}_3$ is equal to $1/\sqrt{w_i}$, for all $i \in [k]$.

\item If $B \in \R^{d \times k}$ is the Moore-Penrose pseudoinverse of
$W^\t$, and $(v,\lambda)$ is a robust eigenvector/eigenvalue pair of
$\wt{M}_3$, then $\lambda Bv = \mu_i$ for some $i \in [k]$.

\end{enumerate}
\end{thm}
The theorem follows by combining the above discussion with the robust
eigenvector characterization of Theorem~\ref{thm:robust}.
Recall that we have taken as convention that eigenvectors have unit norm,
so the $\mu_i$ are exactly determined from the robust
eigenvector/eigenvalue pairs of $\wt{M}_3$ (together with the pseudoinverse
of $W^\t$); in particular, the scale of each $\mu_i$ is correctly
identified (along with the corresponding $w_i$).
Relative to previous works on moment-based estimators for latent variable
models \citep[\emph{e.g.},][]{AHK12,SpectralLDA,HK13-mog}, Theorem~\ref{thm:main}
emphasizes the role of the special tensor structure, which in turn makes
transparent the applicability of methods for orthogonal tensor
decomposition.

\subsubsection{Local Maximizers of (Cross Moment) Skewness}

The variational characterization provides an interesting perspective on the
robust eigenvectors for these latent variable models.
Consider the exchangeable single topic models
(Theorem~\ref{thm:single-topic}), and the objective function
\[
u \mapsto \frac
{\E[(x_1^\t u) (x_2^\t u) (x_3^\t u)]}
{\E[(x_1^\t u) (x_2^\t u)]^{3/2}}
= \frac{M_3(u,u,u)}{M_2(u,u)^{3/2}}
.
\]
In this case, every local maximizer $u^*$ satisfies $M_2(I,u^*) =
\sqrt{w_i} \mu_i$ for some $i \in [k]$.
The objective function can be interpreted as the (cross moment) skewness of
the random vectors $x_1, x_2, x_3$ along direction $u$.

\section{Tensor Power Method}

In this section, we consider the tensor power method of~\citet[Remark
3]{SHOPM} for orthogonal tensor decomposition.
We first state a simple convergence analysis for an orthogonally
decomposable tensor $T$.

When only an approximation $\hat{T}$ to an orthogonally decomposable
tensor $T$ is available (\emph{e.g.}, when empirical moments are used
to estimate population moments), an orthogonal decomposition need not
exist for this perturbed tensor (unlike for the case of matrices), and
a more robust approach is required to extract the approximate
decomposition.  Here, we propose such a variant in
Algorithm~\ref{alg:robustpower} and provide a detailed perturbation
analysis.
We note that alternative approaches such as simultaneous
diagonalization can also be employed (see Appendix~\ref{app:simul}).

\subsection{Convergence Analysis for Orthogonally Decomposable Tensors}
\label{sec:power}

The following lemma establishes the quadratic convergence of the tensor
power method---\emph{i.e.}, repeated iteration of~\eqref{eq:map2}---for
extracting a single component of the orthogonal decomposition.
Note that the initial vector $\theta_0$ determines which robust eigenvector
will be the convergent point.
Computation of subsequent eigenvectors can be computed with deflation,
\emph{i.e.}, by subtracting appropriate terms from $T$.

\begin{lem} \label{lemma:fixed-point}
Let $T \in \bigotimes^3 \R^n$ have an orthogonal decomposition as given in
\eqref{eq:orthogonal-decomp}.
For a vector $\theta_0 \in \R^n$, suppose that the set of numbers $|\lambda_1
v_1^\t\theta_0|, |\lambda_2 v_2^\t\theta_0|, \dotsc, |\lambda_k
v_k^\t\theta_0|$ has a unique largest element.
Without loss of generality, say $|\lambda_1 v_1^\t\theta_0|$ is this
largest value and $|\lambda_2 v_2^\t\theta_0|$ is the second largest value.
For $t = 1, 2, \dotsc$, let
\begin{equation*}
\theta_t
\ := \
\frac{T(I,\theta_{t-1},\theta_{t-1})}{\|T(I,\theta_{t-1},\theta_{t-1})\|} .
\end{equation*}
Then
\begin{equation*}
\|v_1 - \theta_t\|^2
\leq \biggl( 2 \lambda_1^2 \sum_{i=2}^k \lambda_i^{-2} \biggr)
\cdot \biggl|
\frac{\lambda_2 v_2^\t\theta_0}{\lambda_1 v_1^\t\theta_0} \biggr|^{2^{t+1}}
.
\end{equation*}
That is, repeated iteration of~\eqref{eq:map2} starting from $\theta_0$
converges to $v_1$ at a quadratic rate.
\end{lem}

To obtain all eigenvectors, we may simply proceed iteratively using
deflation, executing the power method on $T - \sum_j \lambda_j v_j^{\otimes
3}$ after having obtained robust eigenvector / eigenvalue pairs $\{
(v_j,\lambda_j) \}$.

\begin{proof}
Let $\overline\theta_0, \overline\theta_1, \overline\theta_2, \dotsc$ be the
sequence given by $\overline\theta_0 := \theta_0$ and $\overline\theta_t :=
T(I,\theta_{t-1},\theta_{t-1})$ for $t \geq 1$.
Let $c_i := v_i^\t\theta_0$ for all $i \in [k]$.
It is easy to check that (i) $\theta_t = \overline\theta_t /
\|\overline\theta_t\|$, and (ii)
$\overline\theta_t = \sum_{i=1}^k \lambda_i^{2^t-1} c_i^{2^t} v_i$.
(Indeed, $\overline\theta_{t+1}
= \sum_{i=1}^k \lambda_i (v_i^\t\overline\theta_t)^2 v_i
= \sum_{i=1}^k \lambda_i (\lambda_i^{2^t-1} c_i^{2^t})^2 v_i
= \sum_{i=1}^k \lambda_i^{2^{t+1}-1} c_i^{2^{t+1}} v_i$.)
Then
\begin{align*}
1 - (v_1^\t\theta_t)^2
& =
1 - \frac{(v_1^\t\overline\theta_t)^2}{\|\overline\theta_t\|^2}
= 1 - \frac{\lambda_1^{2^{t+1}-2} c_1^{2^{t+1}}}
{\sum_{i=1}^k \lambda_i^{2^{t+1}-2} c_i^{2^{t+1}}}
\leq \frac{\sum_{i=2}^k \lambda_i^{2^{t+1}-2} c_i^{2^{t+1}}}
{\sum_{i=1}^k \lambda_i^{2^{t+1}-2} c_i^{2^{t+1}}}
\\
& \leq \lambda_1^2 \sum_{i=2}^k \lambda_i^{-2} \cdot \biggl|
\frac{\lambda_2c_2}{\lambda_1c_1} \biggr|^{2^{t+1}}
.
\end{align*}
Since $\lambda_1 > 0$, we have $v_1^\t \theta_t > 0$ and hence $\|v_1 -
\theta_t\|^2 = 2(1 - v_1^\t \theta_t) \leq 2(1 - (v_1^\t \theta_t)^2)$ as
required.
\end{proof}

\subsection{Perturbation Analysis of a Robust Tensor Power Method}
\label{sec:perturbation}

Now we consider the case where we have an approximation $\hat{T}$ to an
orthogonally decomposable tensor $T$.
Here, a more robust approach is required to extract an approximate
decomposition.
We propose such an algorithm in Algorithm~\ref{alg:robustpower}, and
provide a detailed perturbation analysis.
For simplicity, we assume the tensor $\hat{T}$ is of size $k \times k
\times k$ as per the reduction from Section~\ref{sec:estimation}.
In some applications, it may be preferable to work directly with a $n
\times n \times n$ tensor of rank $k \leq n$ (as in
Lemma~\ref{lemma:fixed-point}); our results apply in that setting with
little modification.

\begin{algorithm}
\caption{Robust tensor power method}
\label{alg:robustpower}
\begin{algorithmic}[1]
\renewcommand{\algorithmicrequire}{\textbf{input}}
\renewcommand{\algorithmicensure}{\textbf{output}}
\REQUIRE symmetric tensor $\tilde{T} \in \R^{k \times k \times k}$, number of
iterations $L$, $N$.

\ENSURE the estimated eigenvector/eigenvalue pair; the deflated tensor.

\FOR{$\tau = 1$ to $L$}

\STATE Draw $\th{0}^{(\tau)}$ uniformly at random from the unit sphere in
$\R^k$.

\FOR{$t = 1$ to $N$}

\STATE Compute power iteration update
\begin{eqnarray}
\th{t}^{(\tau)} & := &
\frac{\tilde{T}(I, \th{t-1}^{(\tau)}, \th{t-1}^{(\tau)})}
{\|\tilde{T}(I, \th{t-1}^{(\tau)}, \th{t-1}^{(\tau)})\|}
\label{eq:power-update}
\end{eqnarray}

\ENDFOR

\ENDFOR

\STATE Let $\tau^* := \arg\max_{\tau \in [L]} \{ \tilde{T}(\th{N}^{(\tau)},
\th{N}^{(\tau)}, \th{N}^{(\tau)}) \}$.

\STATE Do $N$ power iteration updates \eqref{eq:power-update} starting from
$\th{N}^{(\tau^*)}$ to obtain $\hat\theta$, and set $\hat\lambda :=
\tilde{T}(\hat\theta,\hat\theta,\hat\theta)$.

\RETURN the estimated eigenvector/eigenvalue pair
$(\hat\theta,\hat\lambda)$; the deflated tensor $\tilde{T} - \hat\lambda \
\hat\theta^{\otimes 3}$.

\end{algorithmic}
\end{algorithm}

Assume that the symmetric tensor $T \in \R^{k \times k \times k}$ is
orthogonally decomposable, and that $\hat{T} = T + E$, where the
perturbation $E \in \R^{k \times k \times k}$ is a symmetric tensor with
small operator norm:
\[ 
\|E\| := \sup_{\|\theta\| = 1} |E(\theta,\theta,\theta)| . 
\]
In our latent variable model applications, $\hat{T}$ is the tensor formed
by using empirical moments, while $T$ is the orthogonally decomposable
tensor derived from the population moments for the given model.
In the context of parameter estimation (as in
Section~\ref{sec:estimation}), $E$ must account for any error amplification
throughout the reduction, such as in the whitening step~\citep[see,
\emph{e.g.},][for such an analysis]{HK13-mog}.

The following theorem is similar to Wedin's perturbation theorem for
singular vectors of matrices~\citep{wedin1972perturbation} in that it bounds
the error of the (approximate) decomposition returned by
Algorithm~\ref{alg:robustpower} on input $\hat{T}$ in terms of the size of
the perturbation, provided that the perturbation is small enough. 

\begin{thm}
\label{thm:robustpower}
Let $\hat{T} = T + E \in \R^{k \times k \times k}$, where $T$ is a
symmetric tensor with orthogonal decomposition $T = \sum_{i=1}^k \lambda_i
v_i^{\otimes 3}$ where each $\lambda_i > 0$, $\{ v_1, v_2, \dotsc, v_k \}$
is an orthonormal basis, and $E$ is a symmetric tensor with operator norm
$\|E\| \leq \eps$.
Define $\lambdamin := \min\{ \lambda_i : i \in [k] \}$, and $\lambdamax :=
\max\{ \lambda_i : i \in [k] \}$.
There exists universal constants $C_1, C_2, C_3 > 0$ such that the
following holds.
Pick any $\eta \in (0,1)$, and suppose
\[
\epsilon \leq C_1 \cdot \frac{\lambdamin}{k} ,
\qquad
N \geq C_2 \cdot \biggl( \log(k) + \log\log\Bigl(
\frac{\lambdamax}{\eps} \Bigr) \biggr)
,
\]
and
\begin{multline*}
\sqrt{\frac{\ln(L/\log_2(k/\eta))}{\ln(k)}}
\cdot \Biggl( 1 - \frac{\ln(\ln(L/\log_2(k/\eta))) +
C_3}{4\ln(L/\log_2(k/\eta))} -
\sqrt{\frac{\ln(8)}{\ln(L/\log_2(k/\eta))}} \Biggr)
\\
\geq 1.02 \Biggl( 1 + \sqrt{\frac{\ln(4)}{\ln(k)}}
\Biggr)
.
\end{multline*}
(Note that the condition on $L$ holds with $L = \poly(k) \log(1/\eta)$.)
Suppose that Algorithm~\ref{alg:robustpower} is iteratively called $k$
times, where the input tensor is $\hat{T}$ in the first call, and in each
subsequent call, the input tensor is the deflated tensor returned by the
previous call.
Let $(\hat{v}_1,\hat\lambda_1), (\hat{v}_2,\hat\lambda_2), \dotsc,
(\hat{v}_k,\hat\lambda_k)$ be the sequence of estimated
eigenvector/eigenvalue pairs returned in these $k$ calls.
With probability at least $1-\eta$, there exists a permutation $\pi$ on
$[k]$ such that
\[
\|v_{\pi(j)}-\hat{v}_j\| \leq 8 \epsilon/\lambda_{\pi(j)}
, \qquad
|\lambda_{\pi(j)}-\hat\lambda_j| \leq 5\epsilon , \quad \forall j \in [k]
,
\]
and
\[
\biggl\|
T - \sum_{j=1}^k \hat\lambda_j \hat{v}_j^{\otimes 3} 
\biggr\| \leq 55\eps .
\]
\end{thm}
The proof of Theorem~\ref{thm:robustpower} is given in
Appendix~\ref{app:analysis}.

One important difference from Wedin's theorem is that this is an
algorithm dependent perturbation analysis, specific to
Algorithm~\ref{alg:robustpower} (since the perturbed tensor need not
have an orthogonal decomposition).
Furthermore, note that Algorithm~\ref{alg:robustpower} uses multiple
restarts to ensure (approximate) convergence---the intuition is that by
restarting at multiple points, we eventually start at a point in which the
initial contraction towards some eigenvector dominates the error $E$ in our
tensor.
The proof shows that we find such a point with high probability within $L =
\poly(k)$ trials.
It should be noted that for large $k$, the required bound on $L$ is very
close to linear in $k$.

We note that it is also possible to design a variant of
Algorithm~\ref{alg:robustpower} that instead uses a stopping criterion to
determine if an iterate has (almost) converged to an eigenvector.
For instance, if $\tilde{T}(\theta,\theta,\theta) > \max \{ \|\tilde{T}\|_F
/ \sqrt{2r} , \|\tilde{T}(I,I,\theta)\|_F / 1.05 \}$, where
$\|\tilde{T}\|_F$ is the tensor Frobenius norm (vectorized Euclidean norm),
and $r$ is the expected rank of the unperturbed tensor ($r = k - \text{\#
of deflation steps}$), then it can be shown that $\theta$ must be close to
one of the eigenvectors, provided that the perturbation is small enough.
Using such a stopping criterion can reduce the number of random restarts
when a good initial point is found early on.
See Appendix~\ref{app:stopping} for details.

In general, it is possible, when run on a general symmetric tensor
(\emph{e.g.}, $\hat T$), for the tensor power method to exhibit
oscillatory behavior \citep[Example 1]{kofidis_regalia_power_convexity}.
This is not in conflict with Theorem~\ref{thm:robustpower}, which effectively bounds the
amplitude of these oscillations; in particular, if $\hat T= T+E$
is a tensor built from empirical
moments, the error term $E$ (and thus the amplitude of the oscillations) can be driven down
by drawing more samples.
The practical value of addressing these oscillations
and perhaps stabilizing the algorithm is an interesting direction for future
research~\citep{SIMAX-080148-Tensor-Eigenvalues}.

A final consideration is that for specific applications, it may be possible
to use domain knowledge to choose better initialization points.
For instance, in the topic modeling applications
(\emph{cf}.~Section~\ref{sec:singletopic}), the eigenvectors are related to
the topic word distributions, and many documents may be primarily composed
of words from just single topic.
Therefore, good initialization points can be derived from these
single-topic documents themselves, as these points would already be close
to one of the eigenvectors.

\section{Discussion}
\label{sec:discussion}

In this section, we discuss some practical and application-oriented issues
related to the tensor decomposition approach to learning latent variable
models.

\subsection{Practical Implementation Considerations}
\label{sec:impl}

A number of practical concerns arise when dealing with moment matrices and
tensors.
Below, we address two issues that are especially pertinent to topic
modeling applications~\citep{AHK12,SpectralLDA} or other settings where the
observations are sparse.

\subsubsection{Efficient Moment Representation for Exchangeable Models}

In an exchangeable bag-of-words model, it is assumed that the words $x_1,
x_2, \dotsc, x_\ell$ in a document are conditionally i.i.d.~given the topic
$h$.
This allows one to estimate $p$-th order moments using just $p$ words per
document.
The estimators obtained via Theorem~\ref{thm:single-topic} (single topic
model) and Theorem~\ref{thm:lda} (LDA) use only up to third-order moments,
which suggests that each document only needs to have three words.

In practice, one should use all of the words in a document for
efficient estimation of the moments.
One way to do this is to average over all $\binom{\ell}{3} \cdot 3!$
ordered triples of words in a document of length $\ell$.
At first blush, this seems computationally expensive (when $\ell$ is
large), but as it turns out, the averaging can be done implicitly, as shown
by~\citet{ZHPA13-contrast}.
Let $c \in \R^d$ be the word count vector for a document of length $\ell$,
so $c_i$ is the number of occurrences of word $i$ in the document, and
$\sum_{i=1}^d c_i = \ell$.
Note that $c$ is a sufficient statistic for the document.
Then, the contribution of this document to the empirical third-order moment
tensor is given by
\begin{multline}
\frac{1}{\binom{\ell}{3}} \cdot \frac{1}{3!} \cdot \biggl(
c \otimes c \otimes c
+ 2\sum_{i=1}^d c_i \ (e_i \otimes e_i \otimes e_i)
\\
- \sum_{i=1}^d \sum_{j=1}^d c_i c_j \ (e_i \otimes e_i \otimes e_j)
- \sum_{i=1}^d \sum_{j=1}^d c_i c_j \ (e_i \otimes e_j \otimes e_i)
- \sum_{i=1}^d \sum_{j=1}^d c_i c_j \ (e_i \otimes e_j \otimes e_j)
\biggr)
.\label{m3wordcount}
\end{multline}
It can be checked that this quantity is equal to
\[
\frac{1}{\binom{\ell}{3}} \cdot \frac{1}{3!} \cdot
\sum_{\text{ordered word triple $(x,y,z)$}}
e_x \otimes e_y \otimes e_z
\]
where the sum is over all ordered word triples in the document.
A similar expression is easily derived for the contribution of the document
to the empirical second-order moment matrix:
\begin{equation}\label{eqn:m2wordcount}
\frac{1}{\binom{\ell}{2}} \cdot \frac{1}{2!} \cdot \biggl(
c \otimes c - \diag(c) \biggr) .
\end{equation}
Note that the word count vector $c$ is generally a sparse vector, so this
representation allows for efficient multiplication by the moment matrices
and tensors in time linear in the size of the document corpus (\emph{i.e.},
the number of non-zero entries in the term-document matrix).

\subsubsection{Dimensionality Reduction}

Another serious concern regarding the use of tensor forms of moments is the
need to operate on multidimensional arrays with $\Omega(d^3)$ values (it is
typically not exactly $d^3$ due to symmetry).
When $d$ is large (\emph{e.g.}, when it is the size of the vocabulary in
natural language applications), even storing a third-order tensor in memory
can be prohibitive.
Sparsity is one factor that alleviates this problem.
Another approach is to use efficient linear dimensionality reduction.
When this is combined with efficient techniques for matrix and tensor
multiplication that avoid explicitly constructing the moment matrices and
tensors (such as the procedure described above), it is possible to avoid
any computational scaling more than linear in the dimension $d$ and the
training sample size.

Consider for concreteness the tensor decomposition approach for the
exchangeable single topic model as discussed in
Section~\ref{sec:estimation}.
Using recent techniques for randomized linear algebra computations
\citep[\emph{e.g.},][]{tropp:svd}, it is possible to efficiently approximate
the whitening matrix $W \in \R^{d \times k}$ from the second-moment matrix
$M_2 \in \R^{d \times d}$.
To do this, one first multiplies $M_2$ by a random matrix $R \in \R^{d
\times k'}$ for some $k' \geq k$, and then computes the top $k$ singular
vectors of the product $M_2R$.
This provides a basis $U \in \R^{d \times k}$ whose span is approximately
the range of $M_2$.
From here, an approximate SVD of $U^\t M_2 U$ is used to compute the
approximate whitening matrix $W$.
Note that both matrix products $M_2R$ and $U^\t M_2 U$ may be performed via
implicit access to $M_2$ by exploiting \eqref{eqn:m2wordcount}, so that
$M_2$ need not be explicitly formed.
With the whitening matrix $W$ in hand, the third-moment tensor $\wt{M}_3 =
M_3(W,W,W) \in \R^{k \times k \times k}$ can be implicitly computed via
\eqref{m3wordcount}.
For instance, the core computation in the tensor power method $\theta' :=
\wt{M}_3(I,\theta,\theta)$ is performed by (i) computing $\eta := W\theta$,
(ii) computing $\eta' := M_3(I,\eta,\eta)$, and finally (iii) computing
$\theta' := W^\t \eta'$.
Using the fact that $M_3$ is an empirical third-order moment tensor, these
steps can be computed with $O(dk + N)$ operations, where $N$ is the number
of non-zero entries in the term-document matrix~\citep{ZHPA13-contrast}.

\subsection{Computational Complexity}

It is interesting to consider the computational complexity of the tensor
power method in the dense setting where $T \in \R^{k \times k \times k}$ is
orthogonally decomposable but otherwise unstructured.
Each iteration requires $O(k^3)$ operations, and assuming at most
$k^{1+\delta}$ random restarts for extracting each eigenvector (for some
small $\delta > 0$) and $O(\log(k) + \log\log(1/\epsilon))$ iterations per
restart, the total running time is $O(k^{5+\delta} (\log(k) +
\log\log(1/\epsilon)))$ to extract all $k$ eigenvectors and eigenvalues.

An alternative approach to extracting the orthogonal decomposition of $T$
is to reorganize $T$ into a matrix $M \in \R^{k \times k^2}$ by flattening
two of the dimensions into one.
In this case, if $T = \sum_{i=1}^k \lambda_i v_i^{\otimes 3}$, then $M =
\sum_{i=1}^k \lambda_i v_i \otimes \vectorize(v_i \otimes v_i)$.
This reveals the singular value decomposition of $M$ (assuming the
eigenvalues $\lambda_1, \lambda_2, \dotsc, \lambda_k$ are distinct), and
therefore can be computed with $O(k^4)$ operations.
Therefore it seems that the tensor power method is less efficient than a
pure matrix-based approach via singular value decomposition.
However, it should be noted that this matrix-based approach fails to
recover the decomposition when eigenvalues are repeated, and can be
unstable when the gap between eigenvalues is small---see
Appendix~\ref{app:simul} for more discussion.

It is worth noting that the running times differ by roughly a factor of
$\Theta(k^{1+\delta})$, which can be accounted for by the random restarts.
This gap can potentially be alleviated or removed by using a more clever
method for initialization.
Moreover, using special structure in the problem (as discussed above) can
also improve the running time of the tensor power method.

\subsection{Sample Complexity Bounds}

Previous work on using linear algebraic methods for estimating latent
variable models crucially rely on matrix perturbation analysis for deriving
sample complexity bounds~\citep{MR06,HKZ12,AHK12,SpectralLDA,HK13-mog}.
The learning algorithms in these works are plug-in estimators that use
empirical moments in place of the population moments, and then follow
algebraic manipulations that result in the desired parameter estimates.
As long as these manipulations can tolerate small perturbations of the
population moments, a sample complexity bound can be obtained by exploiting
the convergence of the empirical moments to the population moments via the
law of large numbers.
As discussed in Appendix~\ref{app:simul}, these approaches do not directly
lead to practical algorithms due to a certain amplification of the error (a
polynomial factor of $k$, which is observed in practice).

Using the perturbation analysis for the tensor power method, improved sample
complexity bounds can be obtained for all of the examples discussed in
Section~\ref{sec:examples}.
The underlying analysis remains the same as in previous works
\citep[\emph{e.g.},][]{SpectralLDA,HK13-mog}, the main difference being the accuracy of the
orthogonal tensor decomposition obtained via the tensor power method.
Relative to the previously cited works, the sample complexity bound will be
considerably improved in its dependence on the rank parameter $k$, as
Theorem~\ref{thm:robustpower} implies that the tensor estimation
error (\emph{e.g.}, error in estimating $\wt{M}_3$ from
Section~\ref{sec:estimation}) is not amplified by any factor explicitly
depending on $k$ (there is a requirement that the error be smaller than
some factor depending on $k$, but this only contributes to a lower-order
term in the sample complexity bound).
See Appendix~\ref{app:simul} for further discussion regarding the stability
of the techniques from these previous works.

\subsection{Other Perspectives}

The tensor power method is simply one approach for extracting the
orthogonal decomposition needed in parameter estimation.
The characterizations from Section~\ref{sec:tensor-decomp} suggest that a
number of fixed point and variational techniques may be possible (and
Appendix~\ref{app:simul} provides yet another perspective based on simultaneous diagonalization).
One important consideration is that the model is often misspecified, and
therefore approaches with more robust guarantees (\emph{e.g.}, for
convergence) are desirable.
Our own experience with the tensor power method (as applied to exchangeable
topic modeling) is that while model misspecification does indeed affect
convergence, the results can be very reasonable even after just a dozen or
so iterations~\citep{SpectralLDA}.
Nevertheless, robustness is likely more important in other applications,
and thus the stabilization approaches
\citep{kofidis_regalia_power_convexity,RK03,Erdogan09,SIMAX-080148-Tensor-Eigenvalues}
may be advantageous.

\acks{%
  We thank Boaz Barak, Dean Foster, Jon Kelner, and Greg Valiant for helpful
  discussions.
  We are also grateful to Hanzhang Hu, Drew Bagnell, and Martial Hebert
  for alerting us of an issue with Theorem~\ref{thm:variational} and
  suggesting a simple fix.
  This work was completed while DH was a postdoctoral researcher at Microsoft
  Research New England, and partly while AA, RG, and MT were visiting the
  same lab.
  AA is supported in part by the NSF Award CCF-1219234, AFOSR Award
  FA9550-10-1-0310 and the ARO Award W911NF-12-1-0404.
}

\appendix

\section{Fixed-Point and Variational Characterizations of Orthogonal Tensor
Decompositions}

We give detailed proofs of Theorems~\ref{thm:robust} and~\ref{thm:variational}
in this section for completeness.

\subsection{Proof of Theorem~\ref{thm:robust}}
\label{app:proof-robust}

\begin{thm2}{thm:robust}
Let $T$ have an orthogonal decomposition as given in
\eqref{eq:orthogonal-decomp}.
\begin{enumerate}
\item The set of $\theta \in \R^n$ which do not converge to some $v_i$
under repeated iteration of \eqref{eq:map2} has measure zero.

\item The set of robust eigenvectors of $T$ is $\{ v_1, v_2, \dotsc, v_k
\}$.

\end{enumerate}
\end{thm2}
\begin{proof}
\sloppy
For a random choice of $\theta \in \R^n$ (under any distribution absolutely
continuous with respect to Lebesgue measure), the values $|\lambda_1
v_1^\t\theta|, |\lambda_2 v_2^\t\theta|, \dotsc, |\lambda_k v_k^\t\theta|$
will be distinct with probability $1$.
Therefore, there exists a unique largest value, say $|\lambda_i
v_i^\t\theta|$ for some $i \in [k]$, and by Lemma~\ref{lemma:fixed-point},
we have convergence to $v_i$ under repeated iteration of~\eqref{eq:map2}.
Thus the first claim holds.

\fussy
We now prove the second claim.
First, we show that every $v_i$ is a robust eigenvector.
Pick any $i \in [k]$, and note that for a sufficiently small ball around
$v_i$, we have that for all $\theta$ in this ball, $\lambda_i v_i^\t\theta$
is strictly greater than $\lambda_j v_j^\t\theta$ for $j \in [k] \setminus
\{i\}$.
Thus by Lemma~\ref{lemma:fixed-point}, $v_i$ is a robust eigenvector.
Now we show that the $v_i$ are the only robust eigenvectors.
Suppose there exists some robust eigenvector $u$ not equal to $v_i$ for any
$i \in [k]$.
Then there exists a positive measure set around $u$ such that all points in
this set converge to $u$ under repeated iteration of \eqref{eq:map2}.
This contradicts the first claim.
\end{proof}

\subsection{Proof of Theorem~\ref{thm:variational}}
\label{app:proof-variational}

\begin{thm2}{thm:variational}
Let $T$ have an orthogonal decomposition as given in
\eqref{eq:orthogonal-decomp}, and consider the optimization problem
\[ \max_{u \in \R^n} T(u,u,u) \ \text{s.t.} \ \|u\| \leq 1 . \]
\begin{enumerate}
\item The stationary points are eigenvectors of $T$.

\item A stationary point $u$ is an isolated local maximizer if and only if
$u = v_i$ for some $i \in [k]$.

\end{enumerate}
\end{thm2}
\begin{proof}
Consider the Lagrangian form of the corresponding constrained maximization
problem over unit vectors $u \in \R^n$:
\[ \calL(u,\lambda) := T(u,u,u) - \frac32 \lambda (u^\t u - 1)
.
\]
Since
\[ \nabla_u \calL(u,\lambda)
= \nabla_u \biggl( \sum_{i=1}^k \lambda_i (v_i^\t u)^3 - \frac32 \lambda
(u^\t u - 1) \biggr)
= 3 \Bigl( T(I,u,u) - \lambda u \Bigr)
,
\]
the stationary points $u \in \R^n$ (with $\|u\| \leq 1$) satisfy
\[ T(I,u,u) = \lambda u \]
for some $\lambda \in \R$, \emph{i.e.}, $(u,\lambda)$ is an
eigenvector/eigenvalue pair of $T$.

Now we characterize the isolated local maximizers.
Observe that if $u\neq0$ and $T(I,u,u) = \lambda u$ for $\lambda < 0$,
then $T(u,u,u) < 0$.
Therefore $u' = (1-\delta)u$ for any $\delta \in (0,1)$
satisfies $T(u',u',u') = (1-\delta)^3 T(u,u,u) > T(u,u,u)$.
So such a $u$ cannot be a local maximizer.
Moreover, if $\|u\| < 1$ and $T(I,u,u) = \lambda u$ for $\lambda > 0$,
then $u' = (1+\delta)u$ for a small enough $\delta \in (0,1)$
satisfies $\|u'\| \leq 1$ and $T(u',u',u') = (1+\delta)^3 T(u,u,u) >
T(u,u,u)$.
Therefore a local maximizer must have $T(I,u,u) = \lambda u$ for some
$\lambda \geq 0$, and $\|u\| = 1$ whenever $\lambda>0$.

Extend $\{ v_1, v_2, \dotsc, v_k \}$ to an orthonormal basis $\{ v_1, v_2,
\dotsc, v_n \}$ of $\R^n$.
Now pick any stationary point $u = \sum_{i=1}^n c_i v_i$ with $\lambda :=
T(u,u,u) = u^\t T(I,u,u)$.
Then
\[
\lambda_i c_i^2
= \lambda_i (u^\t v_i)^2
= v_i^\t T(I,u,u)
= \lambda v_i^\t u
=
\lambda c_i
\geq 0
, \quad i \in [k]
,
\]
and thus
\[
\nabla_u^2 \calL(u,\lambda)
= 6 \sum_{i=1}^k \lambda_i c_i \ v_i v_i^\t - 3 \lambda I
= 3\lambda \biggl( 2 \sum_{i \in \Omega} v_i v_i^\t - I \biggr)
\]
where $\Omega := \{ i \in [k] : c_i \neq 0 \}$.
This implies that for any unit vector $w \in \R^n$,
\[
w^\t \nabla_u^2 \calL(u,\lambda) w
= 3\lambda \biggl( 2 \sum_{i \in \Omega} (v_i^\t w)^2 - 1 \biggr)
.
\]
The point $u$ is an isolated local maximum if the above quantity is
strictly negative for all unit vectors $w$ orthogonal to $u$.
We now consider three cases depending on the cardinality of $\Omega$ and
the sign of $\lambda$.
\begin{itemize}
\item Case 1: $|\Omega| = 1$ and $\lambda > 0$.
This means $u = v_i$ for some $i \in [k]$ (as $u = -v_i$ implies $\lambda =
-\lambda_i < 0$).
In this case,
\[
w^\t \nabla_u^2 \calL(u,\lambda) w = 3\lambda_i (2 (v_i^\t w)^2 - 1)
= -3\lambda_i < 0
\]
for all $w \in \R^n$ satisfying $(u^\t w)^2 = (v_i^\t w)^2 = 0$.
Hence $u$ is an isolated local maximizer.

\item Case 2: $|\Omega| \geq 2$ and $\lambda > 0$.
Since $|\Omega| \geq 2$, we may pick a strict non-empty subset $S
\subsetneq \Omega$ and set
\[ w
:=
\frac1Z \biggl(
\frac1{Z_S} \sum_{i \in S} c_i v_i
- \frac1{Z_{S^c}} \sum_{i \in \Omega \setminus S} c_i v_i
\biggr)
\]
where $Z_S := \sum_{i \in S} c_i^2$, $Z_{S^c} := \sum_{i \in \Omega
\setminus S} c_i^2$, and $Z := \sqrt{1/Z_S + 1/Z_{S^c}}$.
It is easy to check that $\|w\|^2 = \sum_{i \in \Omega} (v_i^\t w)^2 = 1$
and $u^\t w = 0$.
Consider any open neighborhood $U$ of $u$, and pick $\delta > 0$ small
enough so that $\tl{u} := \sqrt{1-\delta^2}u + \delta w$ is contained in
$U$.
Set $u_0 := \sqrt{1-\delta^2} u$.
By Taylor's theorem, there exists $\epsilon \in [0,\delta]$ such that, for
$\bar{u} := u_0 + \epsilon w$, we have
\begin{align*}
T(\tl{u},\tl{u},\tl{u})
& = T(u_0,u_0,u_0) + \nabla_u T(u,u,u)^\t (\tl{u} - u_0) \Big|_{u = u_0}
\\
& \qquad
+ \frac12 (\tl{u} - u_0)^\t \nabla_u^2 T(u,u,u) (\tl{u} - u_0) \Big|_{u =
\bar{u}}
\\
& = (1-\delta^2)^{3/2} \lambda +
\delta (1-\delta^2) \lambda u^\t w
+ \frac12 \delta^2 w^\t \nabla_u^2 T(u,u,u) w \Big|_{u=\bar{u}}
\\
& = (1-\delta^2)^{3/2} \lambda + 0
+ 3\delta^2 \sum_{i=1}^k \lambda_i (v_i^\t
(u_0 + \epsilon w)) (v_i^\t w)^2
\\
& = (1-\delta^2)^{3/2} \lambda + 3\delta^2 \sqrt{1-\delta^2} \sum_{i=1}^k
\lambda_i c_i (v_i^\t w)^2
+ 3\delta^2 \epsilon \sum_{i=1}^k \lambda_i (v_i^\t w)^3
\\
& = (1-\delta^2)^{3/2} \lambda + 3\delta^2 \sqrt{1-\delta^2} \lambda
\sum_{i \in \Omega} (v_i^\t w)^2
+ 3\delta^2 \epsilon \sum_{i=1}^k \lambda_i (v_i^\t w)^3
\\
& = (1-\delta^2)^{3/2} \lambda + 3\delta^2 \sqrt{1-\delta^2} \lambda
+ 3\delta^2 \epsilon \sum_{i=1}^k \lambda_i (v_i^\t w)^3
\\
& = \biggl( 1 - \frac32 \delta^2 + O(\delta^4) \biggr) \lambda
+ 3\delta^2 \sqrt{1-\delta^2} \lambda
+ 3\delta^2 \epsilon \sum_{i=1}^k \lambda_i (v_i^\t w)^3
.
\end{align*}
Since $\epsilon \leq \delta$, for small enough $\delta$, the RHS is
strictly greater than $\lambda$.
This implies that $u$ is not an isolated local maximizer.

\item Case 3: $|\Omega| = 0$ or $\lambda = 0$.
Note that if $|\Omega| = 0$, then $\lambda = 0$, so we just consider
$\lambda = 0$.
Consider any open neighborhood $U$ of $u$, and pick $j \in [n]$ and
$\delta > 0$ small enough so that $\tl{u} := \sqrt{1-\delta^2} u +
\delta v_j$ is contained in $U$.
Then
\begin{align*}
T(\tl{u},\tl{u},\tl{u})
& = (1-\delta^2)^{3/2} T(u,u,u) + 3\lambda_j (1-\delta^2)\delta c_j^2
+ 3\lambda_i \sqrt{1-\delta^2}\delta^2 c_j + \delta^3
> 0 = \lambda
\end{align*}
for sufficiently small $\delta$.
Thus $u$ is not an isolated local maximizer.

\end{itemize}
From these exhaustive cases, we conclude that a stationary point $u$ is an
isolated local maximizer if and only if $u = v_i$ for some $i \in [k]$.
\end{proof}

We are grateful to Hanzhang Hu, Drew Bagnell, and Martial Hebert for
alerting us of an issue with our original statement of
Theorem~\ref{thm:variational} and its proof, and for suggesting a
simple fix.
The original statement used the optimization constraint $\|u\|=1$
(rather than $\|u\| \leq 1$), but the characterization of the
decomposition with this constraint is then only given by isolated
local maximizers $u$ with the additional constraint that $T(u,u,u) >
0$---that is, there can be isolated local maximizers with $T(u,u,u)
\leq 0$ that are not vectors in the decomposition.
The suggested fix of Hu, Bagnell, and Herbert is to relax to $\|u\|
\leq 1$, which eliminates isolated local maximizers with $T(u,u,u)
\leq 0$; this way, the characterization of the decomposition is simply
the isolated local maximizers under the relaxed constraint.


\section{Analysis of Robust Power Method}
\label{app:analysis}

In this section, we prove Theorem~\ref{thm:robustpower}.
The proof is structured as follows.
In Appendix~\ref{app:init}, we show that with high probability, at least
one out of $L$ random vectors will be a good initializer for the tensor
power iterations.
An initializer is good if its projection onto an eigenvector is noticeably
larger than its projection onto other eigenvectors.
We then analyze in Appendix~\ref{app:iter} the convergence behavior of the
tensor power iterations.
Relative to the proof of Lemma~\ref{lemma:fixed-point}, this analysis is
complicated by the tensor perturbation.
We show that there is an initial slow convergence phase (linear rate rather
than quadratic), but as soon as the projection of the iterate onto an
eigenvector is large enough, it enters the quadratic convergence regime
until the perturbation dominates.
Finally, we show how errors accrue due to deflation in
Appendix~\ref{app:deflation}, which is rather subtle and different from
deflation with matrix eigendecompositions.
This is because when some initial set of eigenvectors and eigenvalues are
accurately recovered, the additional errors due to deflation are
effectively only lower-order terms.
These three pieces are assembled in Appendix~\ref{app:main} to complete
the proof of Theorem~\ref{thm:robustpower}.

\subsection{Initialization}
\label{app:init}

Consider a set of non-negative numbers $\tlambda_1, \tlambda_2, \dotsc,
\tlambda_k \geq 0$.
For $\gamma \in (0,1)$, we say a unit vector $\th{0} \in \R^k$ is
\emph{$\gamma$-separated} relative to $i^* \in [k]$ if
\[ \tlambda_{i^*} |\th{i^*,0}| - \max_{i \in [k] \setminus \{i^*\}}
\tlambda_i |\th{i,0}| \geq \gamma \tlambda_i |\th{i^*,0}| \]
(the dependence on $\tlambda_1, \tlambda_2, \dotsc, \tlambda_k$ is
implicit).

The following lemma shows that for any constant $\gamma$, with probability
at least $1-\eta$, at least one out of $\poly(k) \log(1/\eta)$
i.i.d.~random vectors (uniformly distributed over the unit sphere
$S^{k-1}$) is $\gamma$-separated relative to $\arg\max_{i \in [k]}
\tlambda_i$.
(For small enough $\gamma$ and large enough $k$, the polynomial is close to
linear in $k$.)
\begin{lem}\label{lem:initialization}
There exists an absolute constant $c > 0$ such that if positive integer
$L \geq 2$ satisfies
\begin{equation} \label{eq:L}
\sqrt{\frac{\ln(L)}{\ln(k)}}
\cdot \Biggl( 1 - \frac{\ln(\ln(L)) + c}{4\ln(L)} -
\sqrt{\frac{\ln(8)}{\ln(L)}} \Biggr)
\geq \frac{1}{1-\gamma} \cdot \Biggl( 1 + \sqrt{\frac{\ln(4)}{\ln(k)}}
\Biggr)
,
\end{equation}
the following holds.
With probability at least $1/2$ over the choice of $L$ i.i.d.~random
vectors drawn uniformly distributed over the unit sphere $S^{k-1}$ in
$\R^k$, at least one of the vectors is $\gamma$-separated relative to
$\arg\max_{i \in [k]} \tlambda_i$.
Moreover, with the same $c$, $L$, and for any $\eta \in (0,1)$, with
probability at least $1-\eta$ over $L\cdot\log_2(1/\eta)$ i.i.d.~uniform
random unit vectors, at least one of the vectors is $\gamma$-separated.
\end{lem}
\begin{proof}
Without loss of generality, assume $\arg\max_{i \in [k]} \tlambda_i = 1$.
Consider a random matrix $Z \in \R^{k \times L}$ whose entries are
independent~$\N(0,1)$ random variables; we take the $j$-th column of $Z$ to
be comprised of the random variables used for the $j$-th random vector
(before normalization).
Specifically, for the $j$-th random vector,
\[ \th{i,0} := \frac{Z_{i,j}}{\sqrt{\sum_{i'=1}^k Z_{i',j}^2}} , \quad i
\in [n] .
\]
It suffices to show that with probability at least $1/2$, there is a column
$j^* \in [L]$ such that
\[ |Z_{1,j^*}| \geq \frac1{1-\gamma} \max_{i \in [k] \setminus \{1\}}
|Z_{i,j^*}| . \]

Since $\max_{j \in [L]} |Z_{1,j}|$ is a $1$-Lipschitz function of $L$
independent $\N(0,1)$ random variables, it follows that
\[
\Pr\biggl[ \Bigl| \max_{j \in [L]} |Z_{1,j}| - \med \Bigl[ \max_{j \in [L]}
|Z_{1,j}| \Bigr] \Bigr| > \sqrt{2\ln(8)} \biggr] \leq 1/4 .
\]
Moreover,
\[
\med \Bigl[ \max_{j \in [L]} |Z_{1,j}| \Bigr]
\geq
\med \Bigl[ \max_{j \in [L]} Z_{1,j} \Bigr]
=: m .
\]
Observe that the cumulative distribution function of $\max_{j \in [L]}
Z_{1,j}$ is given by $F(z) = \Phi(z)^{L}$, where $\Phi$ is the standard
Gaussian CDF.
Since $F(m) = 1/2$, it follows that $m = \Phi^{-1}(2^{-1/{L}})$.
It can be checked that
\[
\Phi^{-1}(2^{-1/{L}})
\geq \sqrt{2\ln(L)} - \frac{\ln(\ln(L)) + c}{2\sqrt{2\ln(L)}}
\]
for some absolute constant $c > 0$.
Also, let $j^* := \arg\max_{j \in [L]} |Z_{1,j}|$.

Now for each $j \in [L]$, let $|Z_{2:k,j}| := \max \{ |Z_{2,j}|,
|Z_{3,j}|, \dotsc, |Z_{k,j}| \}$.
Again, since $|Z_{2:k,j}|$ is a $1$-Lipschitz function of $k-1$ independent
$\N(0,1)$ random variables, it follows that
\[
\Pr\biggl[ |Z_{2:k,j}| > \E\Bigl[ |Z_{2:k,j}| \Bigr] +
\sqrt{2\ln(4)} \biggr] \leq 1/4 .
\]
Moreover, by a standard argument,
\[
\E\Bigl[ |Z_{2:k,j}| \Bigr] \leq \sqrt{2\ln(k)} .
\]
Since $|Z_{2:k,j}|$ is independent of $|Z_{1,j}|$ for all $j \in [L]$,
it follows that the previous two displayed inequalities also hold with $j$
replaced by $j^*$.

Therefore we conclude with a union bound that with probability at least
$1/2$,
\[
|Z_{1,j^*}| \geq
\sqrt{2\ln(L)} - \frac{\ln(\ln(L)) + c}{2\sqrt{2\ln(L)}}
- \sqrt{2\ln(8)}
\quad \text{and} \quad
|Z_{2:k,j^*}| \leq
\sqrt{2\ln (k)} + \sqrt{2\ln(4)}
.
\]
Since $L$ satisfies~\eqref{eq:L} by assumption, in this event, the $j^*$-th
random vector is $\gamma$-separated.
\end{proof}

\subsection{Tensor Power Iterations}
\label{app:iter}

Recall the update rule used in the power method.
Let $\th{t} = \sum_{i=1}^k \th{i,t} v_i \in \R^k$ be the unit vector at
time $t$.
Then
\begin{align*}
\th{t+1} = \sum_{i=1}^k \th{i,t+1} v_i
& := \tilde{T}(I,\th{t},\th{t}) / \|\tilde{T}(I,\th{t},\th{t})\| .
\end{align*}

In this subsection, we assume that $\tilde{T}$ has the form
\begin{equation} \label{eq:tildeT}
\tilde{T} = \sum_{i=1}^k \tlambda_i
v_i^{\otimes 3} + \tilde{E}
\end{equation}
where $\{ v_1, v_2, \dotsc, v_k \}$ is an orthonormal basis, and, without
loss of generality,
\[ \tlambda_1 |\th{1,t}| = \max_{i \in [k]} \tlambda_i |\th{i,t}| > 0 . \]
Also, define
\[ \tlambdamin := \min \{ \tlambda_i : i \in [k], \ \tlambda_i > 0 \} ,
\quad \tlambdamax := \max \{ \tlambda_i : i \in [k] \} . \]

We further assume the error $\tilde{E}$ is a symmetric tensor such that,
for some constant $p > 1$,
\begin{align}
\|\tilde{E}(I,u,u)\| & \leq \teps , \quad \forall u \in S^{k-1} ;
\label{eq:regular-err}
\\
\|\tilde{E}(I,u,u)\| & \leq \teps / p , \quad \forall u \in S^{k-1} \
\text{s.t.} \ (u^\t v_1)^2 \geq 1 - (3\teps/\tlambda_1)^2
.
\label{eq:smaller-err}
\end{align}
In the next two propositions (Propositions~\ref{prop:simple}
and~\ref{prop:one-step}) and next two lemmas (Lemmas~\ref{lem:r}
and~\ref{lem:R}), we analyze the power method iterations using $\tilde{T}$
at some arbitrary iterate $\th{t}$ using only the
property~\eqref{eq:regular-err} of $\tilde{E}$.
But throughout, the quantity $\teps$ can be replaced by $\teps/p$ if
$\th{t}$ satisfies $(\th{t}^\t v_1)^2 \geq 1 - (3\teps/\tlambda_1)^2$ as
per property~\eqref{eq:smaller-err}.

Define
\begin{equation} \label{eq:defns}
\begin{aligned}
R_{\tau} & := \biggl( \frac{\th{1,\tau}^2}{1 - \th{1,\tau}^2} \biggr)^{1/2} ,
& r_{i,\tau} & := \frac{\tlambda_1\th{1,\tau}}{\tlambda_i |\th{i,\tau}|} ,
\\
\gamma_{\tau} & := 1 - \frac1{\min_{i\neq1}|r_{i,\tau}|} ,
& \delta_{\tau} & := \frac {\teps} {\tlambda_1 \th{1,\tau}^2} ,
& \kappa & := \frac{\tlambdamax}{\tlambda_1}
\end{aligned}
\end{equation}
for $\tau \in \{t,t+1\}$.

\begin{prop} \label{prop:simple}
\begin{align*}
\min_{i\neq1} |r_{i,t}| & \geq \frac{R_t}{\kappa} , &
\gamma_t & \geq 1 - \frac{\kappa}{R_t} , &
\th{1,t}^2 & = \frac{R_t^2}{1+R_t^2}
.
\end{align*}
\end{prop}

\begin{prop} \label{prop:one-step}
\begin{align}
r_{i,t+1}
& \geq
r_{i,t}^2 \cdot \frac{1 - \delta_t}
{1 + \kappa \delta_t r_{i,t}^2}
= \frac{1 - \delta_t}
{\frac1{r_{i,t}^2} + \kappa \delta_t}
, \quad i \in [k] ,
\label{eq:ratio-ineq}
\\
R_{t+1}
& \geq
R_t \cdot \frac{1-\delta_t} {1-\gamma_t + \delta_t R_t}
\geq \frac{1-\delta_t} {\frac{\kappa}{R_t^2} + \delta_t}
.
\label{eq:energy-ineq}
\end{align}
\end{prop}
\begin{proof}
Let $\ut{t+1} := \tilde{T}(I,\th{t},\th{t})$, so $\th{t+1} = \ut{t+1} /
\|\ut{t+1}\|$.
Since $\ut{i,t+1} = \tilde{T}(v_i,\th{t},\th{t}) = T(v_i,\th{t},\th{t}) +
E(v_i,\th{t},\th{t})$, we have
\begin{equation*}
\ut{i,t+1} = \tlambda_i \th{i,t}^2 + E(v_i,\th{t},\th{t}) , \quad i \in [k]
.
\end{equation*}
Using the triangle inequality and the fact $\|E(v_i,\th{t},\th{t})\| \leq
\teps$, we have
\begin{equation} \label{eq:coord-lb}
\ut{i,t+1}
\geq \tlambda_i \th{i,t}^2 - \teps
\geq |\th{i,t}| \cdot \Bigl( \tlambda_i |\th{i,t}| - \teps / |\th{i,t}|
\Bigr)
\end{equation}
and
\begin{equation} \label{eq:coord-ub}
|\ut{i,t+1}|
\leq |\tlambda_i \th{i,t}^2| + \teps
\leq |\th{i,t}| \cdot \Bigl( \tlambda_i |\th{i,t}| + \teps / |\th{i,t}|
\Bigr)
\end{equation}
for all $i \in [k]$.
Combining \eqref{eq:coord-lb} and \eqref{eq:coord-ub} gives
\[
r_{i,t+1}
= \frac {\tlambda_1\th{1,t+1}} {\tlambda_i|\th{i,t+1}|}
= \frac {\tlambda_1\ut{1,t+1}} {\tlambda_i|\ut{i,t+1}|}
\geq
r_{i,t}^2 \cdot \frac {1 - \delta_t} {1 +
\frac{\teps}{\tlambda_i\th{i,t}^2}}
=
r_{i,t}^2 \cdot \frac {1 - \delta_t}
{1 + (\tlambda_i/\tlambda_1) \delta_t r_{i,t}^2}
\geq
r_{i,t}^2 \cdot \frac {1 - \delta_t}
{1 + \kappa \delta_t r_{i,t}^2}
.
\]

Moreover, by the triangle inequality and H\"older's inequality,
\begin{align}
\biggl( \sum_{i=2}^n [\ut{i,t+1}]^2 \biggr)^{1/2}
& = \biggl( \sum_{i=2}^n \Bigl( \tlambda_i \th{i,t}^2 + E(v_i,\th{t},\th{t})
\Bigr)^2 \biggr)^{1/2}
\nonumber \\
& \leq \biggl( \sum_{i=2}^n \tlambda_i^2 \th{i,t}^4 \biggr)^{1/2}
+ \biggl( \sum_{i=2}^n E(v_i,\th{t},\th{t})^2 \biggr)^{1/2}
\nonumber \\
& \leq \max_{i\neq1} \tlambda_i |\th{i,t}|
\biggl( \sum_{i=2}^n \th{i,t}^2 \biggr)^{1/2} + \teps
\nonumber \\
& = (1 - \th{1,t}^2)^{1/2} \cdot \Bigl( \max_{i\neq1} \tlambda_i |\th{i,t}|
+ \teps / (1 - \th{1,t}^2)^{1/2} \Bigr)
.
\label{eq:other-coord-ub}
\end{align}
Combining \eqref{eq:coord-lb} and \eqref{eq:other-coord-ub} gives
\[
\frac {|\th{1,t+1}|} {(1 - \th{1,t+1}^2)^{1/2}}
=
\frac {|\ut{1,t+1}|} {\Bigl( \sum_{i=2}^n [\ut{i,t+1}]^2 \Bigr)^{1/2}}
\geq
\frac{|\th{1,t}|}{(1 - \th{1,t}^2)^{1/2}}
\cdot
\frac {\tlambda_1 |\th{1,t}| - \teps / |\th{1,t}|} {\max_{i\neq1} \tlambda_i
|\th{i,t}| + \teps / (1 - \th{1,t}^2)^{1/2}}
.
\]
In terms of $R_{t+1}$, $R_t$, $\gamma_t$, and $\delta_t$, this reads
\begin{equation*}
R_{t+1} \geq \frac {1-\delta_t}
{(1-\gamma_t) \Bigl( \frac{1-\th{1,t}^2}{\th{1,t}^2} \Bigr)^{1/2} +
\delta_t}
= R_t \cdot \frac{1-\delta_t} {1-\gamma_t + \delta_t R_t}
= \frac{1-\delta_t} {\frac{1-\gamma_t}{R_t} + \delta_t}
\geq \frac{1-\delta_t} {\frac{\kappa}{R_t^2} +
\delta_t}
\end{equation*}
where the last inequality follows from Proposition~\ref{prop:simple}.
\end{proof}

\begin{lem} \label{lem:r}
Fix any $\rho > 1$.
Assume
\[ 0 \leq \delta_t < \min\Bigl\{ \frac1{2(1+2\kappa\rho^2)}, \
\frac{1-1/\rho}{1+\kappa\rho} \Bigr\} \]
and $\gamma_t > 2(1+2\kappa\rho^2) \delta_t$.
\begin{enumerate}
\item If $r_{i,t}^2 \leq 2\rho^2$, then $r_{i,t+1} \geq |r_{i,t}| \bigl( 1
+ \frac{\gamma_t}{2} \bigr)$.

\item If $\rho^2 < r_{i,t}^2$, then $r_{i,t+1} \geq \min\{ r_{i,t}^2 /
\rho, \ \frac{1-\delta_t - 1/\rho}{\kappa\delta_t} \}$.

\item $\gamma_{t+1} \geq \min\{ \gamma_t , 1 - 1/\rho \}$.

\item If $\min_{i\neq1} r_{i,t}^2 > (\rho(1-\delta_t)-1)/(\kappa\delta_t)$,
then $R_{t+1} > \frac{1-\delta_t-1/\rho}{\kappa\delta_t} \cdot
\frac{\tlambdamin}{\tlambda_1} \cdot \frac1{\sqrt{k}}$.

\item If $R_t \leq 1 + 2\kappa\rho^2$, then $R_{t+1} \geq R_t \bigl( 1 +
\frac{\gamma_t}{3} \bigr)$, $\th{1,t+1}^2 \geq \th{1,t}^2$, and
$\delta_{t+1} \leq \delta_t$.

\end{enumerate}
\end{lem}
\begin{proof}
Consider two (overlapping) cases depending on $r_{i,t}^2$.
\begin{itemize}
\item Case 1: $r_{i,t}^2 \leq 2\rho^2$.
By~\eqref{eq:ratio-ineq} from Proposition~\ref{prop:one-step},
\[
r_{i,t+1}
\geq r_{i,t}^2 \cdot \frac{1-\delta_t}{1+\kappa\delta_tr_{i,t}^2}
\geq |r_{i,t}| \cdot \frac{1}{1-\gamma_t} \cdot
\frac{1-\delta_t}{1+2\kappa\rho^2\delta_t}
\geq |r_{i,t}| \Bigl( 1 + \frac{\gamma_t}{2} \Bigr)
\]
where the last inequality uses the assumption $\gamma_t >
2(1+2\kappa\rho^2)\delta_t$.
This proves the first claim.

\item Case 2: $\rho^2 < r_{i,t}^2$.
We split into two sub-cases.
Suppose $r_{i,t}^2 \leq (\rho(1-\delta_t)-1)/(\kappa\delta_t)$.
Then, by~\eqref{eq:ratio-ineq},
\[
r_{i,t+1}
\geq r_{i,t}^2 \cdot \frac{1-\delta_t}{1+\kappa\delta_tr_{i,t}^2}
\geq r_{i,t}^2 \cdot \frac{1-\delta_t}{1+\kappa\delta_t
\frac{\rho(1-\delta_t)-1}{\kappa\delta_t}}
= \frac{r_{i,t}^2}{\rho}
.
\]
Now suppose instead $r_{i,t}^2 > (\rho(1-\delta_t)-1)/(\kappa\delta_t)$.
Then
\begin{equation} \label{eq:r-limit}
r_{i,t+1} \geq
\frac{1-\delta_t}
{\frac{\kappa\delta_t}{\rho(1-\delta_t)-1} + \kappa\delta_t}
= \frac{1-\delta_t - 1/\rho}{\kappa\delta_t}
.
\end{equation}

\end{itemize}
Observe that if $\min_{i \neq 1} r_{i,t}^2 \leq
(\rho(1-\delta_t)-1)/(\kappa\delta_t)$, then $r_{i,t+1} \geq |r_{i,t}|$ for
all $i \in [k]$, and hence $\gamma_{t+1} \geq \gamma_t$.
Otherwise we have $\gamma_{t+1} > 1 - \frac{\kappa\delta_t}{1-\delta_t -
1/\rho} > 1-1/\rho$.
This proves the third claim.

If $\min_{i\neq1} r_{i,t}^2 > (\rho(1-\delta_t)-1)/(\kappa\delta_t)$, then
we may apply the inequality~\eqref{eq:r-limit} from the second sub-case of
Case 2 above to get
\begin{align*}
R_{t+1} &
= \frac1{\Bigl( \sum_{i\neq1} (\tlambda_1/\tlambda_i)^2 / r_{i,t+1}^2
\Bigr)^{1/2}}
>
\biggl( \frac{1-\delta_t-1/\rho}{\kappa\delta_t} \biggr)
\cdot \frac{\tlambdamin}{\tlambda_1}
\cdot \frac{1}{\sqrt{k}}
.
\end{align*}
This proves the fourth claim.

Finally, for the last claim, if $R_t \leq 1 + 2\kappa\rho^2$, then by
\eqref{eq:energy-ineq} from Proposition~\ref{prop:one-step} and the
assumption $\gamma_t > 2(1+2\kappa\rho^2)\delta_t$,
\begin{align*}
R_{t+1}
&
\geq R_t \cdot \frac{1-\delta_t}{1-\gamma_t +\delta_tR_t}
\geq R_t \cdot \frac{1-\frac{\gamma_t}{2(1+2\kappa\rho^2)}}
{1-\gamma_t/2}
\geq R_t \biggl( 1 + \gamma_t \cdot
\frac{\kappa\rho^2}{1+2\kappa\rho^2} \biggr)
\geq R_t \Bigl( 1 + \frac{\gamma_t}3 \Bigr)
.
\end{align*}
This in turn implies that $\th{1,t+1}^2 \geq \th{1,t}^2$ via
Proposition~\ref{prop:simple}, and thus $\delta_{t+1} \leq \delta_t$.
\end{proof}

\begin{lem} \label{lem:R}
Assume $0 \leq \delta_t < 1/2$ and $\gamma_t > 0$.
Pick any $\beta > \alpha > 0$ such that
\[
\frac{\alpha}{(1+\alpha)(1+\alpha^2)} \geq \frac{\teps}{\gamma_t\tlambda_1}
, \quad
\frac{\alpha}{2(1+\alpha)(1+\beta^2)} \geq \frac{\teps}{\tlambda_1}
.
\]
\begin{enumerate}
\item If $R_t \geq 1/\alpha$, then $R_{t+1} \geq 1/\alpha$.

\item If $1/\alpha > R_t \geq 1/\beta$, then $R_{t+1} \geq \min\{
R_t^2 / (2\kappa), \ 1/\alpha \}$.

\end{enumerate}
\end{lem}
\begin{proof}
Observe that for any $c > 0$,
\begin{equation} \label{eq:regime-impl}
R_t \geq \frac1c \quad \Leftrightarrow \quad
\th{1,t}^2 \geq \frac1{1+c^2} \quad \Leftrightarrow \quad
\delta_t \leq \frac{(1+c^2)\teps}{\tlambda_1} .
\end{equation}
Now consider the following cases depending on $R_t$.
\begin{itemize}
\item Case 1: $R_t \geq 1/\alpha$.
In this case, we have
\[
\delta_t
\leq \frac{(1+\alpha^2)\teps}{\tlambda_1}
\leq \frac{\alpha\gamma_t}{1+\alpha}
\]
by~\eqref{eq:regime-impl} (with $c=\alpha$) and the condition on $\alpha$.
Combining this with~\eqref{eq:energy-ineq} from
Proposition~\ref{prop:one-step} gives
\[ R_{t+1} \geq \frac{1-\delta_t}{\frac{1-\gamma_t}{R_t} + \delta_t}
\geq \frac{1 - \frac{\alpha\gamma_t}{1+\alpha}}
{(1-\gamma_t)\alpha + \frac{\alpha\gamma_t}{1+\alpha}}
= \frac1{\alpha}
.
\]

\item Case 2: $1/\beta \leq R_t < 1/\alpha$.
In this case, we have
\[ \delta_t \leq \frac{(1+\beta^2)\teps}{\tlambda_1} \leq
\frac{\alpha}{2(1+\alpha)} \]
by~\eqref{eq:regime-impl} (with $c=\beta$) and the conditions on $\alpha$
and $\beta$.
If $\delta_t \geq 1/(2+R_t^2/\kappa)$, then~\eqref{eq:energy-ineq} implies
\[ R_{t+1} \geq \frac{1 - \delta_t} {\frac{\kappa}{R_t^2} + \delta_t}
\geq \frac{1 - 2\delta_t} {2\delta_t}
\geq \frac{1 - \frac{\alpha}{1+\alpha}} {\frac{\alpha}{1+\alpha}}
= \frac1{\alpha} .
\]
If instead $\delta_t < 1/(2+R_t^2/\kappa)$, then~\eqref{eq:energy-ineq}
implies
\[ R_{t+1}
\geq \frac{1 - \delta_t} {\frac{\kappa}{R_t^2} + \delta_t}
> \frac{1 - \frac{1}{2+R_t^2/\kappa}}
{\frac{\kappa}{R_t^2} + \frac{1}{2+R_t^2/\kappa}}
= \frac{R_t^2} {2\kappa} .
\]
\end{itemize}
\end{proof}

\subsubsection{Approximate Recovery of a Single Eigenvector}

We now state the main result regarding the approximate recovery of a single
eigenvector using the tensor power method on $\tilde{T}$.
Here, we exploit the special properties of the error $\tilde{E}$---both~\eqref{eq:regular-err} and~\eqref{eq:smaller-err}.

\begin{lem} \label{lem:single}
There exists a universal constant $C > 0$ such that the following holds.
Let $i^* := \arg\max_{i \in [k]} \tlambda_i |\th{i,0}|$.
If
\begin{equation*}
\teps < \frac{\gamma_0}{2(1+8\kappa)}
\cdot \tlambdamin \cdot \th{i^*,0}^2 \quad\text{and}\quad
N \geq
C \cdot \biggl( \frac{\log(k\kappa)}{\gamma_0} +
\log\log\frac{p\tlambda_{i^*}}{\teps}
\biggr)
,
\end{equation*}
then after $t \geq N$ iterations of the tensor power method on tensor
$\tilde{T}$ as defined in~\eqref{eq:tildeT} and
satisfying~\eqref{eq:regular-err} and~\eqref{eq:smaller-err}, the final
vector $\th{t}$ satisfies
\[
\th{i^*,t} \geq \sqrt{1 - \biggl( \frac{3\teps}{p\tlambda_{i^*}} \biggr)^2}
,
\quad
\|\th{t} - v_{i^*}\| \leq \frac{4\teps}{p\tlambda_{i^*}}
,
\quad
|\tilde{T}(\th{t},\th{t},\th{t}) - \tlambda_{i^*}|
\leq
\biggl( 27\kappa \Bigl( \frac{\teps}{p\lambda_{i^*}} \Bigr)^2 + 2
\biggr) \frac{\teps}{p}
.
\]
\end{lem}
\begin{proof}
Assume without loss of generality that $i^* = 1$.
We consider three phases: (i) iterations before the first time $t$ such
that $R_t > 1 + 2\kappa\rho^2 = 1 + 8\kappa$ (using $\rho := 2$), (ii) the
subsequent iterations before the first time $t$ such that $R_t \geq
1/\alpha$ (where $\alpha$ will be defined below), and finally (iii)
the remaining iterations.

We begin by analyzing the first phase, \emph{i.e.}, the iterates in $T_1 :=
\{ t \geq 0 : R_t \leq 1 + 2\kappa\rho^2 = 1 + 8\kappa \}$.
Observe that the condition on $\teps$ implies
\[
\delta_0
= \frac{\teps}{\tlambda_1 \th{1,0}^2}
< \frac{\gamma_0}{2(1+8\kappa)} \cdot \frac{\tlambdamin}{\tlambda_1}
\leq \min\biggl\{ \frac{\gamma_0}{2(1+2\kappa\rho^2)} ,
\frac{1-1/\rho}{2(1+2\kappa\rho^2)} \biggr\}
,
\]
and hence the preconditions on $\delta_t$ and $\gamma_t$ of
Lemma~\ref{lem:r} hold for $t = 0$.
For all $t \in T_1$ satisfying the preconditions, Lemma~\ref{lem:r} implies
that $\delta_{t+1} \leq \delta_t$ and $\gamma_{t+1} \geq \min\{\gamma_t ,
1-1/\rho\}$, so the next iteration also satisfies the preconditions.
Hence by induction, the preconditions hold for all iterations in $T_1$.
Moreover, for all $i \in [k]$, we have
\[
|r_{i,0}| \geq \frac{1}{1-\gamma_0} ;
\]
and while $t \in T_1$: (i) $|r_{i,t}|$ increases at a linear rate while
$r_{i,t}^2 \leq 2\rho^2$, and (ii) $|r_{i,t}|$ increases at a quadratic
rate while $\rho^2 \leq r_{i,t}^2 \leq
\frac{1-\delta_t-1/\rho}{\kappa\delta_t}$.
(The specific rates are given, respectively, in Lemma~\ref{lem:r}, claims 1
and 2.)
Since $\frac{1-\delta_t-1/\rho}{\kappa\delta_t} \leq
\frac{\tlambda_1}{2\kappa \teps}$, it follows that $\min_{i \neq 1}
r_{i,t}^2 \leq \frac{1-\delta_t-1/\rho}{\kappa\delta_t}$ for at most
\begin{equation} \label{eq:r-num_iter}
\frac{2}{\gamma_0} \ln\biggl( \frac{\sqrt{2\rho^2}}{\frac{1}{1-\gamma_0}}
\biggr)
+
\ln\biggl( \frac{\ln\frac{\tlambda_1}{2\kappa\teps}}{\ln\sqrt2} \biggr)
= O\biggl( \frac{1}{\gamma_0} + \log\log\frac{\tlambda_1}{\teps} \biggr)
\end{equation}
iterations in $T_1$.
As soon as $\min_{i \neq 1} r_{i,t}^2 >
\frac{1-\delta_t-1/\rho}{\kappa\delta_t}$, we have that in the next
iteration,
\[ R_{t+1} > \frac{1-\delta_t-1/\rho}{\kappa\delta_t} \cdot
\frac{\tlambdamin}{\tlambda_1} \cdot \frac{1}{\sqrt{k}} \geq
\frac{7}{\sqrt{k}} ; \]
and all the while $R_t$ is growing at a linear rate (given in
Lemma~\ref{lem:r}, claim 5).
Therefore, there are at most an additional
\begin{equation} \label{eq:R-num_iter}
1 + \frac{3}{\gamma_0} \ln\biggl( \frac{1+8\kappa}{7/\sqrt{k}} \biggr)
= O\biggl( \frac{\log(k\kappa)}{\gamma_0} \biggr)
\end{equation}
iterations in $T_1$ over that counted in~\eqref{eq:r-num_iter}.
Therefore, by combining the counts in~\eqref{eq:r-num_iter}
and~\eqref{eq:R-num_iter}, we have that the number of iterations in the
first phase satisfies
\[
|T_1|
= O\biggl( \log\log\frac{\tlambda_1}{\teps} + \frac{\log(k\kappa)}{\gamma_0}
\biggr)
.
\]

We now analyze the second phase, \emph{i.e.}, the iterates in $T_2 := \{ t
\geq 0 : t \notin T_1, \ R_t < 1/\alpha \}$.
Define
\[
\alpha := \frac{3\teps}{\tlambda_1} , \quad
\beta := \frac{1}{1+2\kappa\rho^2} = \frac{1}{1+8\kappa}
.
\]
Note that for the initial iteration $t' := \min T_2$, we have that $R_{t'}
\geq 1 + 2\kappa\rho^2 = 1 + 8\kappa = 1/\beta$, and by
Proposition~\ref{prop:simple}, $\gamma_{t'} \geq 1 - \kappa/(1 + 8\kappa) >
7/8$.
It can be checked that $\delta_t$, $\gamma_t$, $\alpha$, and $\beta$
satisfy the preconditions of Lemma~\ref{lem:R} for this initial iteration
$t'$.
For all $t \in T_2$ satisfying these preconditions, Lemma~\ref{lem:R} implies
that $R_{t+1} \geq \min\{ R_t, 1/\alpha \}$, $\th{1,t+1}^2 \geq \min\{
\th{1,t}^2 , 1/(1+\alpha^2) \}$ (via Proposition~\ref{prop:simple}),
$\delta_{t+1} \leq \max\{ \delta_t , (1+\alpha)^2\teps/\tlambda_1 \}$ (using
the definition of $\delta_t$), and $\gamma_{t+1} \geq \min \{ \gamma_t , 1
- \alpha\kappa \}$ (via Proposition~\ref{prop:simple}).
Hence the next iteration $t+1$ also satisfies the preconditions, and by
induction, so do all iterations in $T_2$.
To bound the number of iterations in $T_2$, observe that $R_t$ increases at
a quadratic rate until $R_t \geq 1/\alpha$, so
\begin{equation} \label{eq:R-num_iter2}
|T_2| \leq \ln\biggl( \frac{\ln(1/\alpha)}{\ln((1/\beta)/(2\kappa))} \biggr)
< \ln\biggl( \frac{\ln\frac{\tlambda_1}{3\teps}}{\ln4} \biggr)
= O\biggl( \log\log\frac{\tlambda_1}{\teps} \biggr)
.
\end{equation}
Therefore the total number of iterations before $R_t \geq 1/\alpha$ is
\[
O\biggl( \frac{\log(k\kappa)}{\gamma_0} + \log\log\frac{\tlambda_1}{\teps}
\biggr)
.
\]

After $R_{t''} \geq 1/\alpha$ (for $t'' := \max (T_1 \cup T_2) + 1$), we
have
\[
\th{1,t''}^2
\geq \frac{1/\alpha^2}{1+1/\alpha^2}
\geq 1 - \alpha^2 \geq 1 - \biggl( \frac{3\teps}{\tlambda_1} \biggr)^2 .
\]
Therefore, the vector $\th{t''}$ satisfies the condition for
property~\eqref{eq:smaller-err} of $\tilde{E}$ to hold.
Now we apply Lemma~\ref{lem:R} using $\teps/p$ in place of $\teps$,
including in the definition of $\delta_t$ (which we call
$\odelta_t$):
\[ \odelta_t := \frac{\teps}{p\tlambda_1\th{1,t}^2} ; \]
we also replace $\alpha$ and $\beta$ with $\oalpha$ and $\obeta$, which we
set to
\[ \oalpha := \frac{3\teps}{p\tlambda_1} , \quad \obeta :=
\frac{3\teps}{\tlambda_1} . \]
It can be checked that $\odelta_{t''} \in (0,1/2)$,
$\gamma_{t''} \geq 1 - 3\teps\kappa/\lambda_1 > 0$,
\[
\frac{\oalpha}{(1+\oalpha)(1+\oalpha^2)} \geq
\frac{\teps}{p(1 - 3\teps\kappa/\lambda_1)\tlambda_1} \geq
\frac{\teps}{p\gamma_{t''}\tlambda_1} ,
\quad
\frac{\oalpha}{2(1+\oalpha)(1+\obeta^2)} \geq
\frac{\teps}{p\tlambda_1} .
\]
Therefore, the preconditions of Lemma~\ref{lem:R} are satisfied for the
initial iteration $t''$ in this final phase, and by the same arguments as
before, the preconditions hold for all subsequent iterations $t \geq t''$.
Initially, we have $R_{t''} \geq 1/\alpha \geq 1/\obeta$, and by
Lemma~\ref{lem:R}, we have that $R_t$ increases at a quadratic rate in this
final phase until $R_t \geq 1/\oalpha$.
So the number of iterations before $R_t \geq 1/\oalpha$ can be bounded as
\[
\ln\biggl( \frac{\ln(1/\oalpha)}{\ln((1/\obeta)/(2\kappa))} \biggr)
= \ln\biggl(
\frac{\ln\frac{p\tlambda_1}{3\teps}}{\ln\Bigl(\frac{\lambda_1}{3\teps} \cdot
\frac{1}{2\kappa}\Bigr)} \biggr)
\leq \ln\ln\frac{p\tlambda_1}{3\teps}
= O\biggl( \log\log\frac{p\tlambda_1}{\teps} \biggr)
.
\]
Once $R_t \geq 1/\oalpha$, we have
\[
\th{1,t}^2
\geq 1 - \biggl( \frac{3\teps}{p\tlambda_1} \biggr)^2 .
\]
Since $\sign(\th{1,t}) = r_{1,t} \geq r_{1,t-1}^2 \cdot (1
- \odelta_{t-1}) / (1 + \kappa\odelta_{t-1}r_{1,t-1}^2) = (1 -
\odelta_{t-1}) / (1 + \kappa\odelta_{t-1}) > 0$ by
Proposition~\ref{prop:one-step}, we have $\th{1,t} > 0$.
Therefore we can conclude that
\[
\|\th{t} - v_1\|
= \sqrt{2(1-\th{1,t})}
\leq \sqrt{2\biggl(1-\sqrt{1 - (3\teps/(p\tlambda_1))^2}\biggr)}
\leq 4\teps/(p\tlambda_1)
.
\]
Finally,
\begin{align*}
|\tilde{T}(\th{t},\th{t},\th{t}) - \tlambda_1|
& = \biggl|
\tlambda_1 (\th{1,t}^3 - 1) + \sum_{i=2}^k \tlambda_i \th{i,t}^3
+ \tl{E}(\th{t},\th{t},\th{t})
\biggr|
\\
& \leq \tlambda_1 |\th{1,t}^3 - 1|
+ \sum_{i=2}^k \tlambda_i |\th{i,t}| \th{i,t}^2
+ \|\tl{E}(I,\th{t},\th{t})\|
\\
& \leq \tlambda_1 \bigl( 1 - \th{1,t} + |\th{1,t}(1 - \th{1,t}^2)| \bigr)
+ \max_{i\neq1} \tlambda_i |\th{i,t}| \sum_{i=2}^k \th{i,t}^2
+ \|\tl{E}(I,\th{t},\th{t})\|
\\
& \leq \tlambda_1 \bigl( 1 - \th{1,t} + |\th{1,t}(1 - \th{1,t}^2)| \bigr)
+ \max_{i\neq1} \tlambda_i \sqrt{1 - \th{1,t}^2} \sum_{i=2}^k \th{i,t}^2
+ \|\tl{E}(I,\th{t},\th{t})\|
\\
& = \tlambda_1 \bigl( 1 - \th{1,t} + |\th{1,t}(1 - \th{1,t}^2)| \bigr)
+ \max_{i\neq1} \tlambda_i (1 - \th{1,t}^2)^{3/2}
+ \|\tl{E}(I,\th{t},\th{t})\|
\\
& \leq \tlambda_1 \cdot 3 \biggl( \frac{3\teps}{p\tlambda_1} \biggr)^2
+ \kappa \tlambda_1\cdot \left(\frac{3\teps}{p\tlambda_1}\right)^3
+ \frac{\teps}{p}
\\
& \leq \frac{(27\kappa\cdot (\teps/p\tlambda_1)^2 + 2) \teps}{p}
.
\end{align*}
\end{proof}

\subsection{Deflation}
\label{app:deflation}

\begin{lem} \label{lem:deflation}
Fix some $\teps \geq 0$.
Let $\{ v_1, v_2, \dotsc, v_k \}$ be an orthonormal basis for $\R^k$,
and $\lambda_1, \lambda_2, \dotsc, \lambda_k \geq 0$ with
$\lambdamin := \min_{i \in [k]} \lambda_i$.
Also, let $\{ \hv_1, \hv_2, \dotsc, \hv_k \}$ be a set of unit vectors in
$\R^k$ (not necessarily orthogonal), $\hlambda_1, \hlambda_2, \dotsc,
\hlambda_k \geq 0$ be non-negative scalars, and define
\begin{align*}
\deflate_i & :=
\lambda_i v_i^{\otimes 3}
-
\hlambda_i \hv_i^{\otimes 3}
,
\quad i \in [k] .
\end{align*}
Pick any $t \in [k]$.
If
\begin{align*}
|\hlambda_i - \lambda_i| & \leq \teps , \\
\|\hv_i - v_i\| & \leq \min\{ \sqrt2, \ 2\teps / \lambda_i \}
\end{align*}
for all $i \in [t]$, then for any unit vector $u \in S^{k-1}$,
\begin{multline*}
\biggl\| \sum_{i=1}^t \deflate_i(I,u,u) \biggr\|_2^2
\leq
\biggl(
4(5 + 11\teps/\lambdamin)^2
+ 128 ( 1 + \teps / \lambdamin )^2 (\teps / \lambdamin)^2
\biggr) \teps^2 \sum_{i=1}^t (u^\t v_i)^2
\\
+ 64(1+\teps/\lambdamin)^2 \teps^2 \sum_{i=1}^t (\teps / \lambda_i)^2
+ 2048 ( 1 + \teps / \lambdamin )^2
\teps^2 \biggl( \sum_{i=1}^t (\teps / \lambda_i)^3 \biggr)^2
.
\end{multline*}
In particular, for any $\Delta \in (0,1)$, there exists a constant $\Delta'
> 0$ (depending only on $\Delta$) such that $\teps \leq \Delta'
\lambdamin/\sqrt{k}$ implies
\[
\biggl\| \sum_{i=1}^t \deflate_i(I,u,u) \biggr\|_2^2 \le \biggl( \Delta +
100 \sum_{i=1}^t (u^\t v_i)^2 \biggr) \teps^2 .
\]
\end{lem}
\begin{proof}
For any unit vector $u$ and $i \in [t]$, the error term
\[
\deflate_i(I,u,u)
= \lambda_i (u^\t v_i)^2 v_i - \hlambda_i (u^\t \hv_i)^2 \hv_i
\]
lives in $\operatorname{span}\{ v_i, \hv_i\}$; this space is the same as
$\operatorname{span}\{ v_i, \hv_i^\perp\}$, where
\begin{equation*}
\hv_i^\perp := \hv_i - (v_i^\t \hv_i) v_i
\end{equation*}
is the projection of $\hv_i$ onto the subspace orthogonal to $v_i$.
Since $\|\hv_i - v_i\|^2 = 2(1 - v_i^\t \hv_i)$, it follows that
\begin{equation*}
c_i := v_i^\t \hv_i
= 1 - \|\hv_i - v_i\|^2 / 2
\geq 0
\end{equation*}
(the inequality follows from the assumption $\|\hv_i - v_i\| \leq
\sqrt2$, which in turn implies $0 \leq c_i \leq 1$).
By the Pythagorean theorem and the above inequality for $c_i$,
\begin{equation*}
\|\hv_i^\perp\|^2
= 1 - c_i^2
\leq \|\hv_i - v_i\|^2
.
\end{equation*}
Later, we will also need the following bound, which is easily derived from
the above inequalities and the triangle inequality:
\begin{equation*}
|1 - c_i^3|
= |1 - c_i + c_i (1 - c_i^2)|
\leq 1 - c_i + |c_i (1 - c_i^2)|
\leq 1.5 \|\hv_i - v_i\|^2
.
\end{equation*}

We now express $\deflate_i(I,u,u)$ in terms of the coordinate system
defined by $v_i$ and $\hv_i^\perp$, depicted below.
Define
\begin{equation*}
a_i := u^\t v_i
\quad \text{and} \quad
b_i := u^\t \bigl( \hv_i^\perp / \|\hv_i^\perp\| \bigr)
.
\end{equation*}
(Note that the part of $u$ living in $\operatorname{span}\{v_i,
\hv_i^\perp\}^\perp$ is irrelevant for analyzing $\deflate_i(I,u,u)$.)
We have
\begin{align*}
\deflate_i(I,u,u)
& =
\lambda_i (u^\t v_i)^2 v_i - \hlambda_i (u^\t \hv_i)^2 \hv_i
\\
& =
\lambda_i a_i^2 v_i
- \hlambda_i
\bigl( a_i c_i + \|\hv_i^\perp\| b_i \bigr)^2
\bigl( c_i v_i + \hv_i^\perp \bigr)
\\
& =
\lambda_i a_i^2 v_i
- \hlambda_i
\bigl( a_i^2 c_i^2 + 2 \|\hv_i^\perp\| a_i b_i c_i + \|\hv_i^\perp\|^2
b_i^2 \bigr) c_i v_i
- \hlambda_i \bigl( a_i c_i + \|\hv_i^\perp\| b_i \bigr)^2 \hv_i^\perp
\\
& =
\underbrace{
\Bigl(
(\lambda_i - \hlambda_i c_i^3) a_i^2
- 2 \hlambda_i \|\hv_i^\perp\| a_i b_i c_i^2
- \hlambda_i \|\hv_i^\perp\|^2 b_i^2 c_i
\Bigr)}_{=: A_i} v_i
- \underbrace{
\hlambda_i \|\hv_i^\perp\|
\bigl( a_i c_i + \|\hv_i^\perp\| b_i \bigr)^2
}_{=: B_i}
\bigl( \hv_i^\perp / \|\hv_i^\perp\| \bigr)
\\
& = A_i v_i - B_i \bigl( \hv_i^\perp / \|\hv_i^\perp\| \bigr)
.
\end{align*}
The overall error can also be expressed in terms of the $A_i$ and $B_i$:
\begin{align}
\biggl\| \sum_{i=1}^t \deflate_i(I,u,u) \biggr\|_2^2
& =
\biggl\| \sum_{i=1}^t A_i v_i
-
\sum_{i=1}^t B_i (\hv_i^\perp / \|\hv_i^\perp\|) \biggr\|_2^2
\nonumber \\
& \leq
2 \biggl\| \sum_{i=1}^t A_i v_i \biggr\|^2
+ 2 \biggl\| \sum_{i=1}^t B_i (\hv_i^\perp / \|\hv_i^\perp\|) \biggr\|_2^2
\nonumber \\
& \leq
2 \sum_{i=1}^t A_i^2
+ 2 \biggl( \sum_{i=1}^t |B_i| \biggr)^2
\label{eq:A+B}
\end{align}
where the first inequality uses the fact $(x+y)^2 \leq 2(x^2+y^2)$ and the
triangle inequality, and the second inequality uses the orthonormality of
the $v_i$ and the triangle inequality.

It remains to bound $A_i^2$ and $|B_i|$ in terms of $|a_i|$, $\lambda_i$,
and $\teps$.
The first term, $A_i^2$, can be bounded using the triangle inequality and
the various bounds on $|\lambda_i - \hlambda_i|$, $\|\hv_i - v_i\|$,
$\|\hv_i^\perp\|$, and $c_i$:
\begin{align*}
|A_i|
& \leq
( |\lambda_i - \hlambda_i| c_i^3 + \lambda_i |c_i^3 - 1| ) a_i^2
+ 2 ( \lambda_i + |\lambda_i - \hlambda_i| ) \|\hv_i^\perp\|
|a_i b_i| c_i^2
+ ( \lambda_i + |\lambda_i - \hlambda_i| ) \|\hv_i^\perp\|^2 b_i^2
c_i
\\
& \leq ( |\lambda_i - \hlambda_i| + 1.5 \lambda_i \|\hv_i - v_i\|^2
+ 2(\lambda_i + |\lambda_i - \hlambda_i|) \|\hv_i - v_i\|) |a_i|
+ (\lambda_i + |\lambda_i - \hlambda_i|) \|\hv_i - v_i\|^2
\\
& \leq ( \teps + 7\teps^2/\lambda_i
+ 4\teps + 4\teps^2/\lambda_i ) |a_i|
+ 4 \teps^2/\lambda_i + \teps^3/\lambda_i^2
\\
& = ( 5 + 11\teps/\lambda_i ) \teps |a_i|
+ 4 (1 + \teps/\lambda_i) \teps^2/\lambda_i
,
\end{align*}
and therefore (via $(x+y)^2 \leq 2(x^2 + y^2)$)
\begin{equation*}
A_i^2 \leq 2(5 + 11\teps/\lambda_i)^2 \teps^2 a_i^2
+ 32(1+\teps/\lambda_i)^2 \teps^4 / \lambda_i^2
.
\end{equation*}
The second term, $|B_i|$, is bounded similarly:
\begin{align*}
|B_i|
& \leq
2 ( \lambda_i + |\lambda_i - \hlambda_i| )
\|\hv_i^\perp\|^2
( a_i^2 + \|\hv_i^\perp\|^2 )
\\
& \leq
2 ( \lambda_i + |\lambda_i - \hlambda_i| )
\|\hv_i - v_i\|^2
( a_i^2 + \|\hv_i - v_i\|^2 )
\\
& \leq
8 ( 1 + \teps / \lambda_i )
(\teps^2 / \lambda_i)
a_i^2
+
32 ( 1 + \teps / \lambda_i )
\teps^4 / \lambda_i^3
.
\end{align*}
Therefore, using the inequality from~\eqref{eq:A+B} and again $(x+y)^2 \leq
2(x^2 + y^2)$,
\begin{align*}
\biggl\| \sum_{i=1}^t \deflate_i(I,u,u) \biggr\|_2^2
& \leq
2 \sum_{i=1}^t A_i^2
+ 2 \biggl( \sum_{i=1}^t |B_i| \biggr)^2
\\
& \leq
4(5 + 11\teps/\lambdamin)^2 \teps^2 \sum_{i=1}^t a_i^2
+ 64(1+\teps/\lambdamin)^2 \teps^2 \sum_{i=1}^t (\teps / \lambda_i)^2
\\
& \qquad
+ 2\biggl(
8 ( 1 + \teps / \lambdamin )
(\teps^2 / \lambdamin)
\sum_{i=1}^t a_i^2
+ 32 ( 1 + \teps / \lambdamin )
\teps \sum_{i=1}^t (\teps / \lambda_i)^3
\biggr)^2
\\
& \leq
4(5 + 11\teps/\lambdamin)^2 \teps^2
\sum_{i=1}^t a_i^2
+ 64(1+\teps/\lambdamin)^2 \teps^2 \sum_{i=1}^t (\teps / \lambda_i)^2
\\
& \qquad
+ 128 ( 1 + \teps / \lambdamin )^2 (\teps / \lambdamin)^2 \teps^2
\sum_{i=1}^t a_i^2
\\
& \qquad
+ 2048 ( 1 + \teps / \lambdamin )^2
\teps^2 \biggl( \sum_{i=1}^t (\teps / \lambda_i)^3 \biggr)^2
\\
& =
\biggl(
4(5 + 11\teps/\lambdamin)^2
+ 128 ( 1 + \teps / \lambdamin )^2 (\teps / \lambdamin)^2
\biggr) \teps^2 \sum_{i=1}^t a_i^2
\\
& \qquad
+ 64(1+\teps/\lambdamin)^2 \teps^2 \sum_{i=1}^t (\teps / \lambda_i)^2
+ 2048 ( 1 + \teps / \lambdamin )^2
\teps^2 \biggl( \sum_{i=1}^t (\teps / \lambda_i)^3 \biggr)^2
.
\end{align*}
\end{proof}

\subsection{Proof of the Main Theorem}
\label{app:main}

\begin{thm2}{thm:robustpower}
Let $\hat{T} = T + E \in \R^{k \times k \times k}$, where $T$ is a
symmetric tensor with orthogonal decomposition $T = \sum_{i=1}^k \lambda_i
v_i^{\otimes 3}$ where each $\lambda_i > 0$, $\{ v_1, v_2, \dotsc, v_k \}$
is an orthonormal basis, and $E$ has operator norm $\eps := \|E\|$.
Define $\lambdamin := \min\{ \lambda_i : i \in [k] \}$, and $\lambdamax :=
\max\{ \lambda_i : i \in [k] \}$.
There exists universal constants $C_1, C_2, C_3 > 0$ such that the
following holds.
Pick any $\eta \in (0,1)$, and suppose
\[
\epsilon \leq C_1 \cdot \frac{\lambdamin}{k} ,
\qquad
N \geq C_2 \cdot \biggl( \log(k) + \log\log\Bigl(
\frac{\lambdamax}{\eps} \Bigr) \biggr)
,
\]
and
\begin{multline*}
\sqrt{\frac{\ln(L/\log_2(k/\eta))}{\ln(k)}}
\cdot \Biggl( 1 - \frac{\ln(\ln(L/\log_2(k/\eta))) +
C_3}{4\ln(L/\log_2(k/\eta))} -
\sqrt{\frac{\ln(8)}{\ln(L/\log_2(k/\eta))}} \Biggr)
\\
\geq 1.02 \Biggl( 1 + \sqrt{\frac{\ln(4)}{\ln(k)}}
\Biggr)
.
\end{multline*}
(Note that the condition on $L$ holds with $L = \poly(k) \log(1/\eta)$.)
Suppose that Algorithm~\ref{alg:robustpower} is iteratively called $k$
times, where the input tensor is $\hat{T}$ in the first call, and in each
subsequent call, the input tensor is the deflated tensor returned by the
previous call.
Let $(\hat{v}_1,\hat\lambda_1), (\hat{v}_2,\hat\lambda_2), \dotsc,
(\hat{v}_k,\hat\lambda_k)$ be the sequence of estimated
eigenvector/eigenvalue pairs returned in these $k$ calls.
With probability at least $1-\eta$, there exists a permutation $\pi$ on
$[k]$ such that
\[
\|v_{\pi(j)}-\hat{v}_j\| \leq 8 \epsilon/\lambda_{\pi(j)}
, \qquad
|\lambda_{\pi(j)}-\hat\lambda_j| \leq 5\epsilon , \quad \forall j \in [k]
,
\]
and
\[
\biggl\|
T - \sum_{j=1}^k \hat\lambda_j \hat{v}_j^{\otimes 3}
\biggr\| \leq 55\eps .
\]
\end{thm2}

\begin{proof}
We prove by induction that for each $i \in [k]$ (corresponding to the
$i$-th call to Algorithm~\ref{alg:robustpower}), with probability at least
$1-i\eta/k$, there exists a permutation $\pi$ on $[k]$ such that the
following assertions hold.
\begin{enumerate}
\item For all $j \leq i$, $\|v_{\pi(j)}-\hat{v}_j\| \leq 8
\epsilon/\lambda_{\pi(j)}$ and $|\lambda_{\pi(j)}-\hat\lambda_j| \leq 12
\epsilon$.

\item The error tensor
\begin{align*}
\tilde{E}_{i+1} & :=
\biggl( \hat{T} - \sum_{j \leq i} \hat\lambda_j \hat{v}_j^{\otimes 3} \biggr)
- \sum_{j \geq i+1} \lambda_{\pi(j)} v_{\pi(j)}^{\otimes 3}
= E + \sum_{j \leq i}
\Bigl(
\lambda_{\pi(j)} v_{\pi(j)}^{\otimes 3}
-
\hat\lambda_j \hat{v}_j^{\otimes 3}
\Bigr)
\end{align*}
satisfies
\begin{align}
\|\tl{E}_{i+1}(I,u,u)\|
& \leq 56\eps , \quad \forall u \in S^{k-1} ;
\label{eq:regular-bound} \\
\|\tl{E}_{i+1}(I,u,u)\|
& \leq 2\eps , \quad \forall u \in S^{k-1} \ \text{s.t.}
\ \exists j \geq i+1 \centerdot (u^\t v_{\pi(j)})^2 \geq 1 -
(168\eps/\lambda_{\pi(j)})^2
.
\label{eq:smaller-bound}
\end{align}

\end{enumerate}
We actually take $i = 0$ as the base case, so we can ignore the first
assertion, and just observe that for $i = 0$,
\[
\tilde{E}_1 = \hat{T} - \sum_{j=1}^k \lambda_i v_i^{\otimes 3} = E .
\]
We have $\|\tilde{E}_1\| = \|E\| = \eps$, and therefore the second
assertion holds.

Now fix some $i \in [k]$, and assume as the inductive hypothesis that, with
probability at least $1 - (i-1)\eta/k$, there exists a permutation $\pi$
such that two assertions above hold for $i-1$ (call this $\event_{i-1}$).
The $i$-th call to Algorithm~\ref{alg:robustpower} takes as input
\[ \tilde{T}_i :=
\hat{T} - \sum_{j \leq i-1} \hat\lambda_j \hat{v}_j^{\otimes 3} , \]
which is intended to be an approximation to
\[ T_i := \sum_{j \geq i} \lambda_{\pi(j)} v_{\pi(j)}^{\otimes 3}
.
\]
Observe that
\[ \tilde{T}_i - T_i = \tl{E}_i , \]
which satisfies the second assertion in the inductive hypothesis.
We may write $T_i = \sum_{l=1}^k \tlambda_l v_l^{\otimes 3}$ where
$\tlambda_l = \lambda_l$ whenever $\pi^{-1}(l) \geq i$, and $\tlambda_l =
0$ whenever $\pi^{-1}(l) \leq i-1$.
This form is used when referring to $\tilde{T}$ or the $\tlambda_i$ in
preceding lemmas (in particular, Lemma~\ref{lem:initialization} and
Lemma~\ref{lem:single}).

By Lemma~\ref{lem:initialization}, with conditional probability at least
$1-\eta/k$ given $\event_{i-1}$, at least one of $\th{0}^{(\tau)}$ for
$\tau \in [L]$ is $\gamma$-separated relative to $\pi(j_{\max})$, where
$j_{\max} := \arg\max_{j \geq i} \lambda_{\pi(j)}$, (for $\gamma = 0.01$;
call this $\event_i'$; note that the application of Lemma~\ref{lem:initialization}
determines $C_3$).
Therefore $\Pr[ \event_{i-1} \cap \event_i' ] = \Pr[ \event_i' |
\event_{i-1} ] \Pr[ \event_{i-1} ] \geq (1 - \eta/k) (1 - (i-1) \eta /k)
\geq 1 - i \eta /k$.
It remains to show that $\event_{i-1} \cap \event_i' \subseteq \event_i$;
so henceforth we condition on $\event_{i-1} \cap \event_i'$.

Set
\begin{equation}
    C_1 := \min\left\{
        (56\cdot9\cdot 102)^{-1},
        (100\cdot 168)^{-1},
        \Delta' \textup{ from Lemma~\ref{lem:deflation} with } \Delta = 1/50
    \right\}.
    \label{eq:C1:defn}
\end{equation}
For all $\tau \in [L]$ such that $\th{0}^{(\tau)}$ is $\gamma$-separated
relative to $\pi(j_{\max})$, we have (i) $|\th{j_{\max},0}^{(\tau)}| \geq
1/\sqrt{k}$, and (ii) that by Lemma~\ref{lem:single} (using $\teps/p :=
2\eps$, $\kappa := 1$, and $i^* := \pi(j_{\max})$, and providing $C_2$),
\begin{align*}
|\tilde{T}_i(\th{N}^{(\tau)},\th{N}^{(\tau)},\th{N}^{(\tau)})
- \lambda_{\pi(j_{\max})}|
& \leq 5\eps
\end{align*}
(notice by definition that $\gamma \geq 1/100$ implies
$\gamma_0 \geq 1 - /(1 + \gamma) \geq 1/101$,
thus it follows from the bounds on the other quantities that
$
\teps
= 2p\eps
\leq 56 C_1 \cdot \frac{\lambdamin}{k}
< \frac{\gamma_0}{2(1+8\kappa)}
\cdot \tlambdamin \cdot \th{i^*,0}^2$
as necessary).
Therefore $\th{N} := \th{N}^{(\tau^*)}$ must satisfy
\[
\tilde{T}_i(\th{N},\th{N},\th{N})
= \max_{\tau \in [L]}
\tilde{T}_i(\th{N}^{(\tau)},\th{N}^{(\tau)},\th{N}^{(\tau)})
\geq \max_{j \geq i} \lambda_{\pi(j)} - 5\eps
= \lambda_{\pi(j_{\max})} - 5\eps
.
\]
On the other hand, by the triangle inequality,
\begin{align*}
\tilde{T}_i(\th{N},\th{N},\th{N})
& \leq \sum_{j \geq i} \lambda_{\pi(j)} \th{\pi(j),N}^3
+ |\tilde{E}_i(\th{N},\th{N},\th{N})|
\\
& \leq \sum_{j \geq i}
\lambda_{\pi(j)} |\th{\pi(j),N}| \th{\pi(j),N}^2
+ 56\eps
\\
& \leq \lambda_{\pi(j^*)} |\th{\pi(j^*),N}|
+ 56\eps
\end{align*}
where $j^* := \arg\max_{j \geq i} \lambda_{\pi(j)} |\th{\pi(j),N}|$.
Therefore
\[
\lambda_{\pi(j^*)} |\th{\pi(j^*),N}|
\geq \lambda_{\pi(j_{\max})} - 5\eps - 56\eps
\geq \frac45 \lambda_{\pi(j_{\max})}
.
\]
Squaring both sides and using the fact that $\th{\pi(j^*),N}^2 +
\th{\pi(j),N}^2 \leq 1$ for any $j \neq j^*$,
\begin{align*}
\bigl( \lambda_{\pi(j^*)} \th{\pi(j^*),N} \bigr)^2
& \geq
\frac{16}{25} \bigl( \lambda_{\pi(j_{\max})} \th{\pi(j^*),N} \bigr)^2
+ \frac{16}{25} \bigl( \lambda_{\pi(j_{\max})} \th{\pi(j),N} \bigr)^2
\\
& \geq
\frac{16}{25} \bigl( \lambda_{\pi(j^*)} \th{\pi(j^*),N} \bigr)^2
+ \frac{16}{25} \bigl( \lambda_{\pi(j)} \th{\pi(j),N} \bigr)^2
\end{align*}
which in turn implies
\[
\lambda_{\pi(j)} |\th{\pi(j),N}|
\leq \frac34 \lambda_{\pi(j^*)} |\th{\pi(j^*),N}|
, \quad j \neq j^*
.
\]
This means that $\th{N}$ is $(1/4)$-separated relative to $\pi(j^*)$.
Also, observe that
\[
|\th{\pi(j^*),N}|
\geq \frac45 \cdot
\frac{\lambda_{\pi(j_{\max})}}{\lambda_{\pi(j^*)}}
\geq \frac45 ,
\quad
\frac{\lambda_{\pi(j_{\max})}}{\lambda_{\pi(j^*)}}
\leq \frac54 .
\]
Therefore by Lemma~\ref{lem:single} (using $\teps/p := 2\eps$, $\gamma :=
1/4$, and $\kappa := 5/4$), executing another $N$ power iterations starting
from $\th{N}$ gives a vector $\hat\theta$ that satisfies
\[
\|\hat\theta - v_{\pi(j^*)}\|
\leq \frac{8\eps}{\lambda_{\pi(j^*)}}
, \qquad
|\hat\lambda - \lambda_{\pi(j^*)}|
\leq 5\eps
.
\]
Since $\hat{v}_i = \hat\theta$ and $\hat\lambda_i = \hat\lambda$, the first
assertion of the inductive hypothesis is satisfied, as we can modify the
permutation $\pi$ by swapping $\pi(i)$ and $\pi(j^*)$ without affecting the
values of $\{ \pi(j) : j \leq i-1 \}$ (recall $j^* \geq i$).

We now argue that $\tilde{E}_{i+1}$ has the required properties to
complete the inductive step.
By Lemma~\ref{lem:deflation} (using $\teps := 5\eps$ and $\Delta := 1/50$,
the latter providing one upper bound on $C_1$ as per \eqref{eq:C1:defn}),
we have for any unit vector $u \in S^{k-1}$,
\begin{equation} \label{eq:approx-bound}
\Biggl\|
\biggl(
\sum_{j \leq i}
\Bigl(
\lambda_{\pi(j)} v_{\pi(j)}^{\otimes 3}
-
\hat\lambda_j \hat{v}_j^{\otimes 3}
\Bigr)
\biggr)(I,u,u)
\Biggr\|
\leq \biggl( 1/50 + 100 \sum_{j=1}^i (u^\t v_{\pi(j)})^2 \biggr)^{1/2}
5\eps
\leq 55\eps
.
\end{equation}
Therefore by the triangle inequality,
\[
\|\tilde{E}_{i+1}(I,u,u)\|
\leq \|E(I,u,u)\|
+ \Biggl\|
\biggl(
\sum_{j \leq i}
\Bigl(
\lambda_{\pi(j)} v_{\pi(j)}^{\otimes 3}
-
\hat\lambda_j \hat{v}_j^{\otimes 3}
\Bigr)
\biggr)(I,u,u)
\Biggr\|
\leq 56\eps
.
\]
Thus the bound~\eqref{eq:regular-bound} holds.

To prove that~\eqref{eq:smaller-bound} holds, pick any unit vector $u \in
S^{k-1}$ such that there exists $j' \geq i+1$ with $(u^\t
v_{\pi(j')})^2 \geq 1 - (168\eps/\lambda_{\pi(j')})^2$.
We have, via the second bound on $C_1$ in \eqref{eq:C1:defn} and the corresponding
assumed bound $\epsilon \leq C_1 \cdot \frac{\lambdamin}{k}$,
\[
100 \sum_{j=1}^i (u^\t v_{\pi(j)})^2
\leq 100 \Bigl( 1 - (u^\t v_{\pi(j')})^2 \Bigr)
\leq 100 \biggl( \frac{168\eps}{\lambda_{\pi(j')}} \biggr)^2
\leq \frac1{50}
,
\]
and therefore
\[
\biggl( 1/50 + 100 \sum_{j=1}^i (u^\t v_{\pi(j)})^2 \biggr)^{1/2} 5\eps
\leq (1/50 + 1/50)^{1/2} 5\eps \leq \eps .
\]
By the triangle inequality, we have $\|\tilde{E}_{i+1}(I,u,u)\| \leq
2\eps$.
Therefore~\eqref{eq:smaller-bound} holds, so the second assertion of the
inductive hypothesis holds.
Thus $\event_{i-1} \cap \event_i' \subseteq \event_i$, and $\Pr[ \event_i ]
\geq \Pr[ \event_{i-1} \cap \event_i' ] \geq 1 - i\eta/k$.
We conclude that by the induction principle, there exists a permutation
$\pi$ such that two assertions hold for $i = k$, with probability at least
$1-\eta$.

From the last induction step ($i = k$), it is also clear
from~\eqref{eq:approx-bound} that $\|T - \sum_{j=1}^k \hat\lambda_j
\hat{v}_j^{\otimes 3}\| \leq 55\eps$ (in $\event_{k-1} \cap \event_k'$).
This completes the proof of the theorem.
\end{proof}

\section{Variant of Robust Power Method that uses a Stopping Condition}
\label{app:stopping}

In this section we analyze a variant of Algorithm~\ref{alg:robustpower}
that uses a stopping condition.
The variant is described in Algorithm~\ref{alg:stopping}.
The key difference is that the inner for-loop is repeated until a stopping
condition is satisfied (rather than explicitly $L$ times).
The stopping condition ensures that the power iteration is converging to an
eigenvector, and it will be satisfied within $\poly(k)$ random restarts
with high probability.
The condition depends on one new quantity, $r$, which should be set to $r
:= k - \text{\# deflation steps so far}$ (\emph{i.e.}, the first call to
Algorithm~\ref{alg:stopping} uses $r = k$, the second call uses $r = k-1$,
and so on).

\begin{algorithm}
\caption{Robust tensor power method with stopping condition}
\label{alg:stopping}
\begin{algorithmic}[1]
\renewcommand{\algorithmicrequire}{\textbf{input}}
\renewcommand{\algorithmicensure}{\textbf{output}}
\REQUIRE symmetric tensor $\tilde{T} \in \R^{k \times k \times k}$, number
of iterations $N$, expected rank $r$.

\ENSURE the estimated eigenvector/eigenvalue pair; the deflated tensor.

\REPEAT

\STATE Draw $\th{0}$ uniformly at random from the unit sphere in $\R^k$.

\FOR{$t = 1$ to $N$}

\STATE Compute power iteration update
\begin{eqnarray}
\th{t} & := &
\frac{\tilde{T}(I, \th{t-1}, \th{t-1})}
{\|\tilde{T}(I, \th{t-1}, \th{t-1})\|}
\label{eq:power-update2}
\end{eqnarray}

\ENDFOR

\UNTIL the following stopping condition is satisfied:
\[
|\tilde{T}(\th{N},\th{N},\th{N})|
\geq \max\biggl\{ \frac{1}{2\sqrt{r}} \|\tilde{T}\|_F , \
\frac1{1.05} \|\tilde{T}(I,I,\th{N})\|_F
\biggr\}
.
\]

\STATE Do $N$ power iteration updates \eqref{eq:power-update2} starting from
$\th{N}$ to obtain $\hat\theta$, and set $\hat\lambda :=
\tilde{T}(\hat\theta,\hat\theta,\hat\theta)$.

\RETURN the estimated eigenvector/eigenvalue pair
$(\hat\theta,\hat\lambda)$; the deflated tensor $\tilde{T} - \hat\lambda \
\hat\theta^{\otimes 3}$.

\end{algorithmic}
\end{algorithm}

\subsection{Stopping Condition Analysis}

For a matrix $A$, we use $\|A\|_F := (\sum_{i,j} A_{i,j}^2)^{1/2}$ to
denote its Frobenius norm.
For a third-order tensor $A$, we use $\|A\|_F := (\sum_i
\|A(I,I,e_i)\|_F^2)^{1/2} = (\sum_i \|A(I,I,v_i)\|_F^2)^{1/2}$.

Define $\tilde{T}$ as before in~\eqref{eq:tildeT}:
\begin{equation*}
\tilde{T} := \sum_{i=1}^k \tlambda_i
v_i^{\otimes 3} + \tilde{E} .
\end{equation*}
We assume $\tilde{E}$ is a symmetric tensor such that, for some constant $p
> 1$,
\begin{align*}
\|\tilde{E}(I,u,u)\| & \leq \teps , \quad \forall u \in S^{k-1} ;
\\
\|\tilde{E}(I,u,u)\| & \leq \teps / p , \quad \forall u \in S^{k-1} \
\text{s.t.} \ (u^\t v_1)^2 \geq 1 - (3\teps/\tlambda_1)^2 ;
\\
\|\tilde{E}\|_F \leq \teps_F
.
\end{align*}
Assume that not all $\tlambda_i$ are zero, and define
\begin{align*}
\tlambdamin & := \min \{ \tlambda_i : i \in [k], \ \tlambda_i > 0 \} ,
& \tlambdamax & := \max \{ \tlambda_i : i \in [k] \} , \\
\ell & := |\{ i \in [k] : \tlambda_i > 0 \}| ,
& \tlambdaavg & := \biggl( \frac1{\ell}
\sum_{i=1}^k \tlambda_i^2 \biggr)^{1/2} .
\end{align*}

We show in Lemma~\ref{lem:stopping} that if the stopping condition is
satisfied by a vector $\theta$, then it must be close to an eigenvector of
$\tilde{T}$.
Then in Lemma~\ref{lem:satisfy-stopping}, we show that the stopping
condition is satisfied by $\theta_N$ when $\theta_0$ is a good starting
point (as per the conditions of Lemma~\ref{lem:single}).

\begin{lem} \label{lem:stopping}
Fix any vector $\theta = \sum_{i=1}^k \theta_i v_i$, and let $i^* :=
\arg\max_{i \in [k]} \tlambda_i |\theta_i|$.
Assume that $\ell \geq 1$ and that for some $\alpha \in (0,1/20)$ and
$\beta \geq 2\alpha/\sqrt{k}$,
\[
\teps \leq \alpha \cdot \frac{\tlambdamin}{\sqrt{k}}
, \quad
\teps_F \leq \sqrt\ell \Bigl( \frac12 - \frac{\alpha}{\beta\sqrt{k}} \Bigr)
\cdot \tlambdaavg
.
\]
If the stopping condition
\begin{equation} \label{eq:stopping}
|\tilde{T}(\theta,\theta,\theta)|
\geq \max\biggl\{ \frac{\beta}{\sqrt\ell} \|\tilde{T}\|_F , \
\frac1{1+\alpha} \|\tilde{T}(I,I,\theta)\|_F
\biggr\}
\end{equation}
holds, then
\begin{enumerate}
\item $\tlambda_{i^*} \geq \beta \tlambdaavg / 2$ and $\tlambda_{i^*}
|\theta_{i^*}| > 0$;

\item $\max_{i\neq i^*} \tlambda_i |\theta_i| \leq \sqrt{7\alpha} \cdot
\tlambda_{i^*} |\theta_{i^*}|$;

\item $\theta_{i^*} \geq 1 - 2\alpha$.
\end{enumerate}
\end{lem}
\begin{proof}
Without loss of generality, assume $i^* = 1$.
First, we claim that $\tlambda_1 |\theta_1| > 0$.
By the triangle inequality,
\begin{align*}
|\tilde{T}(\theta,\theta,\theta)|
\leq \sum_{i=1}^k \tlambda_i \theta_i^3 +
|\tilde{E}(\theta,\theta,\theta)|
\leq \sum_{i=1}^k \tlambda_i |\theta_i| \theta_i^2 + \teps
\leq \tlambda_1 |\theta_1| + \teps .
\end{align*}
Moreover,
\begin{align*}
\|\tilde{T}\|_F
& \geq \biggl\| \sum_{i=1}^k \tlambda_i v_i^{\otimes 3} \biggr\|_F
- \|\tilde{E}\|_F
\\
& = \biggl(
\sum_{j=1}^k \biggl\| \sum_{i=1}^k \tlambda_i v_i v_i^\t (v_i^\t v_j)
\biggr\|_F^2 \biggr)^{1/2} - \|\tilde{E}\|_F
\\
& = \biggl( \sum_{j=1}^k \biggl\| \tlambda_j v_j v_j^\t \biggr\|_F^2
\biggr)^{1/2} - \|\tilde{E}\|_F
\\
& = \biggl( \sum_{j=1}^k \tlambda_j^2 \biggr)^{1/2} - \|\tilde{E}\|_F
\\
& \geq \sqrt\ell \tlambdaavg - \teps_F .
\end{align*}
By assumption, $|\tilde{T}(\theta,\theta,\theta)| \geq (\beta/\sqrt\ell)
\|\tilde{T}\|_F$,
so
\[
\tlambda_1 |\theta_1|
\geq \beta \tlambdaavg - \frac{\beta}{\sqrt\ell} \teps_F - \teps
\geq \beta \tlambdaavg - \beta \Bigl( \frac12 -
\frac{\alpha}{\beta\sqrt{k}} \Bigr) \tlambdaavg -
\frac{\alpha}{\sqrt{k}} \tlambdamin
\geq \frac{\beta}{2} \tlambdaavg
\]
where the second inequality follows from the assumptions on
$\teps$ and $\teps_F$.
Since $\beta > 0$, $\tlambdaavg > 0$, and $|\theta_1| \leq 1$, it follows
that
\[ \tlambda_1 \geq \frac{\beta}{2} \tlambdaavg , \quad
\tlambda_1 |\theta_1| > 0 . \]
This proves the first claim.

Now we prove the second claim.
Define $\tilde{M} := \tilde{T}(I,I,\theta) = \sum_{i=1}^k \tlambda_i
\theta_i v_i v_i^\t + \tilde{E}(I,I,\theta)$ (a symmetric $k \times k$
matrix), and consider its eigenvalue decomposition
\[ \tilde{M} = \sum_{i=1}^k \phi_i u_i u_i^\t \]
where, without loss of generality, $|\phi_1| \geq |\phi_2| \geq \dotsb \geq
|\phi_k|$ and $\{ u_1, u_2, \dotsc, u_k \}$ is an orthonormal basis.
Let $M := \sum_{i=1}^k \tlambda_i \theta_i v_i v_i^\t$, so $\tilde{M} = M +
\tilde{E}(I,I,\theta)$.
Note that the $\tlambda_i|\theta_i|$ and $|\phi_i|$ are the singular values
of $M$ and $\tilde{M}$, respectively.
We now show that the assumption on $|\tilde{T}(\theta,\theta,\theta)|$
implies that almost all of the energy in $M$ is contained in its top
singular component.

By Weyl's theorem,
\begin{align*}
|\phi_1|
\leq \tlambda_1 |\theta_1| + \|\tilde{M} - M\|
\leq \tlambda_1 |\theta_1| + \teps
.
\end{align*}
Next, observe that the assumption $\|\tilde{T}(I,I,\theta)\|_F \leq (1 +
\alpha) \tilde{T}(\theta,\theta,\theta)$ is equivalent to $(1+\alpha)
\theta^\t \tilde{M} \theta \geq \|\tilde{M}\|_F$.
Therefore, using the fact that $|\phi_1| = \max_{u \in S^{k-1}} |u^\t
\tilde{M} u|$, the triangle inequality, and the fact $\|A\|_F \leq \sqrt{k}
\|A\|$ for any matrix $A \in \R^{k \times k}$,
\begin{align}
(1+\alpha) |\phi_1|
\geq (1+\alpha) \theta^\t \tilde{M} \theta
& \geq \|\tilde{M}\|_F
\label{eq:M-rank1}
\\
& \geq \biggl\| \sum_{i=1}^k \tlambda_i \theta_i v_i v_i^\t \biggr\|_F -
\bigl\| \tilde{E}(I,I,\theta) \bigr\|_F
\nonumber \\
& \geq \biggl( \sum_{i=1}^k \tlambda_i^2 \theta_i^2 \biggr)^{1/2} -
\sqrt{k} \|\tilde{E}(I,I,\theta)\|
\nonumber \\
& \geq \biggl( \sum_{i=1}^k \tlambda_i^2 \theta_i^2 \biggr)^{1/2} -
\sqrt{k} \teps .
\nonumber
\end{align}
Combining these bounds on $|\phi_1|$ gives
\begin{equation} \label{eq:combine-phi1}
\tlambda_1 |\theta_1| + \teps
\geq
\frac1{1+\alpha} \Biggl[ \biggl( \sum_{i=1}^k \tlambda_i^2 \theta_i^2
\biggr)^{1/2} - \sqrt{k} \teps \Biggr]
.
\end{equation}
The assumption $\teps \leq \alpha \tlambdamin / \sqrt{k}$ implies that
\[ \sqrt{k} \teps \leq \alpha\tlambdamin \leq \alpha \biggl( \sum_{i=1}^k
\tlambda_i^2 \theta_i^2 \biggr)^{1/2} . \]
Moreover, since $\tlambda_1 |\theta_1| > 0$ (by the first claim) and
$\tlambda_1 |\theta_1| = \max_{i \in [k]} \tlambda_i |\theta_i|$, it
follows that
\begin{equation} \label{eq:lambdatheta-lb}
\tlambda_1 |\theta_1|
\geq \tlambdamin \max_{i \in [k]} |\theta_i|
\geq \frac{\tlambdamin}{\sqrt{k}} ,
\end{equation}
so we also have
\[ \teps \leq \alpha \tlambda_1 |\theta_1| . \]
Applying these bounds on $\teps$ to~\eqref{eq:combine-phi1}, we obtain
\begin{equation*}
\tlambda_1 |\theta_1|
\geq
\frac{1-\alpha}{(1+\alpha)^2}
\biggl( \sum_{i=1}^k \tlambda_i^2 \theta_i^2 \biggr)^{1/2}
\geq
\frac{1-\alpha}{(1+\alpha)^2}
\biggl( \tlambda_1^2 \theta_1^2 + \max_{i\neq1} \tlambda_i^2 \theta_i^2
\biggr)^{1/2}
\end{equation*}
which in turn implies (for $\alpha \in (0,1/20)$)
\begin{equation*}
\max_{i\neq1} \tlambda_i^2 \theta_i^2
\leq \biggl( \frac{(1+\alpha)^4}{(1-\alpha)^2} - 1 \biggr)
\cdot \tlambda_1^2 \theta_1^2
\leq 7\alpha \cdot \tlambda_1^2 \theta_1^2
.
\end{equation*}
Therefore $\max_{i\neq1} \tlambda_i |\theta_i| \leq \sqrt{7\alpha} \cdot
\tlambda_1 |\theta_1|$, proving the second claim.

Now we prove the final claim.
This is done by (i) showing that $\theta$ has a large projection onto
$u_1$, (ii) using an SVD perturbation argument to show that $\pm u_1$ is
close to $v_1$, and (iii) concluding that $\theta$ has a large projection
onto $v_1$.

We begin by showing that $(u_1^\t \theta)^2$ is large.
Observe that from~\eqref{eq:M-rank1}, we have $(1+\alpha)^2 \phi_1^2 \geq
\|\tilde{M}\|_F^2 \geq \phi_1^2 + \max_{i\neq1} \phi_i^2$, and therefore
\[
\max_{i\neq1} |\phi_i| \leq \sqrt{2\alpha+\alpha^2} \cdot |\phi_1| .
\]
Moreover, by the triangle inequality,
\begin{align*}
|\theta^\t \tilde{M} \theta|
& \leq \sum_{i=1}^k |\phi_i| (u_i^\t \theta)^2
\\
& \leq |\phi_1| (u_1^\t \theta)^2 + \max_{i\neq1} |\phi_i| \bigl( 1 -
(u_1^\t \theta)^2 \bigr)
\\
& = (u_1^\t \theta)^2 \bigl( |\phi_1| - \max_{i\neq1} |\phi_i| \bigr)
+ \max_{i\neq1} |\phi_i| .
\end{align*}
Using~\eqref{eq:M-rank1} once more, we have $|\theta^\t \tilde{M} \theta|
\geq \|\tilde{M}\|_F / (1+\alpha) \geq |\phi_1| / (1+\alpha)$, so
\begin{equation*}
(u_1^\t \theta)^2
\geq \frac{\frac{1}{1+\alpha} - \max_{i\neq1} \frac{|\phi_i|}{|\phi_1|}}
{1 - \max_{i\neq1} \frac{|\phi_i|}{|\phi_1|}}
= 1 - \frac{\alpha}{(1+\alpha)
\Bigl(1 - \max_{i\neq1} \frac{|\phi_i|}{|\phi_1|}\Bigr)}
\leq 1 - \frac{\alpha}{(1+\alpha) (1 - \sqrt{2\alpha+\alpha^2})}
.
\end{equation*}
Now we show that $(u_1^\t v_1)^2$ is also large.
By the second claim, the assumption on $\teps$,
and~\eqref{eq:lambdatheta-lb},
\[
\tlambda_1 |\theta_1| - \max_{i\neq1} \tlambda_i |\theta_i|
> (1 - \sqrt{7\alpha}) \cdot \tlambda_1 |\theta_1|
\geq (1 - \sqrt{7\alpha}) \cdot \tlambdamin / \sqrt{k}
.
\]
Combining this with Weyl's theorem gives
\[ |\phi_1| - \max_{i\neq1} \tlambda_i |\theta_i| \geq \tlambda_1
|\theta_1| - \teps - \max_{i\neq1} \tlambda_i |\theta_i| \geq (1 - (\alpha
+ \sqrt{7\alpha})) \cdot \tlambdamin / \sqrt{k} , \]
so we may apply
Wedin's theorem to obtain
\[
(u_1^\t v_1)^2
\geq 1 - \biggl( \frac{\|\tilde{E}(I,I,\theta)\|}
{|\phi_1| - \max_{i\neq1} \tlambda_i |\theta_i|} \biggr)^2
\geq 1 - \biggl( \frac{\alpha}{1 - (\alpha + \sqrt{7\alpha})}
\biggr)^2
.
\]
It remains to show that $\theta_1 = v_1^\t \theta$ is large.
Indeed, by the triangle inequality, Cauchy-Schwarz, and the above
inequalities on $(u_1^\t v_1)^2$ and $(u_1^\t \theta)^2$,
\begin{align*}
|v_1^\t \theta|
& = \biggl| \sum_{i=1}^k (u_i^\t v_1) (u_i^\t \theta) \biggr| \\
& \geq |u_1^\t v_1| |u_1^\t \theta| - \sum_{i=2}^k |u_i^\t v_1| |u_i^\t
\theta|
\\
& \geq |u_1^\t v_1| |u_1^\t \theta| -
\biggl( \sum_{i=2}^k (u_i^\t v_1)^2 \biggr)^{1/2}
\biggl( \sum_{i=2}^k (u_i^\t \theta)^2 \biggr)^{1/2}
\\
& = |u_1^\t v_1| |u_1^\t \theta| - \biggl( \Bigl( 1 - (u_i^\t v_1)^2
\Bigr) \Bigl(1 - (u_i^\t \theta)^2 \Bigr) \biggr)^{1/2}
\\
& \geq
\Biggl(
\biggl(
1 - \frac{\alpha}{(1+\alpha)(1 - \sqrt{2\alpha + \alpha^2})}
\biggr)
\biggl(
1 - \biggl( \frac{\alpha}{1 - (\alpha + \sqrt{7\alpha})} \biggr)^2
\biggr)
\Biggr)^{1/2}
\\
& \qquad
- \Biggl(
\frac{\alpha}{(1+\alpha)(1 - \sqrt{2\alpha + \alpha^2})}
\cdot \biggl( \frac{\alpha}{1 - (\alpha + \sqrt{7\alpha})} \biggr)^2
\Biggr)^{1/2}
\\
& \geq 1 - 2\alpha
\end{align*}
for $\alpha \in (0,1/20)$.
Moreover, by assumption we have $\tilde{T}(\theta,\theta,\theta) \geq 0$,
and
\begin{align*}
\tilde{T}(\theta,\theta,\theta)
& = \sum_{i=1}^k \tlambda_i \theta_i^3
+ \tilde{E}(\theta,\theta,\theta)
\\
& = \tlambda_1 \theta_1^3
+ \sum_{i=2}^k \tlambda_i \theta_i^3
+ \tilde{E}(\theta,\theta,\theta)
\\
& \leq \tlambda_1 \theta_1^3
+ \max_{i \neq 1} \tlambda_i |\theta_i| \sum_{i=2}^k \theta_i^2
+ \teps
\\
& \leq \tlambda_1 \theta_1^3
+ \sqrt{7\alpha} \tlambda_1 |\theta_1| (1 - \theta_1^2)
+ \teps
\quad \text{(by the second claim)} \\
& \leq \tlambda_1 |\theta_1|^3 \biggl(
\sign(\theta_1) + \frac{\sqrt{7\alpha}}{(1-2\alpha)^2} - \sqrt{7\alpha} +
\frac{\alpha}{(1-2\alpha)^3}
\biggr)
\quad \text{(since $|\theta_1| \geq 1 - 2\alpha$)} \\
& < \tlambda_1 |\theta_1|^3 \Bigl( \sign(\theta_1) + 1 \Bigr)
\end{align*}
so $\sign(\theta_1) > -1$, meaning $\theta_1 > 0$.
Therefore $\theta_1 = |\theta_1| \geq 1 - 2\alpha$.
This proves the final claim.
\end{proof}

\begin{lem} \label{lem:satisfy-stopping}
Fix $\alpha , \beta \in (0,1)$.
Assume $\tlambda_{i^*} = \max_{i \in [k]} \tlambda_i$ and
\[
\teps \leq \min \biggl\{ \frac{\alpha}{5\sqrt{k} + 7} , \frac{1-\beta}{7}
\biggr\} \cdot \tlambda_{i^*}
, \quad
\teps_F \leq \sqrt\ell \cdot \frac{1-\beta}{2\beta} \cdot \tlambda_{i^*}
.
\]
To the conclusion of Lemma~\ref{lem:single}, it can be added that the
stopping condition~\eqref{eq:stopping} is satisfied by $\theta = \th{t}$.
\end{lem}
\begin{proof}
Without loss of generality, assume $i^* = 1$.
By the triangle inequality and Cauchy-Schwarz,
\begin{align*}
\|\tilde{T}(I,I,\th{t})\|_F
& \leq \tlambda_1|\th{1,t}| + \sum_{i\neq1} \lambda_i |\th{i,t}|
+ \|\tilde{E}(I,I,\th{t})\|_F
\leq \tlambda_1|\th{1,t}| + \tlambda_1 \sqrt{k} \biggl( \sum_{i\neq1}
\th{i,t}^2 \biggr)^{1/2}
+ \sqrt{k} \teps
\\
& \leq \tlambda_1|\th{1,t}| + \frac{3\sqrt{k}\teps}{p}
+ \sqrt{k} \teps
.
\end{align*}
where the last step uses the fact that $\th{1,t}^2 \geq 1 -
(3\teps/(p\tlambda_1))^2$.
Moreover,
\[
\tilde{T}(\th{t},\th{t},\th{t}) \geq \tlambda_1 -
\biggl( 27 \Bigl( \frac{\teps}{p\lambda_1} \Bigr)^2 + 2
\biggr) \frac{\teps}{p}
.
\]
Combining these two inequalities with the assumption on $\teps$ implies
that
\[ \tilde{T}(\th{t},\th{t},\th{t}) \geq \frac{1}{1+\alpha}
\|\tilde{T}(I,I,\th{t})\|_F . \]

Using the definition of the tensor Frobenius norm, we have
\[
\frac1{\sqrt\ell} \|\tilde{T}\|_F
\leq \frac1{\sqrt\ell} \biggl\| \sum_{i=1}^k \tlambda_i v_i^{\otimes 3}
\biggr\|_F + \frac1{\sqrt\ell} \|\tilde{E}\|_F
= \tlambdaavg + \frac1{\sqrt\ell} \|\tilde{E}\|_F
\leq \tlambdaavg + \frac1{\sqrt\ell} \teps_F
.
\]
Combining this with the above inequality implies
\[ \tilde{T}(I,I,\th{t}) \geq \frac{\beta}{\sqrt\ell} \|\tilde{T}\|_F . \]
Therefore the stopping condition~\eqref{eq:stopping} is satisfied.
\end{proof}

\subsection{Sketch of Analysis of Algorithm~\ref{alg:stopping}}

The analysis of Algorithm~\ref{alg:stopping} is very similar to the proof
of Theorem~\ref{thm:robustpower} for Algorithm~\ref{alg:robustpower}, so
here we just sketch the essential differences.

First, the guarantee afforded to Algorithm~\ref{alg:stopping} is somewhat
different than Theorem~\ref{thm:robustpower}.
Specifically, it is of the following form: (i) under appropriate
conditions, upon termination, the algorithm returns an accurate
decomposition, and (ii) the algorithm terminates after $\poly(k)$ random
restarts with high probability.

The conditions on $\eps$ and $N$ are the same (but for possibly different
universal constants $C_1, C_2$).
In Lemma~\ref{lem:stopping} and Lemma~\ref{lem:satisfy-stopping}, there is
reference to a condition on the Frobenius norm of $E$, but we may use the
inequality $\|E\|_F \leq k \|E\| \leq k \eps$ so that the condition is
subsumed by the $\eps$ condition.

Now we outline the differences relative to the proof of
Theorem~\ref{thm:robustpower}.
The basic structure of the induction argument is the same.
In the induction step, we argue that (i) if the stopping condition is
satisfied, then by Lemma~\ref{lem:stopping} (with $\alpha = 0.05$ and
$\beta = 1/2$), we have a vector $\th{N}$ such that, for some $j^* \geq i$,
\begin{enumerate}
\item $\lambda_{\pi(j^*)} \geq \lambda_{\pi(j_{\max})} / (4\sqrt{k})$;

\item $\th{N}$ is $(1/4)$-separated relative to $\pi(j^*)$;

\item $\th{\pi(j^*),N} \geq 4/5$;

\end{enumerate}
and (ii) the stopping condition is satisfied within $\poly(k)$ random
restarts (via Lemma~\ref{lem:initialization} and
Lemma~\ref{lem:satisfy-stopping}) with high probability.
We now invoke Lemma~\ref{lem:single} to argue that executing another $N$
power iterations starting from $\th{N}$ gives a vector $\hat\theta$ that
satisfies
\[
\|\hat\theta - v_{\pi(j^*)}\|
\leq \frac{8\eps}{\lambda_{\pi(j^*)}}
, \qquad
|\hat\lambda - \lambda_{\pi(j^*)}|
\leq 5\eps
.
\]
The main difference here, relative to the proof of
Theorem~\ref{thm:robustpower}, is that we use $\kappa := 4\sqrt{k}$ (rather
than $\kappa = O(1)$), but this ultimately leads to the same guarantee
after taking into consideration the condition $\eps \leq C_1 \lambdamin /
k$.
The remainder of the analysis is essentially the same as the proof of
Theorem~\ref{thm:robustpower}.

\section{Simultaneous Diagonalization for Tensor Decomposition}
\label{app:simul}

As discussed in the introduction, another standard approach to certain
tensor decomposition problems is to simultaneously diagonalize a collection
of similar matrices obtained from the given tensor.
We now examine this approach in the context of our latent variable
models, where
\begin{eqnarray*}
M_2 & = & \sum_{i=1}^k w_i \ \mu_i \otimes \mu_i \\
M_3 & = &
\sum_{i=1}^k w_i \ \mu_i \otimes \mu_i \otimes \mu_i .
\end{eqnarray*}
Let $V := [\mu_1|\mu_2|\dotsb|\mu_k]$ and $D(\eta) := \diag(\mu_1^\t \eta,
\mu_2^\t \eta, \dotsc, \mu_k^\t \eta)$, so
\begin{eqnarray*}
M_2 &  = &  V\diag(w_1,w_2, \ldots w_k) V^\t\\
M_3(I,I, \eta)
& = &
V \diag(w_1,w_2, \ldots w_k) D(\eta) V^\t
\end{eqnarray*}
Thus, the problem of determining the $\mu_i$ can be cast as a simultaneous
diagonalization problem: find a matrix $X$ such that $X^\t M_2 X$ and $X^\t
M_3(I,I,\eta) X$ (for all $\eta$) are diagonal.
It is easy to see that if the $\mu_i$ are linearly independent, then the
solution $X^\t = V^{\dag}$ is unique up to permutation and rescaling of the
columns.

With exact moments, a simple approach is as follows.
Assume for simplicity that $d = k$, and define
\[
M(\eta) := M_3(I, I, \eta) M_2^{-1} = V D(\eta) V^{-1}
.
\]
Observe that if the diagonal entries of $D(\eta)$ are distinct, then the
eigenvectors of $M(\eta)$ are the columns of $V$ (up to permutation and
scaling).
This criterion is satisfied almost surely when $\eta$ is chosen randomly
from a continuous distribution over $\R^k$.

The above technique (or some variant thereof) was previously used
to give the efficient learnability
results, where the computational and sample complexity bounds were
polynomial in relevant parameters of the problem, including the rank
parameter $k$ \citep{MR06,AHK12,SpectralLDA,HK13-mog}.
However, the specific polynomial dependence on $k$ was rather large due to
the need for the diagonal entries of $D(\eta)$ to be well-separated.
This is because with finite samples, $M(\eta)$ is only known
up to some perturbation, and thus the sample complexity bound depends
inversely in (some polynomial of) the separation of the diagonal entries of
$D(\eta)$.
With $\eta$ drawn uniformly at random from the unit sphere in $\R^k$, the
separation was only guaranteed to be roughly $1/k^{2.5}$~\citep{AHK12}
(while this may be a loose estimate, the instability is observed in
practice).
In contrast, using the tensor power method to approximately recover $V$
(and hence the model parameters $\mu_i$ and $w_i$) requires only a mild,
\emph{lower-order} dependence on $k$.

It should be noted, however, that the use of a single random choice of
$\eta$ is quite restrictive, and it is easy to see that a simultaneous
diagonalization of $M(\eta)$ for several choices of $\eta$
can be beneficial.
While the uniqueness of the eigendecomposition of $M(\eta)$ is
only guaranteed when the diagonal entries of $D(\eta)$ are distinct, the
\emph{simultaneous diagonalization} of $M(\eta^{(1)}),
M(\eta^{(2)}), \dotsc, M(\eta^{(m)})$ for vectors
$\eta^{(1)}, \eta^{(2)}, \dotsc, \eta^{(m)}$ is unique as long as the
\emph{columns} of
\begin{equation*}
\begin{bmatrix}
\mu_1^\t \eta^{(1)} &
\mu_2^\t \eta^{(1)} &
\dotsb &
\mu_k^\t \eta^{(1)} \\
\mu_1^\t \eta^{(2)} &
\mu_2^\t \eta^{(2)} &
\dotsb &
\mu_k^\t \eta^{(2)} \\
\vdots & \vdots & \ddots & \vdots \\
\mu_1^\t \eta^{(m)} &
\mu_2^\t \eta^{(m)} &
\dotsb &
\mu_k^\t \eta^{(m)}
\end{bmatrix}
\end{equation*}
are distinct (\emph{i.e.}, for each pair of column indices $i,j$, there
exists a row index $r$ such that the $(r,i)$-th and $(r,j)$-th entries are
distinct).
This is a much weaker requirement for uniqueness, and therefore may
translate to an improved perturbation analysis.
In fact, using the techniques discussed in Section~\ref{sec:estimation}, we
may even reduce the problem to an orthogonal simultaneous diagonalization,
which may be easier to obtain.
Furthermore, a number of robust numerical methods for (approximately)
simultaneously diagonalizing collections of matrices have been proposed and
used successfully in the literature~\citep[\emph{e.g.},][]{bunse1993numerical,CS93,PertDJ,CC96,ziehe2004fast}.
Another alternative and a more stable approach compared to full
diagonalization is a Schur-like method which finds a unitary matrix $U$
which simultaneously triangularizes the respective
matrices~\citep{corless1997reordered}.
It is an interesting open question whether these techniques can yield similar
improved learnability results and also enjoy the attractive computational
properties of the tensor power method.

\vskip 0.2in
\bibliography{power}

\end{document}